\documentclass[lettersize,journal]{IEEEtran}
\usepackage{array}
\usepackage[caption=false]{subfig}
\usepackage{textcomp}
\usepackage{stfloats}
\usepackage{url}
\usepackage{verbatim}
\usepackage{graphicx}
\usepackage{cite}
\usepackage{booktabs}
\usepackage{algorithmicx}%
\usepackage{algpseudocode}%
\usepackage{amsmath,amsfonts}
\usepackage{algorithm}
\usepackage{multirow}
\newcommand{\M}[1]{\mathbf{#1}}
\usepackage{xcolor}
\usepackage{color,xcolor}
\usepackage{colortbl}
\usepackage{multicol} 
\usepackage{multirow}
\usepackage{multicol}
\usepackage{booktabs}
\usepackage{cite}
\usepackage{hyperref}
\usepackage{bm} % For bold math symbols
\begin{document}

\title{Diff-Reg v2: Diffusion-Based Matching Matrix Estimation for Image Matching and 3D Registration}

\author{Qianliang Wu, Haobo Jiang, Yaqing Ding, Lei Luo, Jun Li, Jin Xie, Xiaojun Wu, Jian Yang

\thanks{Qianliang Wu, Lei Luo, Jun Li, and Jian Yang are with PCA Lab, Key Lab of Intelligent Perception and Systems for High-Dimensional Information of Ministry of Education, School of Computer Science and Engineering, Nanjing University of Science and Technology, Nanjing, China.}
\thanks{Haobo Jiang is with Nanyang Technological University, Singapore.}
\thanks{Yaqing Ding is with Visual Recognition Group, Faculty of Electrical Engineering, Czech Technical University in Prague, Prague, Czech Republic.}
\thanks{Jin Xie is with State Key Laboratory for Novel Software Technology, Nanjing University, Nanjing, China, and
School of Intelligence Science and Technology, Nanjing University, Suzhou, China.}
\thanks{Xiaojun Wu is  with the School of Artificial Intelligence and Computer Science,
Jiangnan University, Wuxi, China.}
%wu_xiaojun@jiangnan.edu.cn
% \thanks{$^{\dagger}$Corresponding Author} 
}
%haobo.jiang@ntu.edu.sg
\markboth{Journal of \LaTeX\ Class Files,~Vol.~14, No.~8, August~2021}%
{Shell \MakeLowercase{\textit{et al.}}: A Sample Article Using IEEEtran.cls for IEEE Journals}


\maketitle
\begin{abstract}

Reliable correspondence estimation is critical for a variety of registration tasks, including 2D image matching, 3D point cloud registration, and cross-modal 2D-to-3D alignment. These tasks are often complicated by scale variation, structural symmetries, and non-rigid deformations, which lead to ambiguity in matching. Traditional methods predominantly rely on geometric or semantic features guided by handcrafted priors, but such approaches are inherently limited in flexibility and expressiveness. Furthermore, single-step prediction architectures often suffer from local minima in complex scenarios.
This paper introduces a novel framework that formulates correspondence estimation as a denoising diffusion process in matching matrix space. An initially noisy matching matrix is iteratively refined toward an optimal solution through reverse diffusion sampling. The process is conducted within the space of doubly stochastic matrices for 3D-3D and 2D-3D tasks, and within a constrained subspace—regularized via dual-softmax projection—for 2D matching. Modality-specific matching matrix embedding strategies and a lightweight denoising module are introduced to enhance performance and efficiency.
The proposed framework enables diverse training sample generation, reveals new matching strategies, and imposes high-order geometric consistency. Extensive experiments on multiple benchmarks confirm its effectiveness across both 2D and 3D registration domains.
\end{abstract}
 

\section{Introduction}
\IEEEPARstart{T}he registration problem, which includes image registration and point cloud registration, is a critical and fundamental task in various applications such as 3D reconstruction, localization, VR/AR, and robotics. These applications typically require precise correspondences (matches) between point cloud pairs or image pairs to obtain reliable estimates of rigid transformations or non-rigid deformations. The goal of achieving accurate matchings is to identify the most significant correspondences~\cite{sun2021loftr, edstedt2024roma, zhong2009intrinsic, yu2023rotation, bai2020d3feat} with local or global semantic or geometric consistency~\cite{giang2023topicfm, li20232D3D, deng2018ppfnet, qin2022geometric, wu2023sgfeat}. However, this goal becomes particularly challenging in scenarios with large deformations, scale inconsistencies, and low overlap, which can lead to ambiguous matches.

\begin{figure}
      % \vspace{-0.5cm}666
      \centering
      \includegraphics[width=0.5\textwidth, height=4cm]{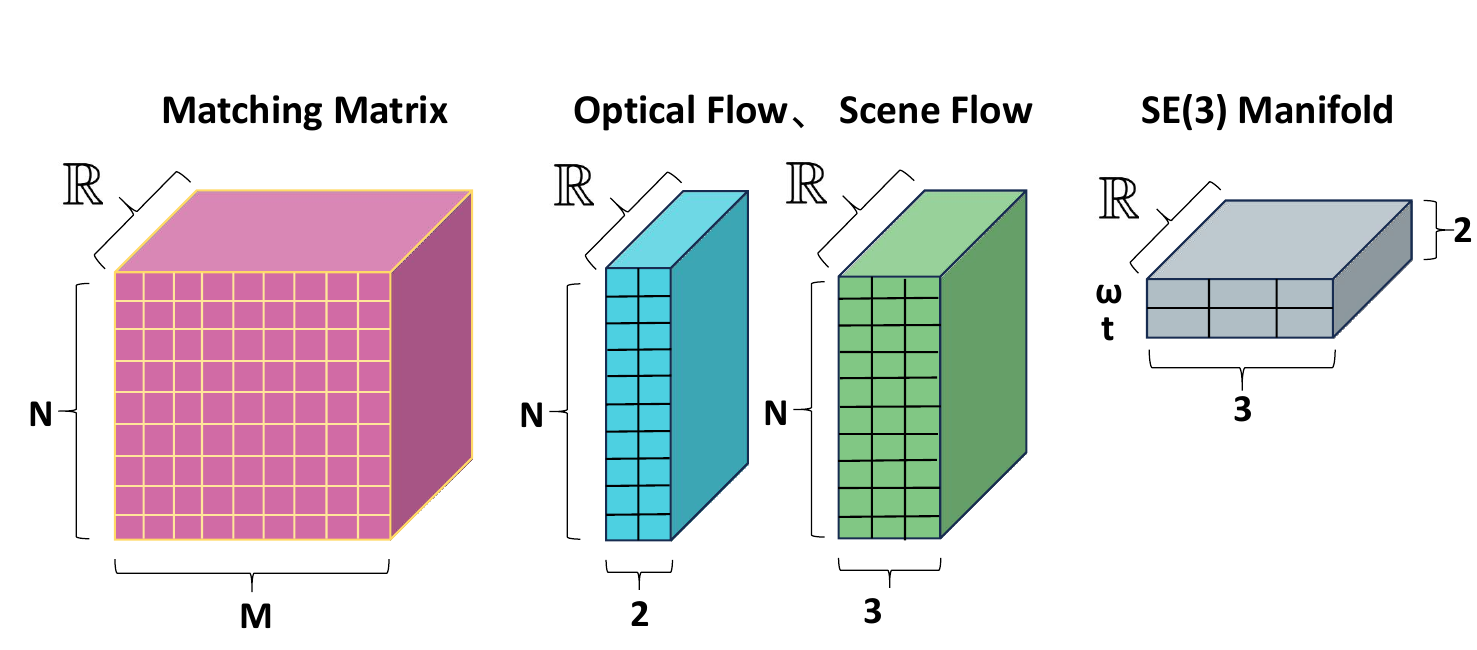}
      
    \caption{Depiction of the four spaces in which current methods primarily operate and innovate. Zoom in for details.} \label{diff_space}    
\vspace{-0.4cm}
\end{figure}

Recent advances in deep learning-based feature matching methods~\cite{qin2022geometric, li2022lepard, yew2022regtr, mei2023unsupervised, huang2021predator, yu2021cofinet, yu2023rotation, yao2023hunter, li20232D3D, sun2021loftr, edstedt2024roma, chen2022aspanformer} have significantly advanced registration tasks using feature backbones~\cite{thomas2019kpconv, he2016deep, ding2021repvgg} to extract point/pixel patches (subsampled) and their associated features. These methods typically compute a coarse-level patch matching matrix based on the patch features, then refine the fine-level local correspondence of the matched patches. To refine the matches, outlier rejection techniques~\cite{yang2020teaser, bai2021pointdsc, chen2022sc2, jiang2023robust, zhang20233dmac, xue2023imp, sarlin2020superglue, lindenberger2023lightglue} are employed to identify more salient and accurate inlier correspondences, utilizing various semantic and geometric strategies~\cite{zhong2009intrinsic, wu2023sgfeat, qi2019deep, fu2021robust, yan2024tri, yan2023desnet, yan2022learning, yan2022rignet, bokman2025affine}. Although significant progress has been made on specific problems, combining these strategies into a unified registration pipeline for complex cases involving multiple challenges remains highly difficult. Methods focusing on the low-dimensional SE(3) space~\cite{urain2023se,jiang2023se,chen2023diffusionpcr} or flow fields~\cite{edstedt2024roma,zhang2024diffsf,nam2023diffusion} (refer to Fig. \ref{diff_space}) often lack global geometric, semantic, or topological "structural" constraints on correspondences, resulting in the loss of critical but nonconforming matches. Furthermore, some approaches~\cite{qin2022geometric,li2022lepard,sun2021loftr,wang2024efficient} rely on single-step correspondence predictions, which may not always yield optimal or accurate results.

Inspired by diffusion models~\cite{ho2020denoising, song2020denoising, vignac2022digress, austin2021structured} and graph matching algorithms~\cite{xie2020fast, leordeanu2009integer, zaslavskiy2008path}, we introduce a diffusion model in the matching matrix space~\cite{86bd1ad6-50bb-38d0-9978-0966b4dfc6d3, sarlin2020superglue, sun2021loftr}. Taking a subset of the matching matrix space as a feasible solution domain, we can train a diffusion model to effectively learn a generalized optimization algorithm specifically designed for the matching matrix estimation and adapted to the characteristics of three registration tasks. Our matching matrix diffusion model consists of some main components: a forward diffusion process, a reverse denoising process, and a denoising module. The forward diffusion process gradually introduces Gaussian noise into the ground truth matching matrix, while the reverse denoising process iteratively refines the noisy matrix to the optimal one. For efficiency, we propose a novel and generalized lightweight denoising module that can be adapted to 2D and 3D registration tasks. For the implementation of the denoising module, we consistently implement "match-to-warp" encoding in different registration tasks.
Finally, we establish a specific variational lower bound associated with our diffusion matching model in the matching matrix space and a simplified version of the objective function to train our framework effectively.

\begin{figure*}
      % \vspace{-0.5cm}
      \centering
      \includegraphics[width=\textwidth, height=4cm]{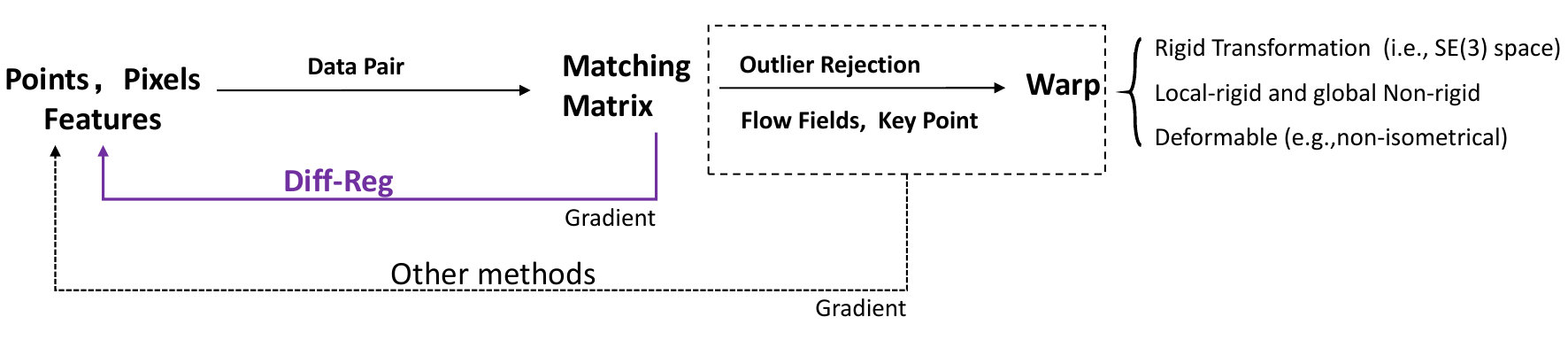}
      
    \caption{Difference between our method and other diffusion-based registration methods. Zoom in for details.} \label{diffmatrix_supervision}    
\vspace{-0.4cm}
\end{figure*}

\textbf{Advantage of diffusion in the matrix space.}
Tasks such as image or point cloud registration face numerous challenges, including scale inconsistency, large deformations, ambiguous matching, and low overlap. In particular, we can identify multiple entry points to these problems, based on which novel algorithms can be devised to improve each specific aspect accordingly. For example, several state-of-the-art studies~\cite{qin2022geometric,yao2023hunter,wu2023sgfeat,wugraph,mei2021cotreg,bai2021pointdsc,zhang20233dmac} have attempted to encode high-order combinational geometric consistency for 3d registration. Also, other works have tried to promote performance from accuracy~\cite{sun2021loftr,edstedt2024roma,edstedt2023dkm,tang2022quadtree,Guan_IJCV,Guan_TCYB} and speed~\cite{li2020dual, wang2024efficient,chen2022aspanformer,giang2023topicfm,lindenberger2023lightglue} for image registration problems.

However, when confronted with complex scenarios, these methods cannot exhaustively cover all effective matching strategies.  
One promoted strategy can be treated as a subset in the matrix space. For a given data pair, the matches generated by this strategy can result in a matching matrix. We can treat the relation between the strategies and the matching matrix feasible set as a surjective mapping. Similarly, one matching matrix ($\in \mathbb{R}^{N \times M}$) can result in a SE(3) ($\in \mathbb{R}^6$) warping or a deformable flow fields ($\in \mathbb{R}^{N \times 2}/\mathbb{R}^{N \times 3}$) (refer to Fig. \ref{diff_space}). The relation between the matching matrix feasible set and the SE(3) warping or flow fields feasible set is also subjective. The proposed diffusion process in the matching matrix space is a practical data augmentation technique that can generate additional and more effective training samples. Furthermore, the diffusion model in matrix space may discover unknown effective matching strategies.

During training, denoising the diffused matching matrix can provide any-order combinational geometric or semantic consistency constraints (refer to Fig. \ref{diffmatrix_supervision}).
During inference, the solution search is conducted within a large neighborhood area around the ground truth. The higher the solution's dimension, the more search directions it can offer. Specifically, in low-dimensional space, the sampling process may fail to identify small directions that could help escape local suboptimal points. In a high-dimensional space, the algorithm has more potential fine-level directions for exploration, allowing the search process to explore and optimize more effectively.

One matching matrix can be relaxed to a doubly stochastic matrix \cite{cuturi2013sinkhorn} or regularized into a more general "dual-softmax" matrix. The doubly stochastic matching matrix imposes a one-to-one mapping relation constraint for any two-view (and any two-modality) matching or registration problem. Similarly, the dual softmax can provide a weak constraint that there is at least one corresponding target point for every source point. Through these two relaxations, we can impose global geometric constraints directly on the matching matrix, thereby providing a holistic perspective. In contrast, existing methods that rely on flow fields or SE(3) warping can only enforce constraints in lower-dimensional spaces (see Fig.~\ref{diff_space}), which risks overlooking critical correspondences that deviate from the assumed transformations.
 
In conclusion, our diffusion model in matrix space offers several advantages over previous registration methods. The forward diffusion process generates diverse training samples, which act as data augmentation for both the feature backbone and the single-step prediction head. A limitation of single-step prediction methods is that if correspondences are predicted at a local minimum, subsequent outlier rejection or post-processing steps may encounter significant challenges. In contrast, our reverse denoising process, guided by the posterior distribution, increases the model's probability of escaping local minima. This lightweight design of the denoising module allows the denoising sampling process to explore a broader solution space, such as the matrix space, more quickly, increasing diversity and facilitating more iterative steps.

Our contributions are summarized as follows:
\begin{itemize}
\item We deploy the diffusion model in the matching atrix space for iteratively searching the optimal matching matrix through the reverse denoising sampling process. We also introduce the diffusion model of the matching matrix into the (semi-)dense image matching task.
\item We implement the matching matrix encoding module and the denoising module in a unified manner for 2D image matching and 3D registration tasks. Additionally, we utilized the pre-trained depth estimation model to convert the 2D-3D registration to 3D registration. The lightweight design of our reverse denoising module results in faster convergence in the reverse sampling process.
\item We conducted comprehensive experiments on the real-world datasets 4DMatch~\cite{li2022non}, 3DMatch~\cite{zeng20173dmatch} RGB-D Scenes V2~\cite{zeng20173dmatch}, 7Scenes~\cite{shotton2013scene}  MegaDepth~\cite{li2018megadepth}, and Scannet~\cite{dai2017Scannet} datasets to validate the effectiveness of our diffusion matching model on 3D registration, 2D-3D registration, and 2D image matching tasks. 
\end{itemize}

\textbf{Differences from Previous Work}:
a) We revisit and critically analyze the rationale behind Diff-Reg v1~\cite{wu2023diff, wu2024diff}, exploring the underlying reasons for its effectiveness. Two theorems are introduced to explain and support the motivation of our method.
b) We propose a diffusion-based approach for matching matrix estimation in (semi-)dense image matching problems.
c) We design the matching matrix encoding module in a unified manner across both image matching and 3D registration tasks, ensuring consistency and seamless integration. Although the implementation is not flawless, the results are nevertheless promising.
d) Based on ELoFTR~\cite{wang2024efficient}, we develop a new framework for training the matching matrix denoising module specifically for the image matching task.
e) We conduct experiments on the MegaDepth~\cite{li2018megadepth} and Scannet~\cite{dai2017Scannet} datasets to validate the effectiveness of our method in image matching.

\section{Related work}

\subsection{3D Registration}
The registration problem estimates the transformation between the point cloud and the image-to-point-cloud pair. Recently, there have been significant advancements in feature learning-based methods for point cloud registration. Many of these state-of-the-art approaches, such as~\cite{bai2020d3feat,huang2021predator,yu2021cofinet,qin2022geometric,yu2022riga,yu2023peal}, leverage a backbone architecture similar to KPConv~\cite{thomas2019kpconv} to downsample points and generate features with larger receptive fields. To further enhance the performance of these methods, they integrate prior knowledge and incorporate learnable outlier rejection modules. For instance, GeoTR~\cite{qin2022geometric} introduces angle-wise and edge-wise embeddings into the transformer encoder, while RoITr~\cite{yu2023rotation} integrates local Point Pair Features (PPF)~\cite{deng2018ppf} to improve rotation invariance.

In addition to feature learning-based methods, another category of registration methods focuses on outlier rejection of candidate correspondences. For instance, PointDSC~\cite{bai2021pointdsc} utilizes a maximum clique algorithm in the local patch to cluster inlier correspondences. SC2-PCR~\cite{chen2022sc2} constructs a second-order consistency graph for candidate correspondences and theoretically demonstrates its robustness. Building on the second-order consistency graph proposed by SC2-PCR~\cite{chen2022sc2}, MAC~\cite{zhang20233dmac} introduces a variant of maximum clique algorithms to generate more reliable candidate inlier correspondences. Moreover, methods such as PEAL~\cite{yu2023peal} and DiffusionPCR~\cite{chen2023diffusionpcr} employ an iterative refinement strategy to enhance the overlap prior information obtained from a pre-trained GeoTr~\cite{qin2022geometric}. 

Recently, significant advancements have been made in 2D-3D registration methods~\cite{Li2021DeepI2PIC, Wang2021P2NetJD, Wang2023FreeRegIC,li20232D3D}. These methods face similar challenges to 3D registration tasks, with the additional complexity of scale inconsistency caused by the perspective projection of images. To address the issue of scale inconsistency, we propose the incorporation of a pre-trained feature backbone, DINO v2~\cite{oquab2023dinov2}, which offers superior multiscale features. Additionally, implementing diffused data augmentation in our diffusion matching model can enhance the ability to identify prominent combinational and consistent correspondences.

Recently, the diffusion model~\cite{ho2020denoising,song2019generative,song2020denoising} has made significant development in many fields, including human pose estimation~\cite{Gong2022DiffPoseTM, Shan2023DiffusionBased3H}, camera pose estimation~\cite{Wang2023PoseDiffusionSP}, object detection~\cite{chen2023diffusiondet}, segmentation~\cite{Baranchuk2021LabelEfficientSS, Gu2022DiffusionInstDM}. These developments have been achieved through a generative Markov Chain process based on the Langevin MCMC~\cite{parisi1981correlation} or a reversed diffusion process~\cite{song2020denoising}. Recognizing the power of the diffusion model to iteratively approximate target data distributions from white noise using hierarchical variational decoders, researchers have started applying it to point cloud registration and 6D pose estimation problems. 

The pioneer work~\cite{urain2023se} that applied the diffusion model in the SE(3) space was accomplished by utilizing NCSN~\cite{song2019generative} to learn a denoising score matching function. This function was then used for reverse sampling with Langevin MCMC in SE(3) space to evaluate 6DoF grasp pose generation. Additionally, \cite{jiang2023se} implemented DDPM~\cite{ho2020denoising} in the SE(3) space for 6D pose estimation by employing a surrogate point cloud registration baseline model. Similarly, GeoTR~\cite{qin2022geometric} served as a denoising module in~\cite{chen2023diffusionpcr}, gradually denoising the overlap prior given by the pre-trained model, following a similar approach to PEAL~\cite{yu2023peal}.

\subsection{Image Matching}

% \textbf{Detector-based Methods:} 
Detector-based methods are a crucial research direction in image matching tasks. These approaches focus on identifying keypoints within images and establishing correspondences across different views. Two widely recognized methods in this domain, the Harris Corner Detector~\cite{harris1988combined} and SIFT~\cite{lowe2004distinctive}, are well-known for their speed and robustness. Recently, learning-based methods have gained considerable attention and success. Notably, SuperPoint~\cite{detone2018superpoint}, which jointly trains both the detector and descriptor, has emerged as a prominent technique in this field, with many subsequent studies building upon it. Once keypoints are detected, outlier correspondences must be pruned to ensure accuracy. The pioneering work SuperGlue~\cite{sarlin2020superglue} leverages an attention-based Graph Neural Network (GNN) to learn keypoint features for pruning, achieving highly accurate feature matching. However, this method becomes computationally expensive when dealing with a large number of candidate correspondences. To improve efficiency, subsequent methods such as SGMNet~\cite{chen2021learning}, ClusterGNN~\cite{shi2022clustergnn}, and LightGlue~\cite{lindenberger2023lightglue} attempt to reduce the size of attention scales or introduce sparse matching network designs that can adapt to the matching difficulty. In sparse image matching, DiffGlue~\cite{zhang2024diffglue} introduces the denoised matching matrix attention layer between the self and cross-attention layers of the transformer as prior guidance to improve inlier correspondences pruning. Despite these advancements, the robustness of keypoint detection remains challenging, particularly in texture-less regions, repetitive patterns, and under extreme viewpoint changes.

Detector-free methods, including semi-dense and dense types, aim to directly predict dense matches instead of relying on keypoint detection and matching. The pioneering work LoFTR~\cite{sun2021loftr} utilizes the Transformer to extend the feature’s receptive field and capture long-range dependencies, providing richer contextual information. To further enhance accuracy, Matchformer~\cite{wang2022matchformer} and AspanFormer~\cite{chen2022aspanformer} introduce multi-scale attention modules to improve feature representation. In AspanFormer, a novel optical flow estimation technique is employed to guide local attention. Quadtree attention~\cite{tang2022quadtree} constructs token pyramids and selects top-K patches based on attention scores, reducing computational costs. TopicFM~\cite{giang2023topicfm} categorizes features into topics with similar semantics and constrains attention computation within these topics for greater efficiency. Efficient LoFTR~\cite{wang2024efficient} introduces an aggregate attention module at the coarse level to enhance efficiency.
\section{The proposed Approach}
% \vspace{-0.1cm}
\subsection{Problem Formulation}
\textbf{Point Cloud Registration.} Given source point clouds $\M P \in \mathbb{R}^{N{\times}3}$ and target point clouds $\M Q \in \mathbb{R}^{M{\times}3}$, the 3D registration task is to find top-k correspondences $\mathcal{C}$ from matching matrix $\M E$ and to conduct warping transformation ($\M{\Gamma} \in$ SE(3) for rigid transformations, and 3D flow fields for non-rigid transformations) to align the overlap region of $\M P$ and $\M Q$. 
\begin{align}
    \M \Gamma^* &= \underset{\Gamma \in \text{SE}(3), N_1 \times \mathbb{R}^3}{\min} \sum_{i=1}^{N_1} \sum_{j=1}^{N_2} w_{ij} \, \left\| \M \Gamma(\M x_i) - \M y_j \right\|_2 \notag \\
    & \quad \text{subject to} \quad w \in [0, 1]^{N_1 \times N_2} \notag
\end{align}

\textbf{Image-to-Point Cloud Registration.} In the context of 2D-3D registration, with a source image $\M X \in \mathbb{R}^{H{\times}W{\times}2}$ and target point cloud $\M Y \in \mathbb{R}^{M{\times}3}$, the standard pipeline involves determining the top-k correspondences $\mathcal{C} = \{(\M x_i,\M y_j)|\M x_i \in \mathbb{R}^2, \M y_j \in \mathbb{R}^3\}$, and then estimating the rigid transformation $\M{\Gamma} \in$ SE(3) by minimizing the 2D projection error:
\begin{equation}
\nonumber
    \M{\Gamma}^* = \underset{\M{\Gamma} \in \text{SE}(3)}{\min} \sum \limits_{\M x_i,\M y_j\in \mathcal{C}} w_{ij}||\text{Proj}(\M{\Gamma}(\M y_j),\M K)-\M x_i||_2
\end{equation}
where $\M K$ represents the camera intrinsic matrix, and $\text{Proj}(\cdot, \cdot)$ denotes the projection function from 3D space to the image plane. $w_{ij}$ denote the correspondence weight. In this paper, we utilize a pre-trained depth estimation model to give the depth map of image $x_i$ and convert the 2D-3D registration to 3D registration.

\textbf{Image Matching.} Given source image $\M{X} \in {\mathbb{R}}^{H \times W \times 2}$ and the target image $\M{Y} \in {\mathbb{R}}^{H' \times W' \times 2}$, the goal of image registration is to find reliable correspondences between them.

\subsection{Preliminaries}

\vspace{1mm}
\subsubsection{Doubly Stochastic Matrix}
We can represent the point clouds $P$ and $Q$ as two graphs, denoted as $\mathcal{G}_1 = \left\{P, E^P\right\}$ and $\mathcal{G}_2 = \left\{Q, E^Q\right\}$, where $E^P$ and $E^Q$ are respective edge sets. The matching matrix between these two graphs is a one-to-one mapping $E \in \{0,1\}^{N{\times}M}$. In cases where $N \neq M (e.g., N>M)$, we can introduce $N-M$ dummy points in $Q$ to make a square matching matrix, also known as a permutation matrix $\mathcal{M} = \left\{A: A1_N = 1_N, A^{T}1_N=1_N, A\geq 0\right\}$.
Then, we further employ sinkhorn iterations\cite{cuturi2013sinkhorn} to convert this non-negative real matrix into a ``doubly stochastic'' matrix, which has uniform row sum $M$ and column sum $N$~\cite{86bd1ad6-50bb-38d0-9978-0966b4dfc6d3}. 

\begin{figure*}
      % \vspace{-0.5cm}
      \centering
      \includegraphics[width=15cm, height=7cm]{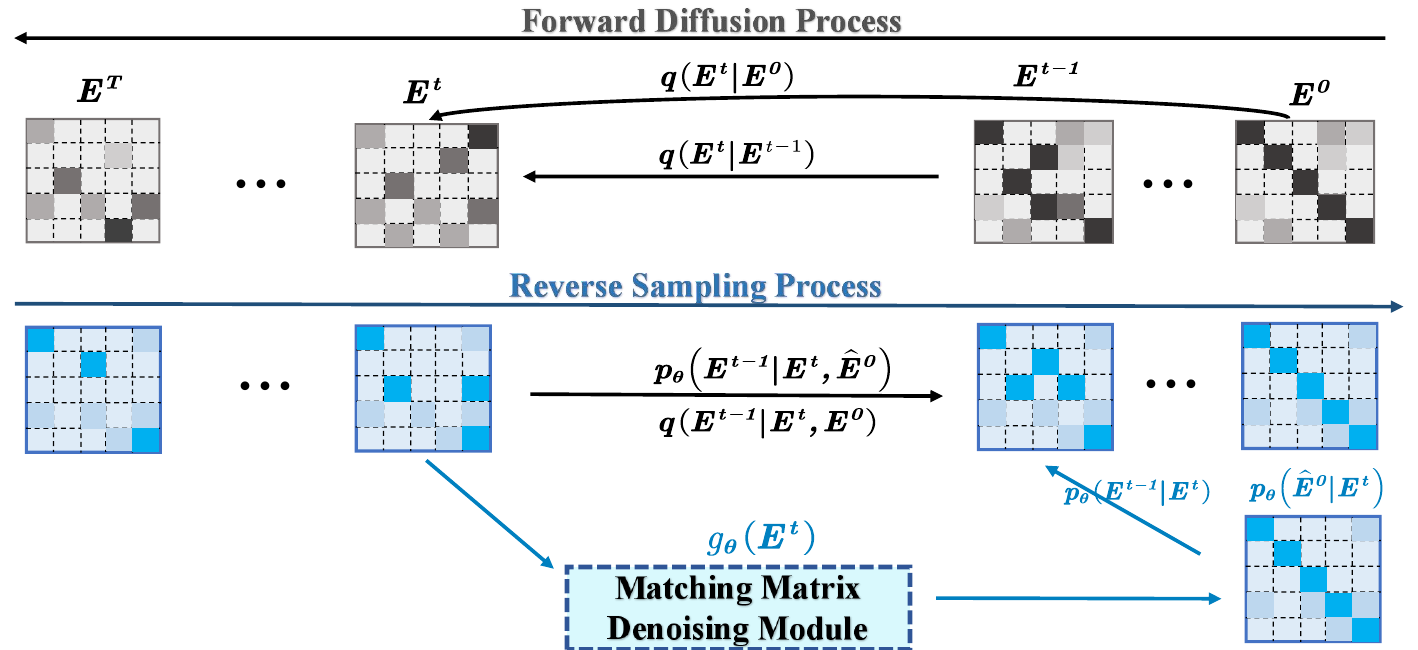}
    
\caption{Illustration of our diffusion matching model for matching matrix estimation. The forward diffusion process is driven by the Gaussian transition kernel $q(E^t|E^{t-1})$, which has a closed form $q(E^t|E^0)$. The denoising model $g_\theta(E^t)$ learns a reverse denoising gradient that points to the target solution $E^0$. During inference in the reverse sampling process, we utilize the predicted $\hat{E}_0$ and DDIM(\cite{song2020denoising}) to sample $E^{t-1}$ from the current state $E^t$.} \label{framework_of_matching_matrix_ddpm}    
\vspace{-0.4cm}
\end{figure*}

\subsubsection{Diffusion Model} According to \cite{song2020denoising,song2020score}, the forward diffusion process obeys a forward stochastic differential equation (SDE) starting from $\M{x}_0 \sim q(x_0)$:
\begin{equation}
d\M{x}_t\ = \ f(\M{x}_t,t)\M{x}_tdt + g(t)d\M{w}_t 
 ,\label{ddpm_diffusion_forward}
\end{equation}
where t $\in [0,T]$, f and g are the drift and diffusion terms, respectively, and $\M{w}$ is a standard Wiener process. To sample the $\M{x}_0$, we can solve the reverse process and  probability flow ODE\cite{song2020score}:
\begin{eqnarray}
\begin{aligned}
    d\M{x}_t\ &=\ \left[f(\M{x}_t,t) - g(t)^2\nabla _{\M{x}_t}\log q(\M{x}_t)\right]dt + g(t)d\M{w}_t,\\
    d\M{x}_t\ &=\ \left[f(\M{x}_t,t) - \frac{1}{2}g(t)^2\nabla _{\M{x}_t}\log p(\M{x}_t)\right]dt,\label{ddpm_diffusion_reverse}
\end{aligned}    
\end{eqnarray}
where $p(\M{x}_t)$ is the marginal distribution at timestamp $t$.
The discrete reverse sampling process is defined as follows:
\begin{equation}
\begin{split}
\M{x}_{t_{k+1}}=\M{x}_{t_{k}}&+\left[f(\M{x}_{t_{k}},t_{k})-g(t_{k})^{2}\nabla_{\M{x}_{t_{k}}} \log q(\M{x}_{t_{k}})\right]\Delta t\\
&+g(t_{k})\sqrt{\Delta t}\cdot z_{k},    
\end{split}\label{dis_reverse_sm}
\end{equation}

The discrete DDPM~\cite{ho2020denoising,SohlDickstein2015DeepUL,song2020score} form of the reverse sampling process (equivalent to Eqn.\eqref{dis_reverse_sm}) is:
\begin{equation}
    \mathbf{x}_{t-1} = \frac{1}{\sqrt{\alpha_t}} \left( \mathbf{x}_t - \frac{1 - \alpha_t}{\sqrt{1 - \bar{\alpha}_t}} \boldsymbol{\epsilon}_\theta(\mathbf{x}_t, t) \right) + \sigma_t \mathbf{z}, \quad \mathbf{z} \sim \mathcal{N}(0, \mathbf{I}),\label{dis_reverse}
\end{equation}
where \( \alpha_t \in (0, 1) \) denotes the noise schedule at timestep \( t \), and \( \bar{\alpha}_t = \prod_{s=1}^t \alpha_s \). Given the model prediction \( \hat{\mathbf{x}}_0 = \mathbf{x}_0^\theta(\mathbf{x}_t, t) \), the mean of the reverse process is derived as:
\begin{equation}
\mu_\theta(\mathbf{x}_t, t) = \frac{1}{\sqrt{\alpha_t}} \left( \mathbf{x}_t - \frac{1 - \alpha_t}{\sqrt{1 - \bar{\alpha}_t}} (\mathbf{x}_t - \sqrt{\bar{\alpha}_t} \hat{\mathbf{x}}_0) \right),\label{rev_x0_u}
\end{equation}
and the reverse process is sampled by:
\begin{equation}
\mathbf{x}_{t-1} = \mu_\theta(\mathbf{x}_t, t) + \sigma_t \mathbf{z}, \quad \mathbf{z} \sim \mathcal{N}(0, \mathbf{I}),\label{rev_x0}
\end{equation}
where \( \sigma_t \) can be a fixed predefined variance or a learned one. When \( \sigma_t = 0 \), this recovers a deterministic process as in DDIM\cite{song2020denoising}.

\subsection{Overview}
In this section, we provide an overview of the following sections. First, we introduce two theorems to explain and support the motivation for employing the diffusion model to estimate the matching matrix (see Section~\ref{motivate_theo}). Second, we provide a high-level description of the diffusion model for the matching matrix, including the forward diffusion and reverse sampling processes, as well as the derivation of the training objective (see Section~\ref{matrix_diffusion_model}). Third, in Section~\ref{mme}, we describe how to encode the matching matrix using a "match-to-warp" embedding ${\M F^{\hat{P}}_{E^t}}$, given point features and positional encoding. We also provide an adaptive implementation of the "match-to-warp" embedding for both 3D and 2D cases. Based on this matching matrix encoding, we develop the denoising module $g_{\theta}$ (see Section~\ref{denoising_module}), which is used in the reverse sampling process. Then, in Section~\ref{framework_training_inference}, we describe the training framework for three types of registration tasks. Finally, we present the experimental details and results in Section~\ref{exp_section} to validate our approach.

\subsection{Theoretical Foundations}\label{motivate_theo}
% In this section, we give theoretical support for using the diffusion model in matrix space to estimate correspondences.

% As described in the introduction, there exists a surjective relation between the matching matrix space and the warping operation space (e.g., SE(3), scene flow fields $\mathbb{R}^{3N}$, and optical flow fields $\mathbb{R}^{2N}$). A diffusion model operating in the matching matrix space can sample training data at a denser resolution with greater diversity. This implies that optimizing the registration problem in the matching matrix space can yield a tighter (i.e., lower) lower bound on the residual error of the registration objective than when optimizing in the lower-dimensional warping operation space:

In this section, we provide theoretical support for using the diffusion model in matrix space to estimate correspondences.

As described in the introduction, there exists a surjective relationship between the matching matrix space and the warping operation space (e.g., SE(3), scene flow fields $\mathbb{R}^{3N}$, and optical flow fields $\mathbb{R}^{2N}$). A diffusion model operating in the matching matrix space can sample training data at a higher resolution and with greater diversity. This suggests that optimizing the registration problem in the matching matrix space can yield a tighter (i.e., lower) bound on the residual error of the registration objective than optimizing in the lower-dimensional warping operation space.

\textbf{Theorem 1.} Let $W^E$ be the warping function generated by matching matrix $E$, and let $\hat{E}^W$ be the matching matrix computed from the warping $W$. The following inequality holds:
\begin{align}
    \underset{E \in \mathcal{M}}{\min} \sum_{i=1}^{N} \sum_{j=1}^{M} E_{ij} \cdot \left\| W^E(p_i) - q_j \right\|_2 &\leq \notag \\
    \mathop{\min}\limits_{W \in  SE(3)/\mathbb{R}^{3N}/ \mathbb{R}^{2N} } \sum_{i=1}^{N} \sum_{j=1}^{M} \hat{E}^W_{ij} \cdot \left\| W(p_i) - q_j \right\|_2.
\end{align}

\noindent\textbf{Proof:} Please refer to the supplementary materials for detailed proof.

We also provide another theoretical support for why we utilize the reverse denoising sampling process to search for the matching matrix solution:

\textbf{Theorem 2.} The matching matrix estimation problem:
\begin{equation}
    E^* = \underset{E \in \mathcal{M}}{\min} \sum_{i=1}^{N} \sum_{j=1}^{M} E_{ij} \cdot \left\| W^E(p_i) - q_j \right\|_2
\end{equation}
can be solved by a sequence of optimal transport sub-problems:
\begin{equation}
E^{n+1} = \arg\min\limits_{E \in \Pi(\M p,\M q)} \langle \tilde{\M{C}}^n(E^n), E \rangle+ \epsilon H(E),\label{convexot}
\end{equation}
where $\tilde{\M{C}}^n$($E^n$) is a similarity matrix based on the point features and $E^n$.

\textbf{Proof:} Please refer to the supplementary materials for detailed proof.

Theorem 2 provides an important insight: the optimal solution can be obtained by searching along a trajectory or path. Ideally, this trajectory should be guided by an optimal gradient, which can be learned through a neural network. Fortunately, the reverse diffusion sampling process can be interpreted as a gradient-guided stochastic optimization trajectory in a high-dimensional probability space, where the guiding gradient is learned from the diffusion samples.

Building on Theorem 2 and the reverse sampling processes defined in equations~\eqref{dis_reverse_sm} and~\eqref{dis_reverse}, we implement an iterative update scheme to solve for the matching matrix. At each iteration, we apply the Sinkhorn algorithm to project the intermediate result into the space of doubly stochastic matrices. To ensure the convergence of the reverse sampling process to a fixed-point solution, we adopt the DDIM framework~\cite{song2020denoising}, which provides a deterministic mapping and yields a stable prediction of the matching matrix.

\subsection{Diffusion Model for Matching Matrix}\label{matrix_diffusion_model}
In this section, we introduce the construction of our diffusion matching model for generating the matching matrix between two scans (refer to Fig.~\ref{framework_of_matching_matrix_ddpm}). We denote the matching matrix as $E\in \{0,1\}^{N{\times}M}$, and we assume $E$ is defined in a nonsquare ``doubly stochastic'' matrix space $\mathcal{M}$.

\vspace{1mm}
\noindent\textbf{Forward Diffusion Process.}
Following DDPM\cite{ho2020denoising}, we can fix the forward diffusion process to a Markovian chain, denoted as $q(E^{1:T}|E^0)$, which generates a sequence of latent variables $E^t$ by gradually injecting Gaussian noise into the ground truth matching matrix $E^0$. The diffused matching matrix $E^t$ at arbitrary timestep \textit{t} has an analytic closed form:
\begin{eqnarray}
    E^t \sim q(E^t|E^0) = N(E^t;\sqrt{\bar{\alpha}}E^0,(1-\bar{\alpha})\textbf{I}))\label{diff_10}.
\end{eqnarray}
where the added noise over each element of the matrix is sampled independently and identically distributed (i.i.d.). To improve performance, we modified the formulation \eqref{diff_10} to the following:
\begin{eqnarray}
\begin{aligned}
% \nonumber
\rm{(Rigid)}\quad \M E^t &= \sqrt{\bar{\alpha_t}}\M E^0 + \sqrt{1-\bar{\alpha_t}}f_{\epsilon}(\epsilon_0), \\
% \tilde{\M E}^t &= \M E^t - \rm{Min}(\M E^t), \\
\rm{(Deformable)}\quad \M E^t &= \sqrt{\bar{\alpha_t}}\M E^0 + \sqrt{1-\bar{\alpha_t}}\epsilon_0.  \\ 
\end{aligned}
\end{eqnarray}
where $f_{\epsilon}=(\epsilon\%1)(abs(\epsilon)/\epsilon)\eta$. We empirically set $\eta=1.5$ and $\epsilon_0 \sim \mathcal{N}(\epsilon;0,\textbf{I})$.

% This diffused $E^t \sim q(E^t|E^0)$ is a continuous matrix in $R^{N \times M}$, which is outside the feasible solution space of matching matrices (i.e., doubly stochastic matrix manifolds). 

To constrain the diffused $E^t \sim q(E^t|E^0)$ to close to the discrete matching matrix sets, we apply the following projection to confine the matrix $E^t$ to the feasible solution space.
For 3D registration and 2D registration, we utilize the Sinkhorn and Softmax projections, respectively:
\begin{eqnarray}
% \begin{aligned}
    \M E^t = \mathbf{f}_{\rm{sinkhorn}}^{\rm{3d}}(\M E^t), \quad
    \M E^t = \mathbf{f}_{\rm{softmax}}^{\rm{2d}}(\M E^t)
% \end{aligned}    
\end{eqnarray}
where the $\mathbf{f}_{\rm{sinkhorn}}^{\rm{3d}}$ operation is from the Sinkhorn algorithm~\cite{cuturi2013sinkhorn} and $\mathbf{f}_{\rm{softmax}}^{\rm{2d}}$ is a softmax operation.

\vspace{1mm}
\noindent\textbf{Reverse Denoising Sampling Process.}
Given a diffusion Markovian chain $E^0 \rightarrow E^1 \rightarrow ... \rightarrow E^T$, we need to learn a reverse transition kernel with the posterior distribution $q(E^{t-1}|E^t, E^0)$ to sample the reverse Markovian chain $E^T \rightarrow E^{T-1} \rightarrow ... \rightarrow E^0$ from a white noise $E^T$ to achieve the target matching matrix. The posterior distribution $q(E^{t-1}|E^t,E^0)$ conditioned on $E^0$ and $E^t$ is defined as: 
\begin{eqnarray}
\begin{aligned}
    &q(E^{t-1}|E^t,E^0) = \frac{q(E^t|E^{t-1},E^0)q(E^{t-1}|E^0)}{q(E^t|E^0)} \\
    &\propto N(E^{t-1};\underbrace{\frac{\sqrt{\alpha_t}(1-\bar{\alpha}_{t-1})E^t+\sqrt{\bar{\alpha}_{t-1}}(1-\alpha_{t})E^0}{1-\Bar{\alpha}_t}}_{\mu_q(E^t,E^0)},\\
    &\underbrace{\frac{(1-\alpha_t)(1-\bar{\alpha}_{t-1})}{1-\bar{\alpha}_t}\textbf{I}}_{\Sigma_q(t)} ).
\end{aligned}
\end{eqnarray}
\label{eqn_rds_matrix}

\noindent\textbf{Training objective.} For latent variable $E^{1:T}$, the Evidence Lower Bound (ELBO) for $E^0$ with distribution $q$ formulates as:
\begin{eqnarray}
% \nonumber
\begin{aligned}
&\log(p_{\theta}(E^0)) \geq L_{vb}(E^0)\\ &=\mathbb{E}_{q(E^{1:T}|E^0)}\left[ \log\left( \frac{p_{\theta}(E^{0:T})}{q(E^{1:T}|E^0)} \right) \right] \\
% &=\mathbb{E}_{q(E^1|E^0)}\left[ \log(p_{\theta}(E^0|E^1))\right] + \mathbb{E}_{q(E^{T}|E^0)}\left[ \log(\frac{p_{\theta}(E^T)}{q(E^T|E^0)})\right]\\
% &+\sum_{t=2}^{T}\mathbb{E}_{q(E^{t},E^{t-1}|E^0)}\left[\frac{p_{\theta}(E^{t-1}|E^{t})}{q(E^{t-1}|E^{t},E^0)}\right]\\
&\approx -\sum_{t=2}^{T}\underbrace{ \mathbb{E}_{q(E^{t}|E^0)}\left[KL\left(q(E^{t-1}|E^{t},E^0) || p_{\theta}(E^{t-1}|E^{t})\right)\right]}_{\text{denoising matching term}}.\label{elbo}
\end{aligned}
\end{eqnarray}

Based on equations~\eqref{rev_x0_u},\eqref{rev_x0}, and\eqref{elbo}, we propose a lightweight denoising network $g_{\theta}(E^t)$ that seeks to predict $E^0$ from any noisy input $E^t$, and derive a simplified version of the loss to train $p_\theta(E^0|E^t)$:
\begin{eqnarray}
\begin{aligned}
    &L_{simple} =-\mathbb{E}_{q(E^0)}\left[\sum \limits_{t-1}^T\mathbb{E}_{q(E^t|E^0)}logp_\theta(E^0|E^t)\right]\label{L_simple}.
\end{aligned}    
\end{eqnarray}
The details of the  of $L_{simple}$ can be found in the supplementary material.

\begin{figure*}
      % \vspace{-0.5cm}
      \centering
      \includegraphics[width=\textwidth, height=9cm]{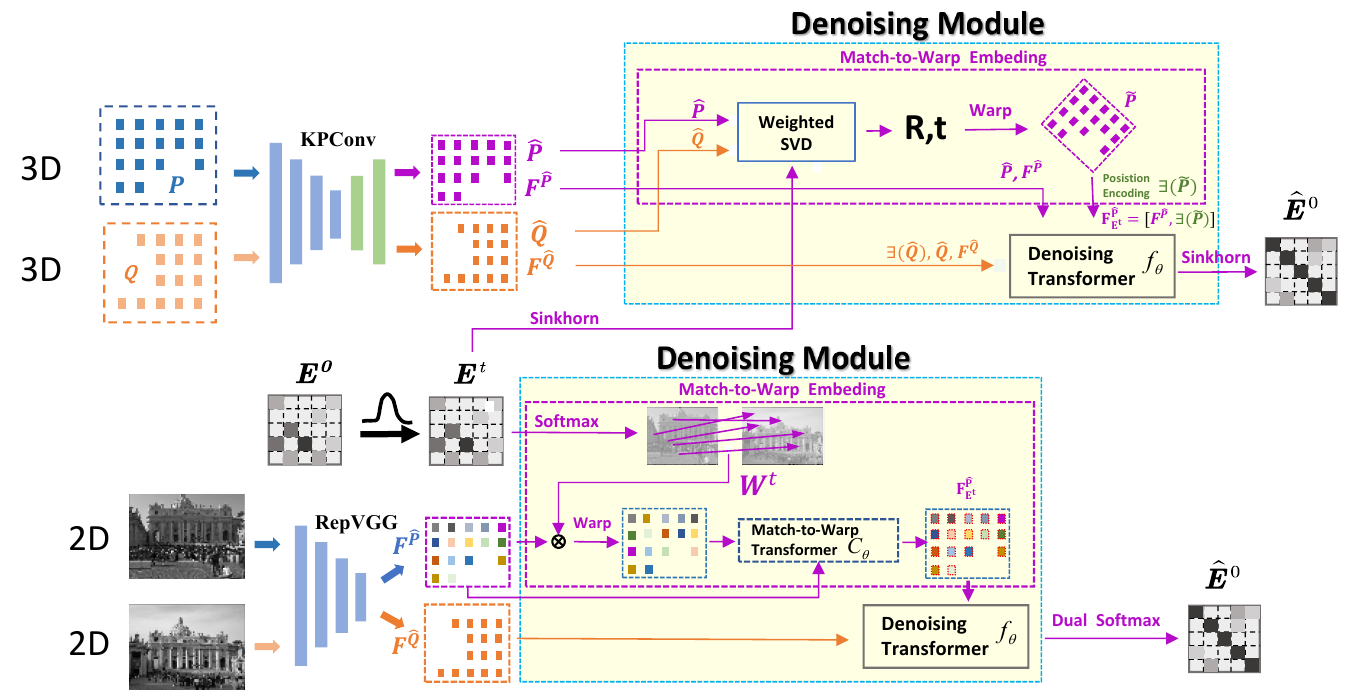}
      
\caption{Detail of our denoising module. For both 2D image matching and 3D registration, we consistently implement the denoising module to accommodate the distinct characteristics of each task. The matching matrix $E^t$ is embedded into a new feature, denoted as $F^{\hat{P}}_{E^t}$, along the coordinates in the source frame $\hat{P}$. (refer to section \ref{mme}).} \label{framework_2}    
\vspace{-0.4cm}
\end{figure*}

\subsection{\textit{“Matching-to-Warp”} Matching Matrix Encoding}\label{mme}

In this section, we describe the details of how to embed the matching matrix. The motivation is to encode the matching matrix with geometric information derived from the warped source points or pixels.

In 3D registration, we treat each element of the matching matrix (e.g., $E$) as a correspondence weight in the weighted SVD algorithm~\cite{arun1987least} for computing the 6D pose $\M T$. After applying the transformation $\M T$ to the source point cloud ${\hat{P}}$, we obtain a new warped point cloud ${\tilde{P}}$. The new positional encoding $\exists ({\tilde{P}})$ represents the geometric encoding of matrix $E$ on the source point cloud ${\hat{P}}$.

In 2D registration, it is difficult to directly obtain geometric encodings of the warped source image because 2D pixel coordinates lie in a projected coordinate system. The warping generated by the diffused matching matrix $E^t$ may result in invalid mappings of source image pixels, causing the warped image to become disordered and meaningless. To address this, we convert the matching matrix $E^t$ into a flow field $\M{F^t}$, which maps the source pixel coordinates into the valid region of the target image's coordinate system. The positional encoding of these warped source pixel coordinates serves as the geometric encoding of the matching matrix $E^t$.

In practice, after Sinkhorn projection with a limited number of iterations, the elements in each column may not sum to 1.0. This is acceptable in 3D registration, as the sorted matrix elements produced by the Sinkhorn iterations are sufficient for the weighted SVD algorithm. However, in 2D image matching, where the matrix must be transformed into a flow field within the image coordinate system, it is necessary for the elements in each column to sum to 1.0. Therefore, we apply a softmax operation to the matching matrix $E^t$ in the 2D image matching case.

\subsubsection{Matching Matrix Embeddding in 3D Registration}
In 3D registration, we take the downsampled superpoints $\hat{P}$ and $\hat{Q}$ along with the points features $F^{\hat{P}}$ and $F^{\hat{Q}}$ as the inputs.
Similarly, in 2D-3D registration, the downsampled superpixels $\hat{X}$ and point cloud $\hat{Q}$ with associated inputs $\{F^{\hat{X}}, D^{\hat{X}}, F^{\hat{X}}_{dino}, F^{\hat{Q}}\}$ are taken. 

For the 3D point cloud registration, matching-to-Warp embedding module take the noisy matching matrix $E^t$ as input and outputs the geometric embedding $\M {F}^{\hat{P}}_{E^t}=[F^{\hat{P}}, \exists({\tilde{P}})]$ (see the top in Fig. \ref{framework_2}). The details of the algorithm are as follows:
\begin{eqnarray}
\begin{aligned}
&\tilde{E}_t = \text{F.sinkhorn}(E^t),\\
&\M{\hat{R}}_t, \M{\hat{t}}_t = \M{soft\_procrustes}(\tilde{E}_t, \hat{ P}, \hat{ Q}), \\
&\tilde{P} =\M{\hat{R}}_t\hat{P}+\M{\hat{t}}_t,\\
&{\M F^{\hat{P}}_{E^t}}=[F^{\hat{P}}, \exists(\tilde{P})]
\end{aligned}\label{3d_mtw}
\end{eqnarray}
where $\exists$ is the rotationary positional embedding\cite{li2022lepard} and F.sinkhorn is the sinkhorn algorithm\cite{cuturi2013sinkhorn}.

For the 2D-3D image-to-point cloud registration task, we take the point cloud $\hat{Q}$ as the source point cloud and carry out matching-to-warp embedding on it.

\subsubsection{Matching Matrix Embedding in 2D Image Matching}
In this section, we describe the geometric embedding of the matching matrix ${E^t}$ in 2D registration. First, we project the matching matrix $E^t$ using the softmax function, transforming it into a flow field $\M F^t_{\hat{Q} \rightarrow \hat{P}}$. Next, the source feature map $\M F^{\hat{P}}$ is warped into $\M F^{\tilde{Q}}$, which is in the interior of the target image frame. The source pixels $\hat{P}$ are warped to $\tilde{Q}$ (in the target pixel coordinates). We then exploit a transformer $C_{\theta}$ to transfer the positional information of $\tilde{Q}$ back to $\hat{P}$'s pixel coordinates, thereby obtaining the geometric embedding $\M F^{\hat{P}}_{E^t}$ (refer to the bottom in Fig. \ref{framework_2}) of the matching matrix $E^t$. The details of the algorithm are as follows:
\begin{eqnarray}
\begin{aligned}
&\M W^t = \text{F.softmax}(E^t).T \\
&\M F^t_{\hat{Q}\rightarrow\hat{P}} = \M W^t \text{Grid}^{H \times W}_{\hat{P}} - \text{Grid}^{H \times W}_{\hat{Q}}\\
&\text{Sample\_Grid}_{\hat{P}}^{E^t} = \M F^t_{\hat{Q}\rightarrow\hat{P}} +  \text{Grid}^{H \times W}_{\hat{Q}}\\
&\M F^{\tilde{Q}}_{E^t} = \text{F.grid\_sample}(\M F^{\hat{P}}, \text{Sample\_Grid}_{\hat{P}}^{E^t})\\
&\M F^{\hat{P}}_{E^t} = C_{\theta}(\M F^{\hat{P}},F^{\tilde{Q}}_{E^t})
\end{aligned}\label{2d_mtw}
\end{eqnarray}
% For robustness and effectiveness, we exploit a transformer $C_{\theta}$ to convert the warped position encoding $\exists(\tilde{Q})$ back to the superpixel space $\hat{P}$' instead of directly converting by hard coding.
% Compared to the explicit embedding ${\M F^{\hat{P}}_{E^t}}$, including point feature $F^{\hat{P}}$ and position information $\exists(\tilde{P})$ in Eqn.\eqref{3d_mtw}, we obtained the implicit embedding ${\M F^{\hat{P}}_{E^t}}$ in Eqn.\eqref{2d_mtw} by the transformer $C_{\theta}$, in which we transfer the warped position encoding $\exists(\tilde{Q})$ into the pixel coordinates of $\hat{P}$. The $\tilde{Q}$ in Eqn.\eqref{2d_mtw} is equivalent or similar to the $\tilde{P}$ in Eqn.\eqref{3d_mtw}. The warped position encoding $\exists(\tilde{Q})$ is embedded in transformer  $C_{\theta}(\M F^{\hat{P}},F^{\tilde{Q}}_{E^t})$. We utilize the aggregated Attention Module in the transformer of Efficient LoFTR~\cite{wang2024efficient} as the definition of attention in $C_{\theta}$. 
Compared to the explicit embedding ${\M F^{\hat{P}}_{E^t}}$, which includes the point feature $F^{\hat{P}}$ and positional information $\exists(\tilde{P})$ in Eqn.~\eqref{3d_mtw}, we obtain the implicit embedding ${\M F^{\hat{P}}_{E^t}}$ in Eqn.\eqref{2d_mtw} via the transformer $C_{\theta}$, where the warped positional encoding $\exists(\tilde{Q})$ is transferred into the pixel coordinates of $\hat{P}$. The $\tilde{Q}$ in Eqn.\eqref{2d_mtw} is equivalent or similar to $\tilde{P}$ in Eqn.~\eqref{3d_mtw}. The warped positional encoding $\exists(\tilde{Q})$ is embedded through the transformer $C_{\theta}(\M F^{\hat{P}}, F^{\tilde{Q}}_{E^t})$. We adopt the aggregated attention module in the transformer of Efficient LoFTR~\cite{wang2024efficient} as the definition of attention in $C_{\theta}$.

The reason we use a weaker projection (e.g., softmax) for image matching instead of the Sinkhorn projection used in 3D registration is that the warping of square pixels may extend beyond the coordinate range of the target image. This can create difficulties when converting the warped positional encoding $\exists(\tilde{Q})$ back to the superpixel space of $\hat{P}$. To address this, we apply softmax to ensure that each pixel in the source image has at least one corresponding pixel in the target image. However, due to low-overlap regions caused by viewpoint changes, some warped pixels may be incorrectly positioned. To correct this, we employ the learnable transformer $C_{\theta}$ to select the correct warped pixels and reassign their positional encodings to the corresponding pixels in $\hat{P}$, instead of relying on hard-coded mappings, to improve robustness.

\subsection{Denoising Module \texorpdfstring{$g_{\theta}$}{g\_theta}}\label{denoising_module}
In this section, we describe the denoising module $g_\theta$ for point cloud registration and image matching. Given the matching matrix geometric embeddings $\M F^{\hat{P}}{E^t}$, the denoised matching matrix is obtained by introducing a denoising module $g_{\theta}$, which utilizes the warped source embeddings $\M F^{\hat{P}}_{E^t}$, target embeddings $\M F^{\hat{Q}}$, superpoints (or superpixels), and other associated features as inputs. $g_{\theta}$ primarily consists of Matching Matrix Embedding (refer to~\ref{mme}), Denoising Transformer~\cite{vaswani2017attention}, Matching Function, and Projection Function.

\vspace{1mm}
\noindent\textbf{{Denoising Transformer $f_{\theta}$}}: 
We exploit a lightweight Transformer as our denoising network. Specifically, we utilize a 6-layer interleaved attention layers transformer $f_\theta$ for denoising feature embedding.

\vspace{1mm}
\noindent\emph{Attention Layer.} In 3D registration task, we follow the transformer implementation from \cite{li2022lepard,wang2024efficient}, the $q,k,v$ in the self-attention are computed as:
\begin{equation}
\begin{aligned}
&q_i = \exists(p_i)\mathbf{W}_qf^{\hat{p}_i}, \
k_j = \exists(p_j)\mathbf{W}_kf^{\hat{p}_j}, \
v_j = \mathbf{W}_vf^{\hat{p}_j}, \\
&f^{\hat{p}_i} = f^{\hat{p}_i} + \text{MLP}(\text{cat}[f^{\hat{p}_i},\Sigma_j\alpha_{ij}v_j]),
\end{aligned}\label{att_12}
\end{equation} where $\mathbf{W}_q,\mathbf{W}_k,\mathbf{W}_v \in {\mathbb{R}}^{d{\times}d}$ are the attention weights, ${\alpha}_{ij} = softmax(q_{i}k_{j}^T/{\sqrt{d}})$, and $\exists(\cdot)$ is a relative rotationary position encoding~\cite{li2022lepard}. MLP$(\cdot)$ is a 3-layer fully connected network, and $cat[{\cdot},\cdot]$ is the concatenating operator. The cross-attention layer is the standard form in which $q$ and $k,v$ are computed by source and target point clouds, respectively. Other operations are the same as self-attention.
In the 2D-3D registration task, we take image inputs $\{\hat{X},F^{\hat{X}},F^{\hat{X}}_{dino}\}$ and point cloud input $\{\hat{Q},F^{\hat{Q}}\}$ to compute $q,k,v$ by utilizing standard attention layers\cite{vaswani2017attention}. We also take the Fourier embedding function\cite{mildenhall2021nerf} to embed superpixels $\hat{X}$ and super points $\hat{Q}$ as positional encoding. 

In the 2D image matching task, we adopt the self-attention and cross-attention from Efficient LoFTR~\cite{wang2024efficient}, i.e., the aggregation attention: 
\begin{eqnarray} f^{\hat{p}_i} = \text{Conv2D}(f^{\hat{p}_i}), \quad f^{\hat{p}_j} = \text{MaxPool}(f^{\hat{p}_j}),
\end{eqnarray} 
where Conv2D is implemented using a stride depthwise convolution with a kernel size of $s \times s$, matching that of the max-pooling layer. The position encoding employs relative position encoding~\cite{su2024roformer} to enhance robustness.

\vspace{1mm}
\noindent\textbf{Matching Function}: $\mathbf{matching\_logits}(\cdot,\cdot,\cdot,\cdot)$. \label{mlogits}
We compute matching ``logits''between $\hat{P}$ and $\hat{Q}$ by features $F^{\hat{P}}$ (or $F^{\hat{X}}$) and $F^{\hat{Q}}$: $\tilde{E}(i,j) = \frac{1}{\sqrt{d}}\left<f^{\hat{p}_i}, f^{\hat{q}_j}\right>$.

\vspace{1mm}
\noindent\textbf{Projection Function}: 
% In 2D image matching, we transform the matching matrix into flow fields, from which the generated warped source image pixels must ensure that the overlap region is within the target image's coordinate system. To achieve this, we exploit softmax projection to convert the matching matrix to flow fields. In 3D registration, the elements of the matching matrix are treated as correspondence weights in the weighted SVD for 6D pose estimation. Only the numerical sorting relationship between the matrix elements is required. Therefore, we can impose stricter constraints on the matching matrix by using the Sinkhorn algorithm~\cite{cuturi2013sinkhorn} to project the matching matrix into the space of doubly stochastic matrices.
% For the sake of clarity, we provide pseudo-code in Algorithm \ref{denoising_algorithm_2D} to describe the details of the denoising module $g_{\theta}$. The implementation of \textit{Matching-to-Warp} step is in the Eqn.(\ref{3d_mtw}) and Eqn.(\ref{2d_mtw}), respectively. 
In 2D image matching, we transform the matching matrix into flow fields, where the resulting warped source image pixels must ensure that the overlapping region lies within the target image's coordinate system. To achieve this, we employ softmax projection to convert the matching matrix into flow fields.

In 3D registration, the elements of the matching matrix are treated as correspondence weights in the weighted SVD used for 6D pose estimation. Only the relative ranking of the matrix elements is required. Therefore, we impose stricter constraints on the matching matrix by applying the Sinkhorn algorithm~\cite{cuturi2013sinkhorn} to project it into the space of doubly stochastic matrices.

For clarity, we provide pseudo-code in Algorithm~\ref{denoising_algorithm_2D} to describe the details of the denoising module $g_{\theta}$. The implementation of the \textit{Matching-to-Warp} step is shown in Eqn.\eqref{3d_mtw} and\eqref{2d_mtw}, respectively.

\begin{algorithm}
\caption{Denoising Module $g_\theta$'s definition.}
\label{denoising_algorithm_2D}
\begin{algorithmic}[1]
\Require Sampled matching matrix $\M E^t \in \mathbb{R}^{N\times M}$; Point clouds pair $\M {\hat{P}}, \M {\hat{Q}} \in \mathbb{R}^{3}$ (or or image pair $\M {\hat{P}}, \M {\hat{Q}} \in \mathbb{R}^{2}$) and associated point/pixel features $\M F^{\hat{\M P}}, \M F^{\hat{\M Q}}$.
\Ensure Target matching matrix $\hat{\M E}_0$.
\Function{$g_{\theta}$}{$\M E^t, \hat{\M P}, \hat{\M Q}, \M F^{\hat{\M P}}, \M F^{\hat{\M Q}}$}
      
        \State $\M F^{\hat{P}}_{E^t} \gets \text{Matching-to-Warp}(\M E_t, \hat{\M P},\hat{\M Q}, \M F^{\hat{P}})$
        \State $\tilde{\M F}^{\hat{\M P}_t}, \tilde{\M F}^{\hat{\M Q}_t} \gets f_{\theta}(\hat{\M P}_t, \hat{\M Q}, \M F^{\hat{P}}_{E^t}, \M F^{\hat{\M Q}})$
        \State $\tilde{\M E}_0 \gets \M{matching\_logits}(\tilde{\M F}^{\hat{\M P}_t}, \tilde{\M F}^{\hat{\M Q}_t})$
        \If{Image Matching}
         \State $\hat{\M E}_0 \gets \text{F.softmax}(\tilde{\M E}_0)$
        \Else
         \State $\hat{\M E}_0 \gets \text{F.sinkhorn}(\tilde{\M E}_0)$        
        \EndIf
        \State \textbf{return} $\hat{\M E}_0$
\EndFunction
\end{algorithmic}
\end{algorithm}

\subsection{Framework Implemantation Details}\label{framework_training_inference}
We aim to design the matching matrix denoising module $g_\theta$ as a nearly "plug-and-play" component, which can be trained in parallel with transformer-based methods for registration tasks. To the best of our knowledge, we empirically believe that our denoising module $g_\theta$ can be integrated with most semi-dense or dense image matching methods, as well as point cloud registration methods—an avenue we plan to explore in future work.

We train the denoising module by integrating it with three state-of-the-art registration methods: Efficient LoFTR~\cite{wang2024efficient}, Lepard~\cite{li2022lepard}, and 2D-3D-MATR~\cite{li20232D3D}, in order to evaluate it on image matching, point cloud registration, and image-to-point cloud registration tasks. More details about the framework design can be found in the supplementary materials.

\section{Experiments}\label{exp_section}
In this section, we conduct experiments on 2D image matching, 3D point cloud registration, and 2D-3D image-to-point cloud registration tasks to prove the effectiveness of our method. Specifically, we test our diffusion matching model on the 3DMatch\cite{zeng20173dmatch}, 4DMatch\cite{li2022lepard}, RGB-D SCENES V2~\cite{lai2014unsupervised}, 7Scenes~\cite{shotton2013scene}, MegaDepth~\cite{li2018megadepth}, and Scannet~\cite{dai2017Scannet} benchmarks.

\subsection{2D-2D Image Matching}\label{design_2d2d_task}
% \subsubsection{\noindent\textbf{Implementation Details.}}
Our model is trained on MegaDepth and Scannet datasets. Following LoFTR~\cite{sun2021loftr}, in MegaDepth, the test split is sampled from the training split. We use the AdamW optimizer with an initial learning rate of $8 \times 10^{-3}$. We trained and evaluated models on 4 NVIDIA V100 GPUs. In this section, we conduct relative pose estimation to evaluate our method's effectiveness.

\vspace{1mm}
\noindent\textbf{Datasets.}
Following Efficient LoFTR~\cite{wang2024efficient} and AspanFormer~\cite{chen2022aspanformer}, we conduct relative pose estimation evaluation on the MegaDepth~\cite{li2018megadepth} and Scannet~\cite{dai2017Scannet} datasets to demonstrate the effectiveness of our method.

The MegaDepth dataset~\cite{li2018megadepth} is a large-scale outdoor dataset consisting of 196 scenes and 1 million internet images. The refined poses and depth maps, which serve as ground truth, are computed using COLMAP and multiview stereo. Following~\cite{sun2021loftr,wang2024efficient,chen2022aspanformer}, we sample 1500 pairs from the “Sacre Coeur” and “St. Peter’s Square” scenes for validation. The images are resized so that the longer dimension is set to 832/640 for training and 1152/1056 for inference. The main challenges of this dataset include significant viewpoint and illumination changes, as well as the presence of repetitive patterns.

The Scannet dataset~\cite{dai2017Scannet} consists of 1613 sequences, each with ground truth camera poses and depth maps for every RGB image. The key challenges in this dataset include significant viewpoint changes, as well as repetitive or textureless patterns. Following~\cite{sarlin2020superglue,sun2021loftr}, we sampled 230K and 1500 image pairs for training and inference, respectively. All images are resized to 480$\times$640.

\vspace{1mm}
\noindent\textbf{Baselines.}
% We mainly compare our method with the two most competitive semi-dense baselines: Efficient LoFTR~\cite{wang2024efficient} and AspanFormer~\cite{chen2022aspanformer}. Since we find that the batch size setting during training is crucial, we provide two pre-trained models, Diff-Reg$_{px832/bs4}$ and Diff-Reg$_{px640/bs16}$, for comparison with the two baselines. As shown in Tab.~\ref{tab_2d2d}, we present several re-trained versions of the baselines. For example, ELOFTR$_{px832/bs20}$(px1152/R5): the subscript "px832/bs20" indicates that the model was trained with a resolution of 832 and a batch size of 20. The number in parentheses, "px1056", indicates that testing was performed at a resolution of 1056, and "R5" indicates the RANSAC algorithm was run 5 times. The "s4" in Diff-Reg$_{px640/bs16}$(px1056/s4) refers to setting the reverse sampling steps to 4. The pre-trained model weights provided by the authors\cite{wang2024efficient,chen2022aspanformer} are ELOFTR$_{px832/bs20}$ and AspanFormer$_{px832/bs8}$, which were trained at a resolution of 832 with batch sizes of 20 and 8, respectively. Other baselines include sparse keypoint detection and matching methods (i.e., SuperPoint (SP), SuperGlue (SG), and LightGlue (LG)) and semi-dense methods (i.e., LoFTR, QuadTree Attention, MatchFormer, and TopicFM), where their results are imported from the Efficient LoFTR\cite{wang2024efficient} and AspanFormer~\cite{chen2022aspanformer} papers.
We mainly compare our method with the two most competitive semi-dense baselines: Efficient LoFTR~\cite{wang2024efficient} and AspanFormer~\cite{chen2022aspanformer}. Since we find that the batch size setting during training is crucial, we provide two pre-trained models, Diff-Reg$_{px832/bs4}$ and Diff-Reg$_{px640/bs16}$, for comparison with the two baselines. As shown in Tab.~\ref{tab_2d2d}, we present several re-trained versions of the baselines. For example, ELOFTR$_{px832/bs20}$(px1152/R5): the subscript "px832/bs20" indicates that the model was trained with a resolution of 832 and a batch size of 20. The number in parentheses, "px1056", indicates that testing was performed at a resolution of 1056, and "R5" indicates that the RANSAC algorithm was run 5 times. The "s4" in Diff-Reg$_{px640/bs16}$(px1056/s4) refers to setting the reverse sampling steps to 4. The pre-trained model weights provided by the authors\cite{wang2024efficient,chen2022aspanformer} are ELOFTR$_{px832/bs20}$ and AspanFormer$_{px832/bs8}$, which were trained at a resolution of 832 with batch sizes of 20 and 8, respectively. Other baselines include sparse keypoint detection and matching methods (i.e., SuperPoint (SP), SuperGlue (SG), and LightGlue (LG)) and semi-dense methods (i.e., LoFTR, QuadTree Attention, MatchFormer, and TopicFM), whose results are imported from the original Efficient LoFTR~\cite{wang2024efficient} and AspanFormer~\cite{chen2022aspanformer} papers.

\begin{table*}
\caption{Quantitative results on the MegaDepth and Scannet benchmarks.}
\centering
\resizebox{\textwidth}{!}{
\begin{tabular}{c|ccc|c|ccc|c}
\toprule
\midrule
\multirow{2}{*}{Method}  & \multicolumn{3}{c}{MegaDepth Dataset} &\multirow{2}{*}{Method}& \multicolumn{3}{c}{Scannet Dataset} & \multirow{2}{*}{Time (ms)}\\
 &  AUC@5$^\circ$ & AUC@10$^\circ$ & AUC@20$^\circ$ &&  AUC@5$^\circ$ &AUC@10$^\circ$ & AUC@20$^\circ$ \\
\midrule
SP+SG &49.7&67.1&80.6&-&16.2(7.5) &33.8(18.6) &51.8(32.10)& 10.8\\   
SP+LG&49.9&67.0&80.1&-&-(16.2) &-(32.8) &-(49.7)& 48.3\\
\midrule
% DRC-Net&27.0&42.9&58.3&-&7.7&17.9 &30.5 &328.0\\
LoFTR&52.8&69.2&81.2&-&22.0(16.9)&40.8(33.6) &57.6(50.6)& 66.2\\
QuadTree&54.6&70.5&82.2&-&24.9(19.0)& 44.7(37.3) &61.8(53.5)& 100.7\\
MatchFormer&53.3&69.7&81.8&-&24.3(15.8)& 43.9(32.0)& 61.4(48.0)& 128.9\\
TopicFM&54.1&70.1&81.6&-&-(17.3) &-(35.5) &-(50.9)& 66.4\\
\midrule
ELOFTR$_{px832/bs20}$(px1152/R1) &55.06&71.61&83.32&-&-(18.47)&-(36.56)&-(53.35)&25.6\\
ELOFTR$_{px832/bs20}$(px1152/R5) &55.20&71.30&83.00&-&-&-&-&-\\
% \rowcolor[gray]{0.9} 
AspanFormer$_{px832/bs8}$(px1152)&53.06&69.77&81.84&AspanFormer$_{px640/bs24}$&26.14(22.85)&46.58(41.35)&63.83(58.23)&59.8\\
\midrule

% \rowcolor[gray]{0.9} 
ELOFTR$_{px640/bs16}$(px1056/R1) &53.96&70.59&82.43&ELOFTR$_{px640/bs16}$&20.50(17.67)&38.79(34.96)&55.77(51.57)&-\\ 
% \rowcolor[gray]{0.9} 
AspanFormer$_{px832/bs4}$(px1056)& 53.78&69.98 &82.09&{AspanFormer$_{px640/bs8}$}&-&-&-&-\\

Diff-Reg$_{px832/bs4}$(px1056/s5) & \underline{55.65}& \underline{71.57} & \underline{83.14}&{Diff-Reg$_{px640/bs8}$}& - &- &- &-\\
% \rowcolor[gray]{0.9} 
Diff-Reg$_{px640/bs16}$(px1056/s4)  & \textbf{55.68} & \textbf{71.99}&\textbf{83.34}&Diff-Reg$_{px640/bs16}$(s3)&23.17(15.97)  &42.27(32.18) &59.32(48.99) &164.2\\

\bottomrule
\end{tabular}
}\label{tab_2d2d}
\end{table*}

\begin{figure*}
      % \vspace{-0.5cm}
      \centering
      \includegraphics[width=\textwidth, height=5.6cm]{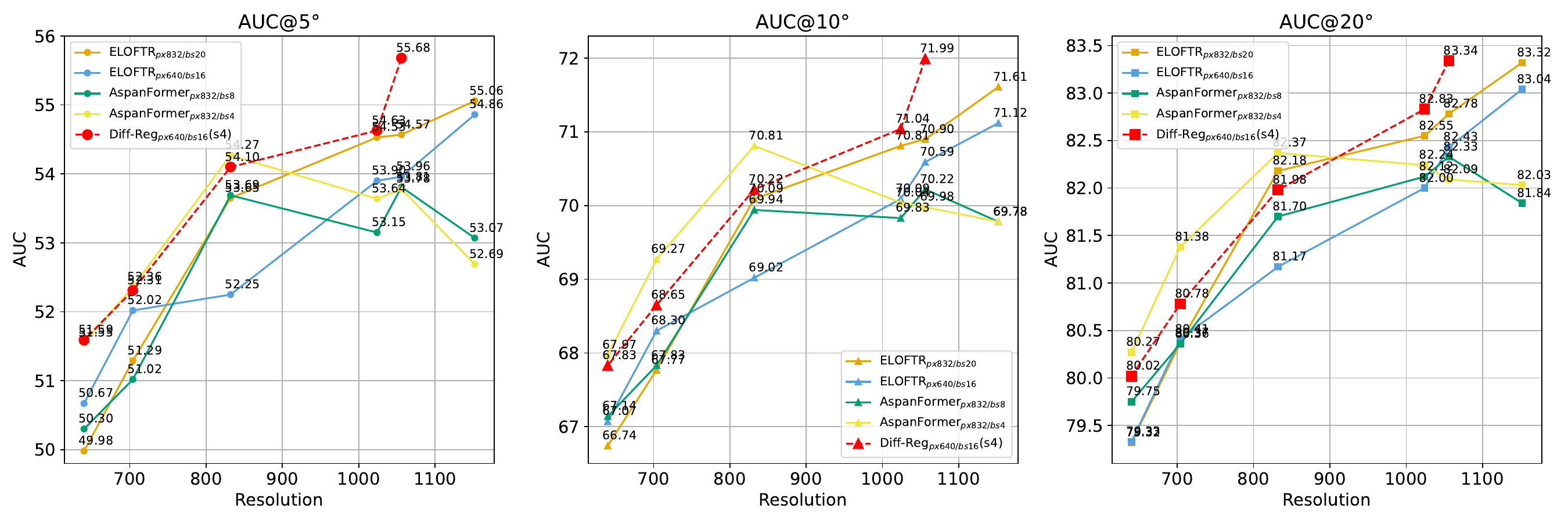}

\caption{Results of comparing method on the MegaDepth datasets on several resolutions: [640,704,832,1024,1056,1152]} \label{AUC_fig_res}    
% \vspace{-1cm}
\end{figure*}

\begin{figure*}[ht]
    \centering
    
    \begin{minipage}{\textwidth}
        \includegraphics[width=0.3\textwidth,height=2.2cm]{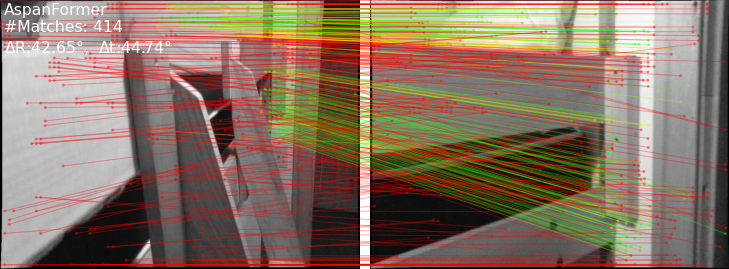}
        \hfill  % 插入空格以将图片分开
        \includegraphics[width=0.3\textwidth,height=2.2cm]{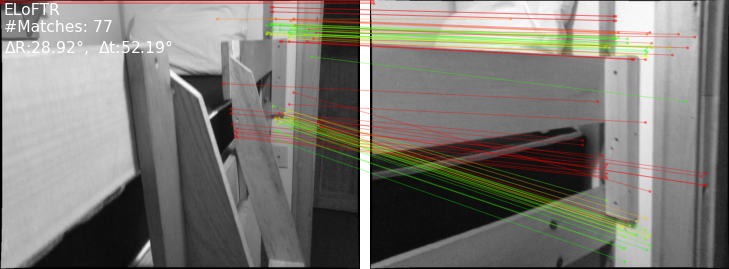}
        \hfill  % 插入空格以将图片分开
        \includegraphics[width=0.3\textwidth,height=2.2cm]{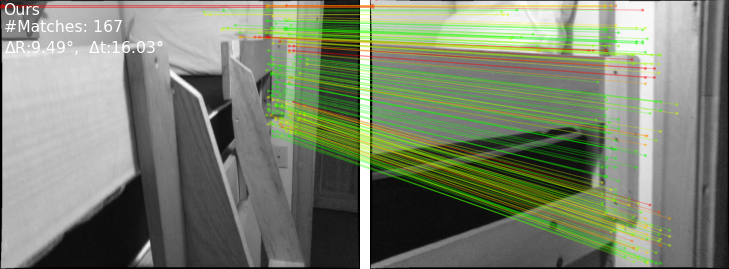} 
    \end{minipage}

    \begin{minipage}{\textwidth}

        \subfloat[AspanFormer$_{px640/bs24}$]{\includegraphics[width=0.3\textwidth,height=2.2cm]{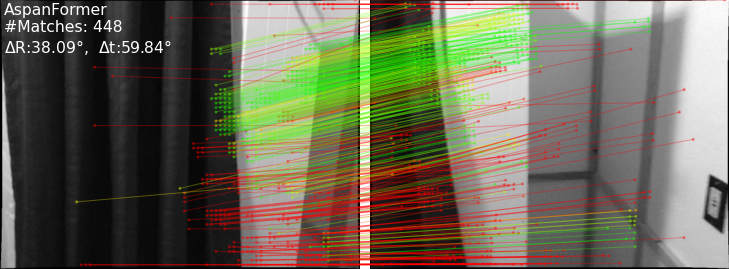}}
        \hfill  % 插入空格以将图片分开
        \subfloat[ELOFTR$_{px640/bs16}$(R1)]
        {\includegraphics[width=0.3\textwidth,height=2.2cm]{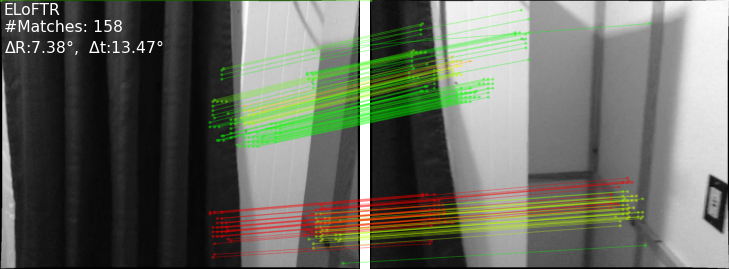}}
        \hfill  % 插入空格以将图片分开
        \subfloat[Diff-Reg$_{px640/bs16}$(s3)]
        {\includegraphics[width=0.3\textwidth,height=2.2cm]{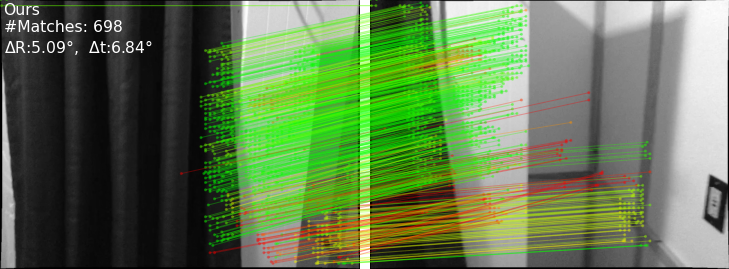}}
    \end{minipage}

    \begin{minipage}{\textwidth}

       \includegraphics[width=0.3\textwidth,height=2.2cm]{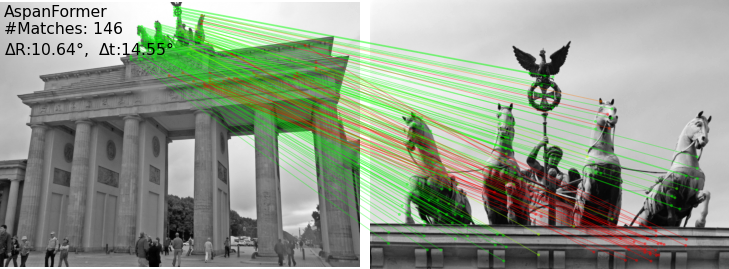}
        \hfill  % 插入空格以将图片分开
        \includegraphics[width=0.3\textwidth,height=2.2cm]{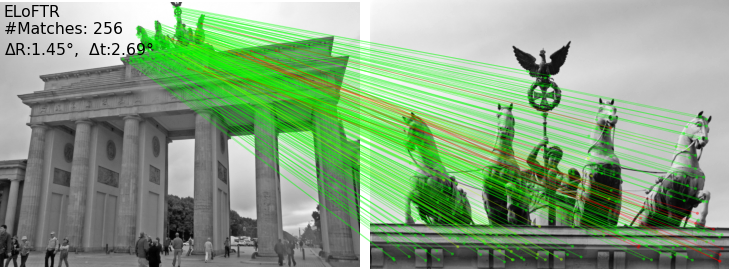}
        \hfill  % 插入空格以将图片分开
        \includegraphics[width=0.3\textwidth,height=2.2cm]{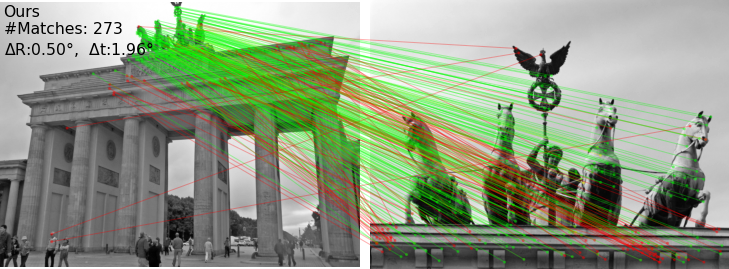}
    \end{minipage}

    \begin{minipage}{\textwidth}

        \subfloat[AspanFormer$_{px832/bs8}$]{\includegraphics[width=0.3\textwidth,height=2.2cm]{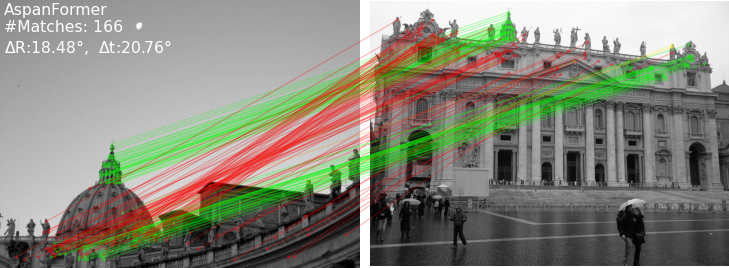}}
        \hfill  % 插入空格以将图片分开
        \subfloat[ELOFTR$_{px832/bs20}$(R1)]
        {\includegraphics[width=0.3\textwidth,height=2.2cm]{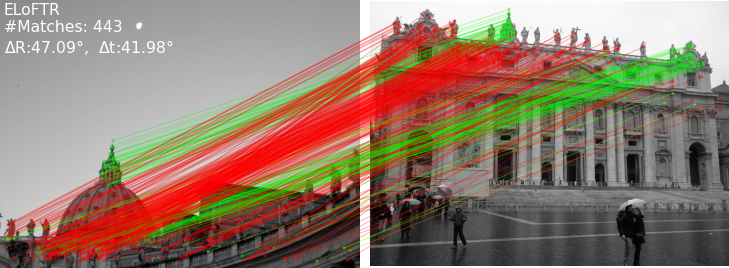}}
        \hfill  % 插入空格以将图片分开
        \subfloat[Diff-Reg$_{px640/bs16}$(s4)]
        {\includegraphics[width=0.3\textwidth,height=2.2cm]{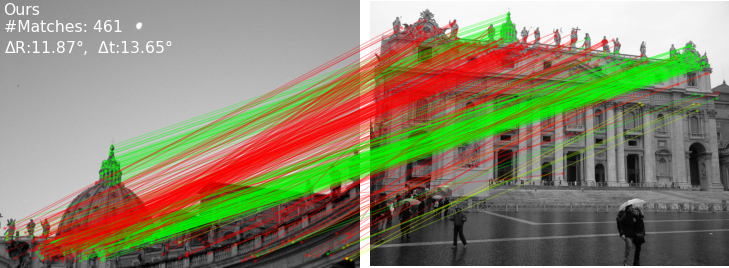}}
    \end{minipage}

    \caption{Qualitative results of MegaDepth and Scannet Benchmark.}
    \label{fig:im_quanitative_results}
\end{figure*}

\begin{figure*}
    \centering

    \begin{minipage}{\textwidth}
        \includegraphics[width=0.3\textwidth,height=2.2cm]{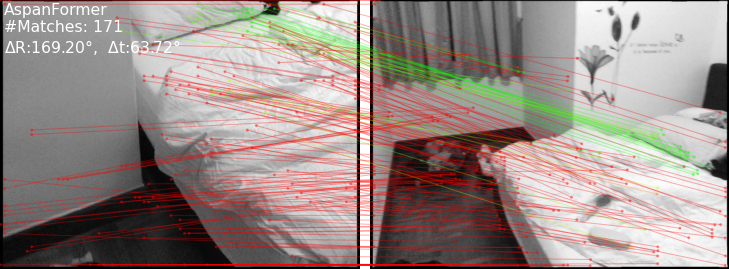}
        \hfill  % 插入空格以将图片分开
        \includegraphics[width=0.3\textwidth,height=2.2cm]{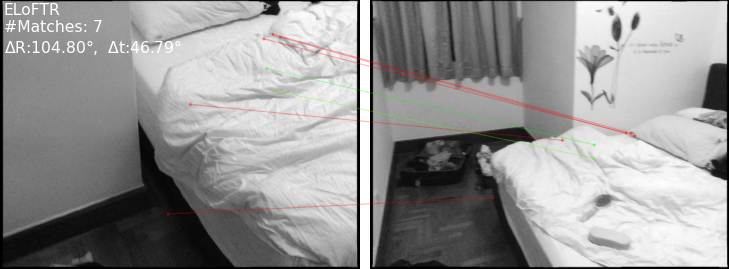}
        \hfill  % 插入空格以将图片分开
        \includegraphics[width=0.3\textwidth,height=2.2cm]{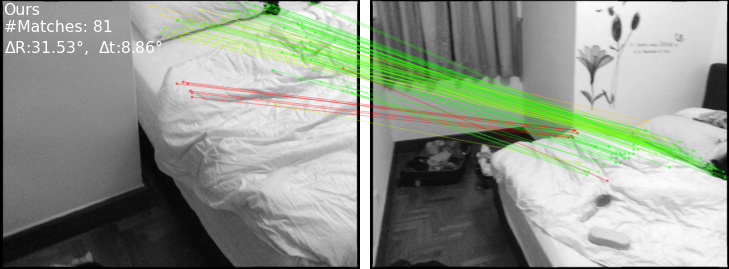} 
    \end{minipage}
    
    \begin{minipage}{\textwidth}

        \subfloat[AspanFormer$_{px640/bs24}$]{\includegraphics[width=0.3\textwidth,height=2.2cm]{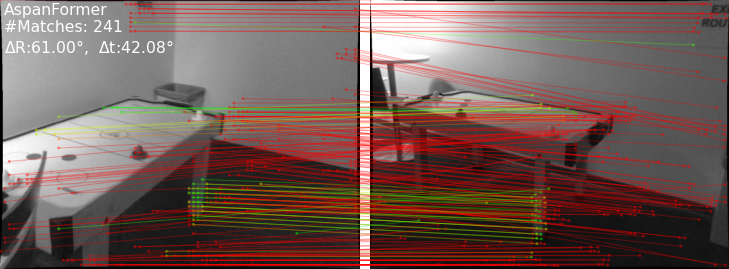}}
        \hfill  % 插入空格以将图片分开
        \subfloat[ELOFTR$_{px640/bs16}$(R1)]
        {\includegraphics[width=0.3\textwidth,height=2.2cm]{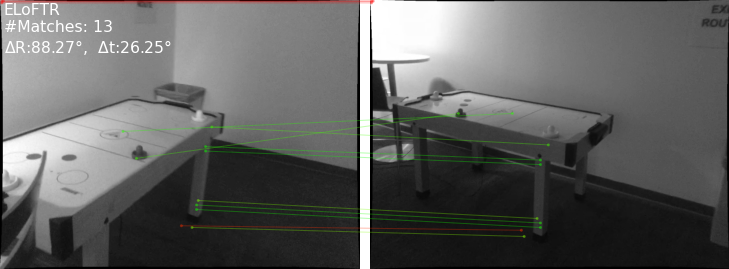}}
        \hfill  % 插入空格以将图片分开
        \subfloat[Diff-Reg$_{px640/bs16}$(s3)]
        {\includegraphics[width=0.3\textwidth,height=2.2cm]{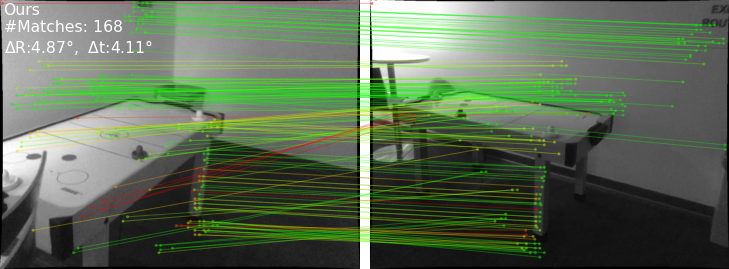}}
    \end{minipage}    
    
    \begin{minipage}{\textwidth}
        \includegraphics[width=0.3\textwidth,height=2.2cm]{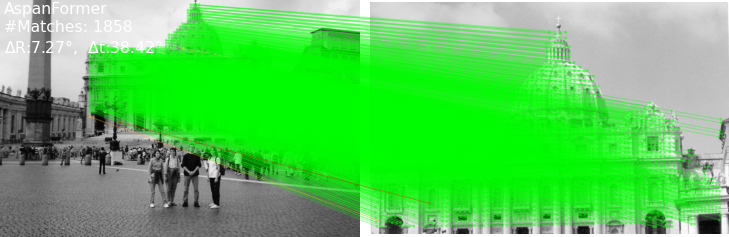}
        \hfill  % 插入空格以将图片分开
        \includegraphics[width=0.3\textwidth,height=2.2cm]{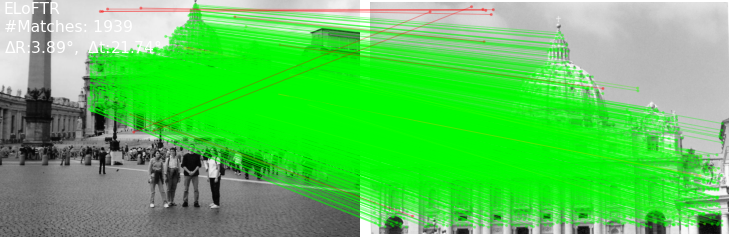}
        \hfill  % 插入空格以将图片分开
        \includegraphics[width=0.3\textwidth,height=2.2cm]{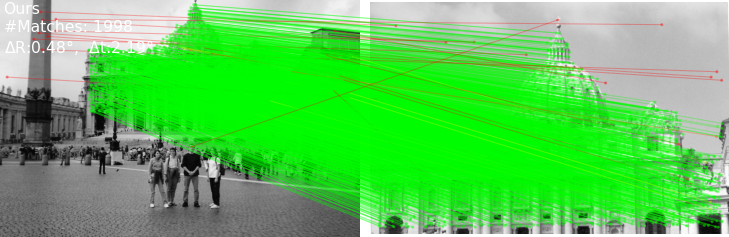} 
    \end{minipage}
    
    \begin{minipage}{\textwidth}
        \includegraphics[width=0.3\textwidth,height=2.2cm]{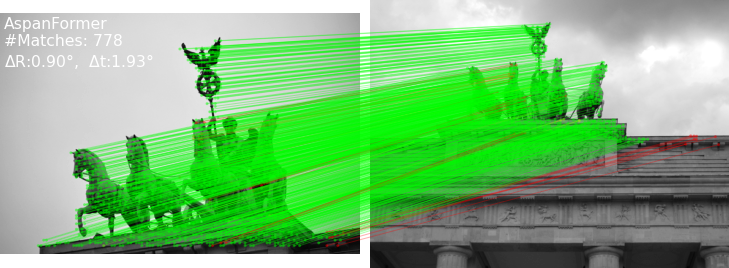}
        \hfill  % 插入空格以将图片分开
        \includegraphics[width=0.3\textwidth,height=2.2cm]{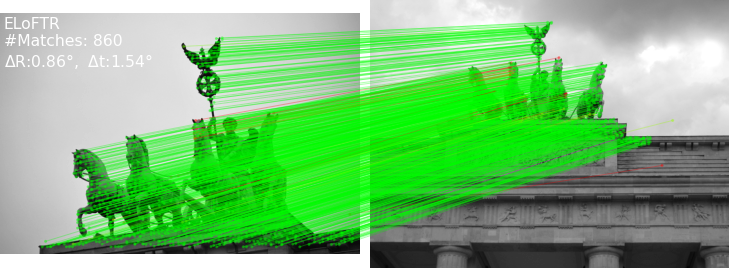}
        \hfill  % 插入空格以将图片分开
        \includegraphics[width=0.3\textwidth,height=2.2cm]{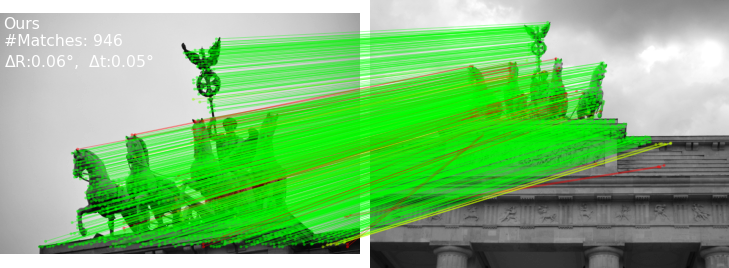} 
    \end{minipage}

    \begin{minipage}{\textwidth}

        \subfloat[AspanFormer$_{px832/bs8}$]{\includegraphics[width=0.3\textwidth,height=2.8cm]{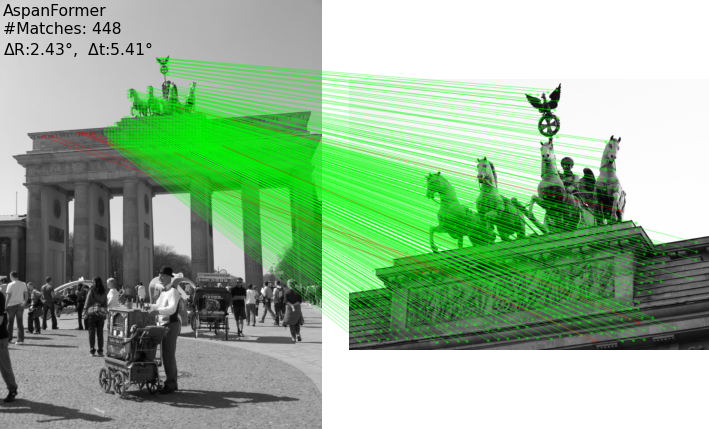}}
        \hfill  % 插入空格以将图片分开
        \subfloat[ELOFTR$_{px832/bs20}$(R1)]
        {\includegraphics[width=0.3\textwidth,height=2.8cm]{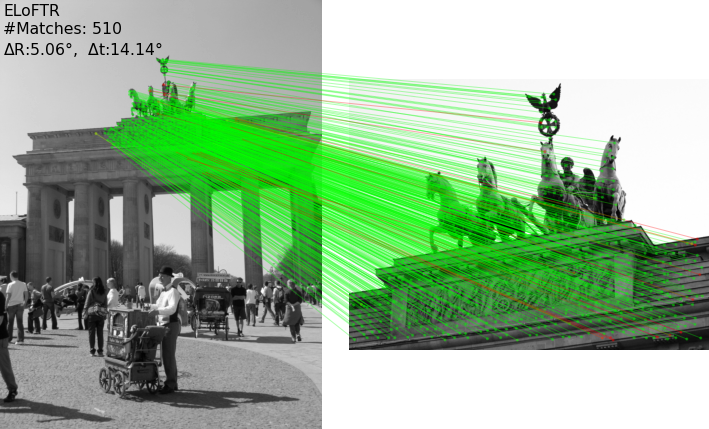}}
        \hfill  % 插入空格以将图片分开
        \subfloat[Diff-Reg$_{px640/bs16}$(s4)]
        {\includegraphics[width=0.3\textwidth,height=2.8cm]{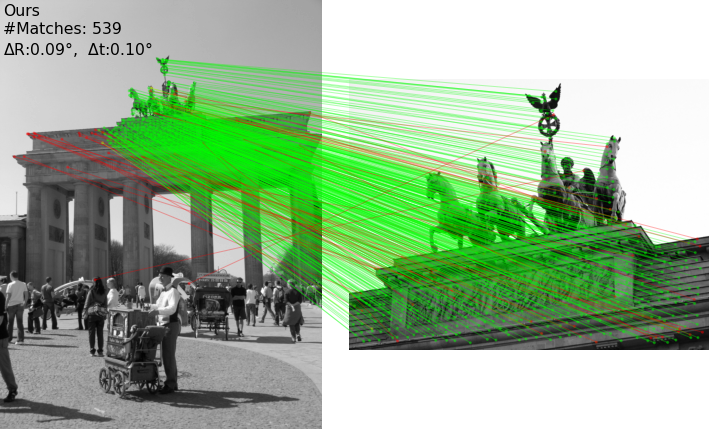}}
    \end{minipage}

    \caption{More qualitative results of MegaDepth and Scannet Benchmark. }
    \label{fig:more_im_quanitative_results}
\end{figure*}

\vspace{1mm}
\noindent\textbf{Metrics.}
Following~\cite{sun2021loftr,chen2022aspanformer}, we compared our method with baselines by the pose errors metric, which is defined as the biggest error of rotation and translation angular errors. We report the AUC of the pose error over three thresholds (5$^\circ$,10$^\circ$,20$^\circ$). We also report the running time of all methods for matching each image pair in the Scannet dataset on an NVIDIA RTX 4090 GPU.

\noindent\textbf{Quantitative results.}
Due to GPU memory limitations, for the MegaDepth benchmark, we resize the images such that the longest dimension is 1056 pixels. We compare Diff-Reg$_{px640/bs16}$ with ELOFTR$_{px640/bs16}$, and compare Diff-Reg$_{px832/bs4}$ with AspanFormer$_{px832/bs4}$. 

Additionally, we use one-time RANSAC for ELOFTR (i.e., ELOFTR$_{px640/bs16}$(px1056/R1)), since the performance of ELOFTR$_{px640/bs20}$(px1152/R1) is similar to that of ELOFTR$_{px640/bs20}$(px1152/R5). The pose error results in Tab.~\ref{tab_2d2d} show that Diff-Reg$_{px640/bs16}$(px1056/s4) outperforms ELOFTR$_{px640/bs16}$(px1056/R1), and  Diff-Reg$_{px832/bs4}$ outperforms AspanFormer$_{px832/bs4}$ across all thresholds. We also find that Diff-Reg$_{px640/bs16}$(px1056/s4) achieves competitive performance compared to ELOFTR$_{px832/bs20}$(px1152/R1) and AspanFormer$_{px832/bs8}$(px1152), which indicate that our method has the potential to achieve further performance gains at a higher resolution (e.g., 1152).

% Due to GPU memory constraints, we resize images in the MegaDepth benchmark such that the longest dimension is 1056 pixels. We first compare Diff-Reg and ELOFTR with settings (px640/bs16, resolution 1056). Similarly, we evaluate Diff-Reg against AspanFormer under the (px832/bs4) setting. For ELOFTR, we additionally apply one-time RANSAC (px640/bs16, 1056/R1), as its performance is similar to that of (px640/bs20, 1152/R5). Pose error results (Tab.~\ref{tab_2d2d}) show that Diff-Reg (px640/bs16, 1056/s4) clearly outperforms ELOFTR under the same setting, and also surpasses AspanFormer (px832/bs4) across all thresholds. Moreover, Diff-Reg (px640/bs16, 1056/s4) achieves comparable performance to both ELOFTR (px832/bs20, 1152/R1) and AspanFormer (px832/bs8, 1152), suggesting its potential to yield further gains at higher resolutions (e.g., 1152).

Due to the iterative nature of the reverse sampling steps in our method's inference, Diff-Reg requires more time (refer to the last column in Tab.~\ref{tab_2d2d}) to find the optimal coarse-level matching matrix. In this paper, our main objective is to introduce diffusion-based matching matrix estimation into image matching. However, with the advances in diffusion acceleration research~\cite{song2023consistency,liu2022flow} (e.g., one-step denoising), there is still room for improvement in sampling speed. We leave this enhancement for future work.

Since AspanFormer demonstrates strong performance on the indoor Scannet dataset, we also conduct a study to determine whether it can achieve similar results on the outdoor MegaDepth dataset. We perform experiments at resolutions of 640, 704, 832, 1024, 1056, and 1152. The results are shown in Fig.~\ref{AUC_fig_res}, where we observe that AspanFormer performs well at lower resolutions but exhibits poor performance at higher resolutions. This phenomenon may indicate that the multi-level or Adaptive Span Attention design does not work effectively at higher resolutions. Both the pre-trained model AspanFormer$_{px832/bs8}$ provided by the authors and our retrained version AspanFormer$_{px832/bs4}$ show similar results. In contrast, our method demonstrates progressively greater performance gains as the image resolution increases, which aligns with common sense and intuition: higher resolution or a broader receptive field leads to better correspondences.

\noindent\textbf{Qualitative results.}
We also present the visualization in Fig.~\ref{fig:im_quanitative_results}. Our method tends to identify more correspondences within a greater number of potential matching regions (i.e., rows 2 and 4 in Fig.~\ref{fig:im_quanitative_results}). In addition to the visualizations provided in Fig.~\ref{fig:more_im_quanitative_results}, our method further refines the correspondences, achieving more accurate matches. For example, in row 3 of Fig.~\ref{fig:more_im_quanitative_results}, the fine-grained matching obtained by AspanFormer still appears as a coarse block-like region (please zoom in), which results in a decrease in the accuracy of the computed pose.

\subsection{3D Registration}
Our method is trained and evaluated using PyTorch on an NVIDIA RTX 3090 GPU. We train the model for approximately 30 epochs on the 3DMatch and 4DMatch datasets with a batch size of 2. We adopt the training/validation/test split strategy from~\cite{huang2021predator,li2022lepard} for 3DMatch and 4DMatch datasets. The total diffusion steps during training are set to 1000. The dimension $d$ of superpoint features $F^{\hat{P}}$ and $F^{\hat{Q}}$ is set as $d = 528$. 

\subsubsection{\textbf{Non-Rigid 4DMatch/4DLoMatch Benchmark}}
\vspace{1mm}

\noindent\textbf{Datasets.}
4DMatch/4DLoMatch\cite{li2022lepard} is an benchmark generated by the animation sequences from DeformingThings4D\cite{li20214dcomplete}. We follow the dataset split provided in \cite{li2022lepard}, which has a wide range of overlap ratios, that $45\%$-$92\%$ in 4DMatch and $15\%$-$45\%$ in 4DLoMatch. 

\noindent\textbf{Metrics.}
Following Lepard \cite{li2022lepard}, we utilize two evaluation metrics to assess the quality of predicted matches. (1) Inlier ratio (\textbf{IR}): This metric represents the correct fraction in the correspondences prediction $\mathcal{K}_{pred}$. (2) Non-rigid Feature Matching Recall (\textbf{NFMR}): This metric calculates the fraction of ground truth correspondences $(u,v)\in \mathcal{K}_{gt}$ that can be successfully recovered by using the predicted correspondences $\mathcal{K}_{pred}$ as anchors. The NFMR metric provides a better characterization of the global rationality of overall body deformation, directly indicating whether the anchor $\mathcal{K}_{pred}$ effectively captures the body movements. More details of the metric definition can be found in the supplementary materials.

\noindent\textbf{Quantitative results.}
We compare our method with two categories of state-of-the-art methods. The first category includes Scene Flow Methods such as PWC\cite{wu2019pointpwc} and FLOT\cite{puy2020flot}. The second category encapsulates Feature Matching-Based Methods, namely D3Feat\cite{bai2020d3feat}, Predator\cite{huang2021predator}, Lepard\cite{li2022lepard}, GeoTR\cite{qin2022geometric}, and RoITr\cite{yu2023rotation}. As illustrated in Table \ref{tab_4dmatch}, our method demonstrates significant improvements compared to the single-step baselines. Diff-Reg(Backbone) refers to the single-step prediction head (i.e., reposition transformer in Lepard \cite{li2022lepard}), while Diff-Reg(steps=1) and Diff-Reg(steps=20) utilize our denoising module $g_{\theta}$ with one single step and 20 steps of reverse denoising sampling. For both NFMR and IR metrics, Diff-Reg(steps=20) achieves the best performance. The improvement in NFMR of Diff-Reg(Backbone) compared to the baselines indicates that our diffused training samples in the matching matrix space enhance the feature backbone's representation, enabling the capture of crucial salient correspondences that are helpful for consistent global deformation. The significant enhancement of Diff-Reg(steps=20) over Diff-Reg(steps=1) demonstrates that the reverse denoising sampling process indeed searches for a better solution guided by the learned posterior distribution. To validate that the predicted correspondences indeed improve deformable registration, we conducted experiments using the state-of-the-art registration method GraphSCNet~\cite{qin2023deep}. As indicated in Table \ref{non_rigid_registration}, our predicted correspondences are beneficial for deformable registration, particularly in the more challenging 4DLoMatch benchmark.

\begin{table*}
% \vspace{-0.5cm}
\caption{Non-rigid registration results of 4DMatch/4DLoMatch. Given predicted correspondences, we utilize the non-rigid registration method GraphSCNet\cite{qin2023deep} to conduct the deformable registration. We retrain RoITr$^*$ using the authors' code.}
\centering
\resizebox{1.0\textwidth}{!}{
\begin{tabular}{c|cccc|cccc}
\toprule
 \midrule
\multirow{2}{*}{Method}& \multicolumn{4}{c}{4DMatch-F} & \multicolumn{4}{c}{4DLoMatch-F} \\
 &   EPE$\downarrow$& AccS$\uparrow$ & AccR$\uparrow$ & Outlier$\downarrow$& EPE$\downarrow$& AccS$\uparrow$ & AccR$\uparrow$ & Outlier$\downarrow$     \\
 \midrule
 PointPWC\cite{wu2019pointpwc} & 0.182& 6.25& 21.49& 52.07 &0.279 &1.69& 8.15& 55.70\\
 FLOT\cite{puy2020flot} & 0.133& 7.66& 27.15 &40.49& 0.210 &2.73& 13.08& 42.51\\
 GeomFmaps [9]& 0.152 &12.34& 32.56 &37.90& 0.148& 1.85 &6.51 &64.63\\
 Synorim-pw [19] &0.099 &22.91& 49.86&26.01 &0.170 &10.55 &30.17& 31.12\\

Lepard\cite{li2022lepard}$+$GraphSCNet\cite{qin2023deep}&\underline{0.042}&70.10 &83.80& \underline{9.20}&  \underline{0.102}& \underline{40.00}& \underline{59.10}& \underline{17.50}  \\

 GeoTR\cite{qin2022geometric}$+$GraphSCNet\cite{qin2023deep}& 0.043 &\underline{72.10} &\underline{84.30}& 9.50& 0.119& 41.00& 58.40& 20.60   \\
RoITr$^*$\cite{yu2023rotation} $+$GraphSCNet\cite{qin2023deep}& 0.056 &59.60 &80.50& 12.50&0.118&32.30&56.70&20.50  \\
Diff-Reg$+$GraphSCNet\cite{qin2023deep}& \textbf{0.041} &\textbf{73.20} &\textbf{85.80}& \textbf{8.30}& \textbf{0.095}& \textbf{43.80}& \textbf{62.90}& \textbf{15.50}   \\
 \bottomrule
\end{tabular}}
\label{non_rigid_registration}

\end{table*}

\begin{figure*}[!t]
    \centering
    % 第一排图片
    \begin{minipage}{\textwidth}
        \centering
        \includegraphics[width=0.15\textwidth,height=2.2cm]{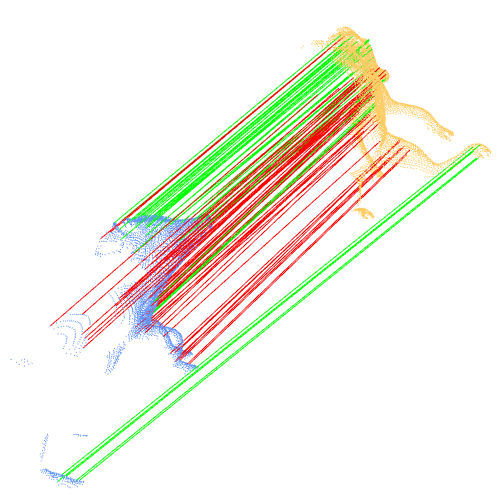}
        \hfill  % 插入空格以将图片分开
        \includegraphics[width=0.15\textwidth,height=2.2cm]{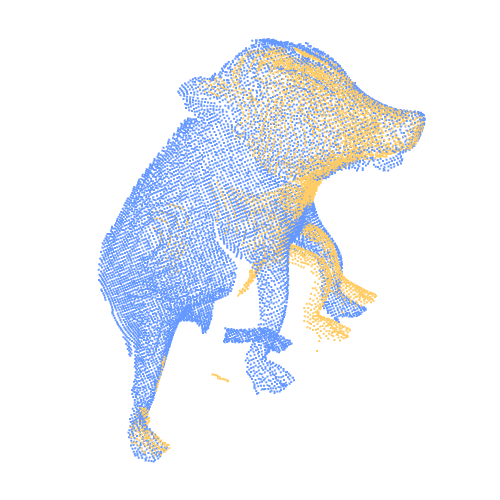}
        \hfill  % 插入空格以将图片分开
        \includegraphics[width=0.15\textwidth,height=2.2cm]{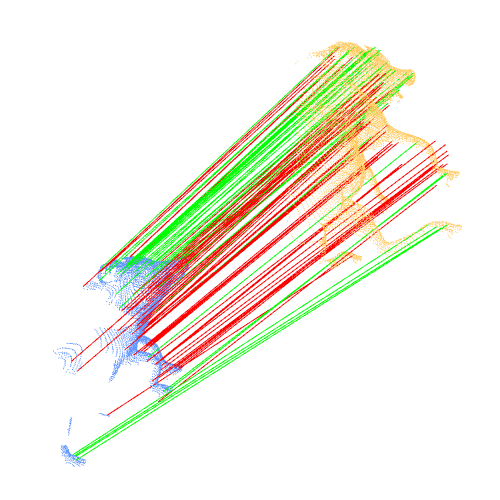}
        \hfill  % 插入空格以将图片分开
        \includegraphics[width=0.15\textwidth,height=2.2cm]{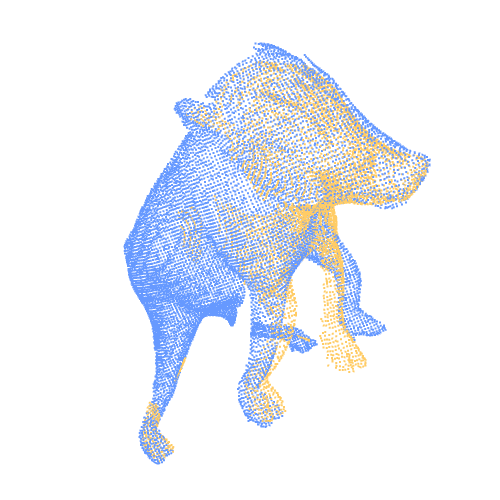} 
        \hfill  % 插入空格以将图片分开
        \includegraphics[width=0.15\textwidth,height=2.2cm]{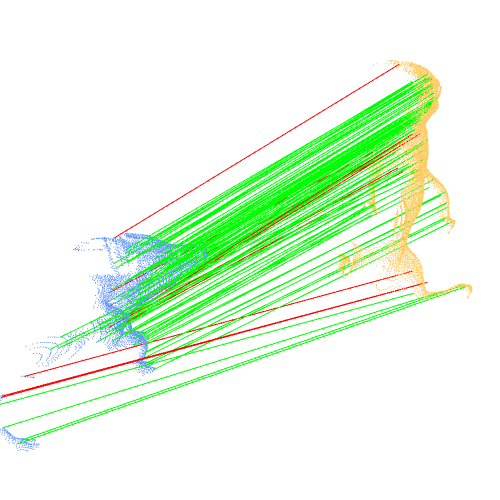} 
        \includegraphics[width=0.15\textwidth,height=2.2cm]
        {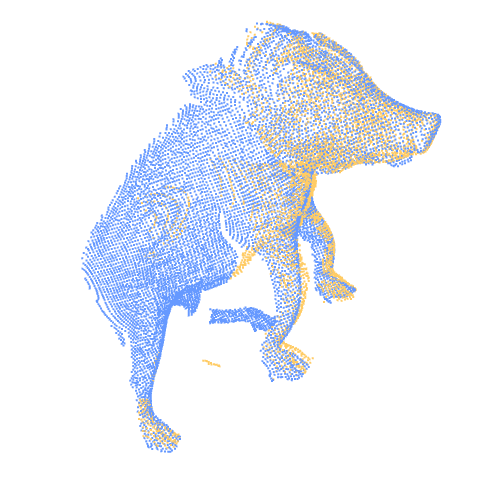} 
    \end{minipage}

    \begin{minipage}{\textwidth}
        \includegraphics[width=0.15\textwidth,height=2.2cm]{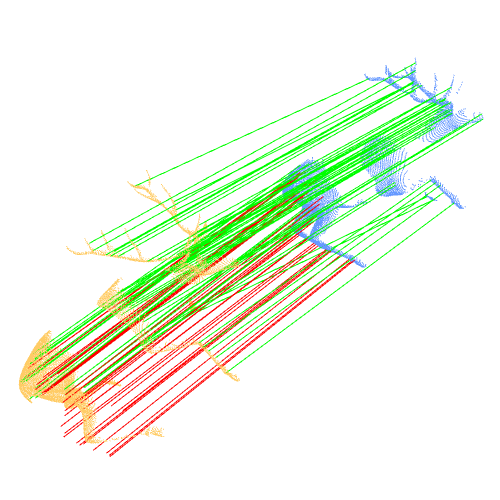}
        \hfill  % 插入空格以将图片分开
        \includegraphics[width=0.15\textwidth,height=2.2cm]{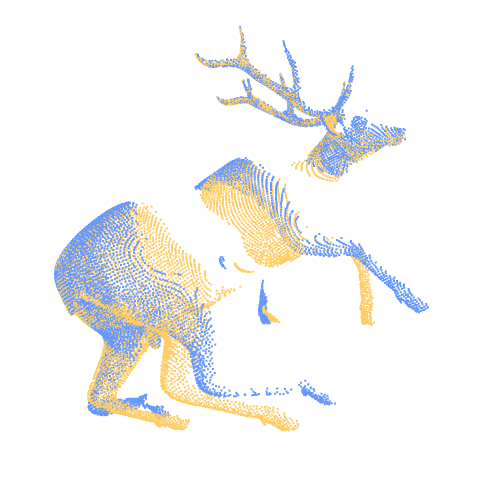}
        \hfill  % 插入空格以将图片分开
        \includegraphics[width=0.15\textwidth,height=2.2cm]{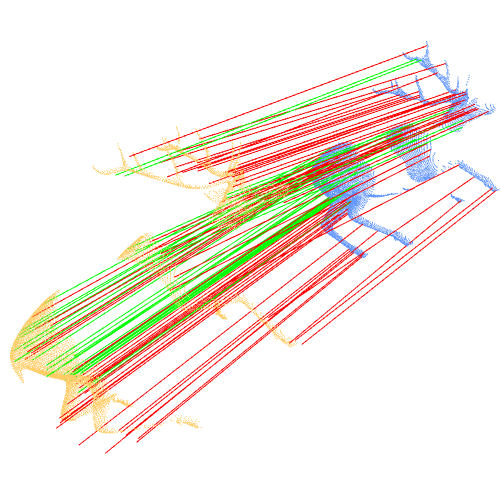}
        \hfill  % 插入空格以将图片分开
        \includegraphics[width=0.15\textwidth,height=2.2cm]{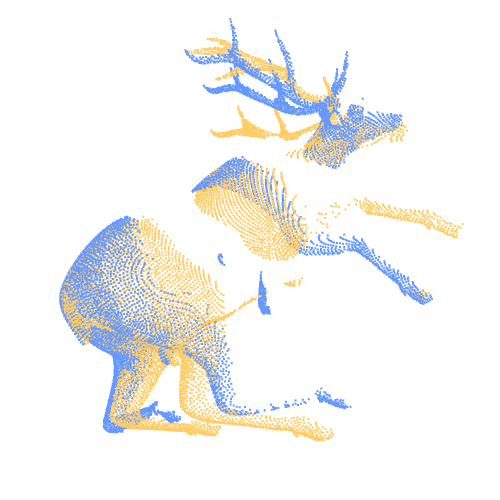} 
        \hfill  % 插入空格以将图片分开
        \includegraphics[width=0.15\textwidth,height=2.2cm]{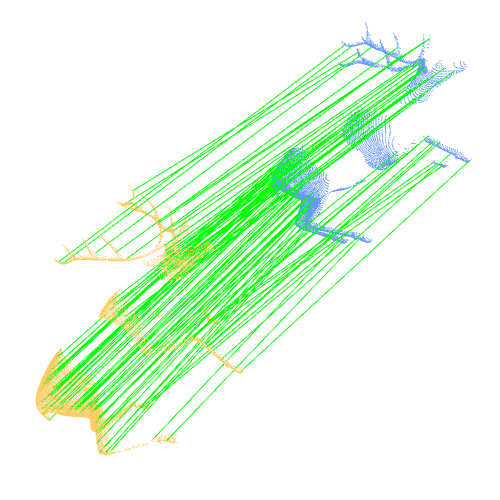} 
        \includegraphics[width=0.15\textwidth,height=2.2cm]
        {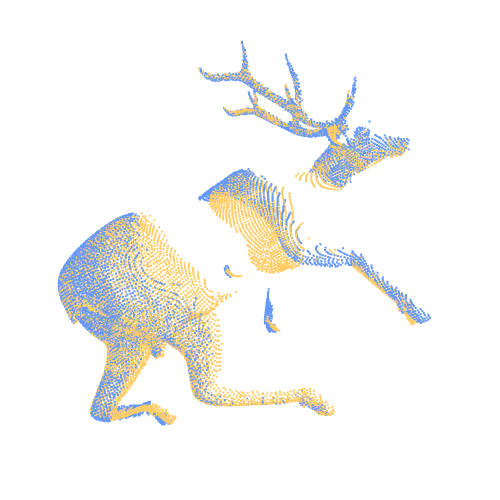} 
    \end{minipage}
    
    % % 空一行作为图片间距
    % \vspace{1cm}
    
    % 第二排图片
    \begin{minipage}{\textwidth}
        \centering
        \includegraphics[width=0.15\textwidth,height=2.2cm]{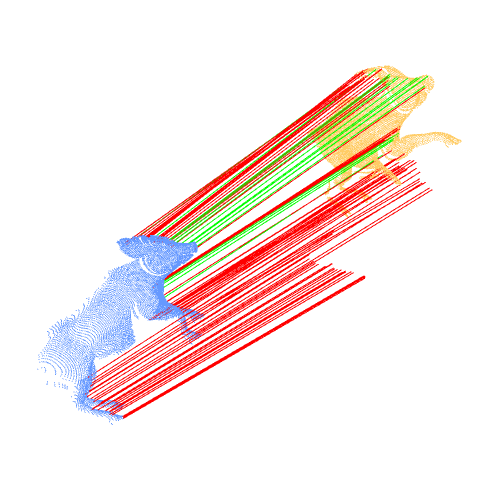}
        \hfill  % 插入空格以将图片分开
        \includegraphics[width=0.15\textwidth,height=2.2cm]{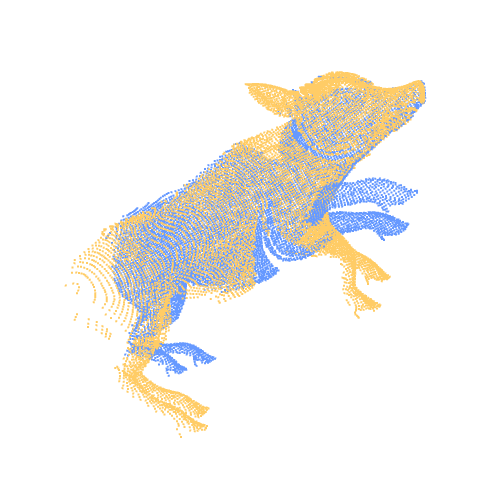}
        \hfill  % 插入空格以将图片分开
        \includegraphics[width=0.15\textwidth,height=2.2cm]{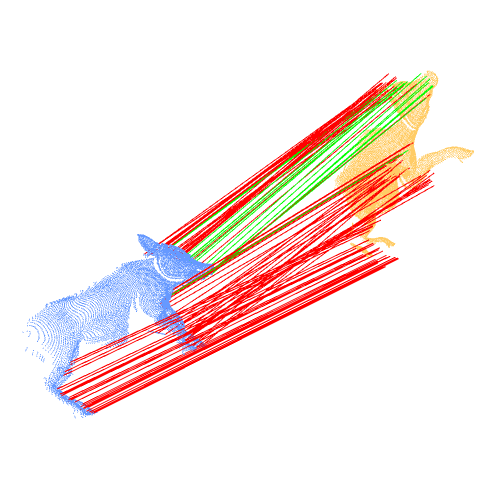}
        \hfill  % 插入空格以将图片分开
        \includegraphics[width=0.15\textwidth,height=2.2cm]{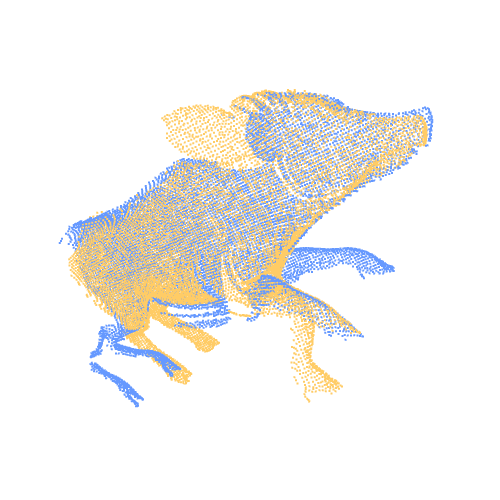} 
        \hfill  % 插入空格以将图片分开
        \includegraphics[width=0.15\textwidth,height=2.2cm]{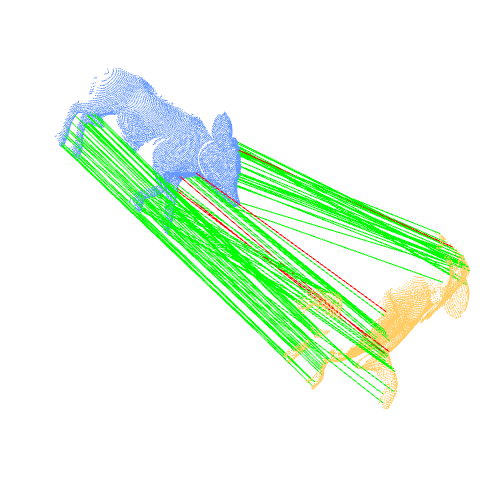} 
        \includegraphics[width=0.15\textwidth,height=2.2cm]{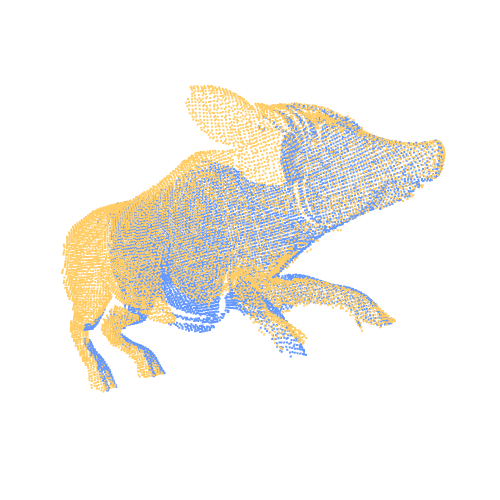}     
    \end{minipage}
    
    \begin{minipage}{\textwidth}
        \includegraphics[width=0.15\textwidth,height=2.2cm]{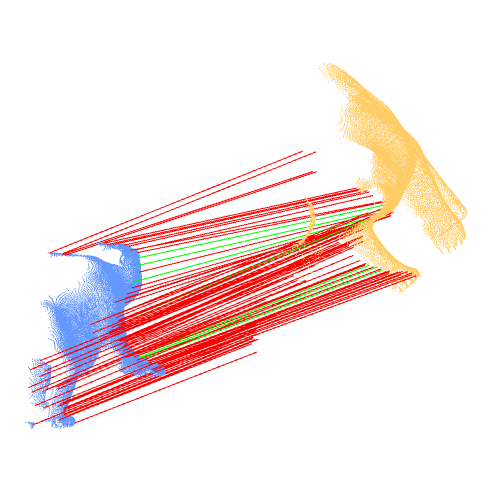}
        \hfill  % 插入空格以将图片分开
        \includegraphics[width=0.15\textwidth,height=2.2cm]{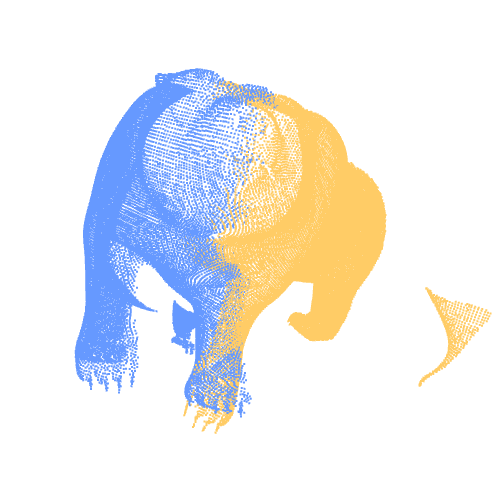}
        \hfill  % 插入空格以将图片分开
        \includegraphics[width=0.15\textwidth,height=2.2cm]{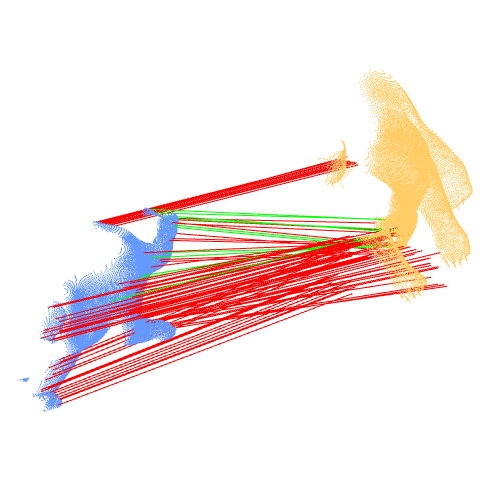}
        \hfill  % 插入空格以将图片分开
        \includegraphics[width=0.15\textwidth,height=2.2cm]{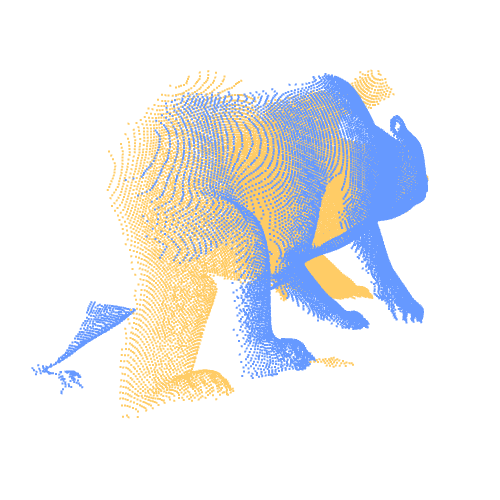} 
        \hfill  % 插入空格以将图片分开
        \includegraphics[width=0.15\textwidth,height=2.2cm]{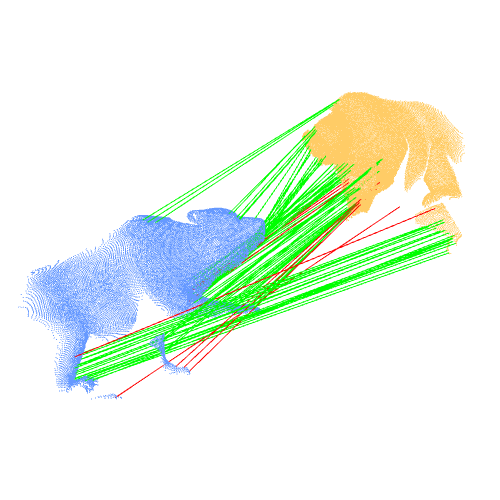} 
        \includegraphics[width=0.15\textwidth,height=2.2cm]{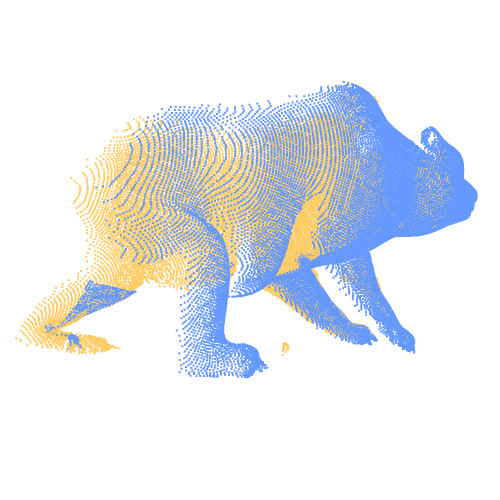}  
    \end{minipage}
    
    \begin{minipage}{\textwidth}
     
    \quad  % 插入一些空格以将图片分开
    \subfloat[GeoTR\cite{qin2022geometric}]{%
        \includegraphics[width=0.15\textwidth, height=2.2cm]{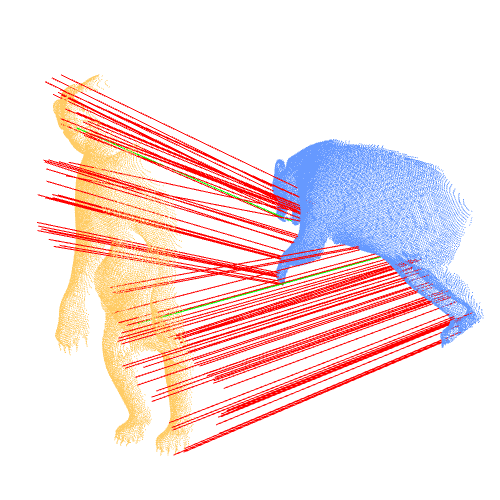} 
         \hfill
        \includegraphics[width=0.15\textwidth, height=2.2cm]{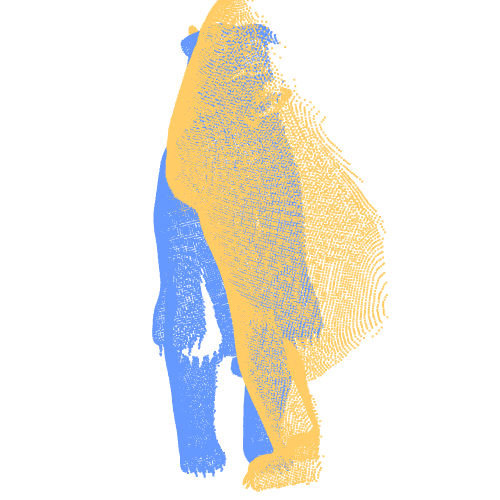}%
        \label{fig:sub5a}
    }
    \quad  % 插入一些空格以将图片分开
    \subfloat[RoITr\cite{yu2023rotation}]{%
        \includegraphics[width=0.15\textwidth, height=2.2cm]{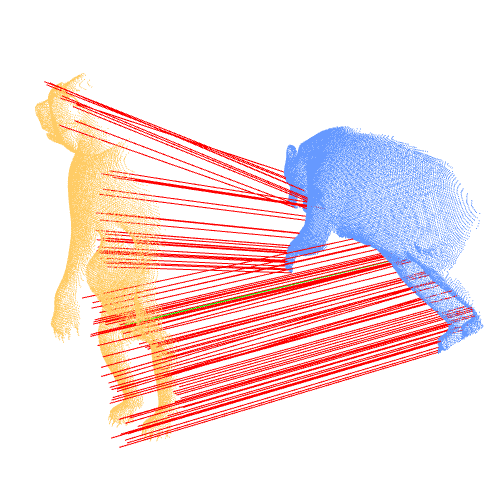} 
         \hfill
        \includegraphics[width=0.15\textwidth, height=2.2cm]{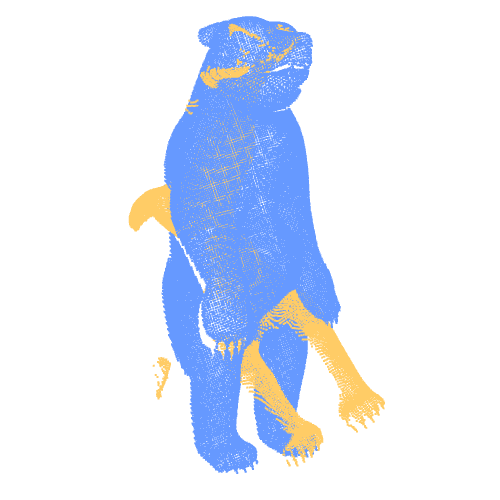}%
        \label{fig:sub5b}
    }
    \quad  % 插入一些空格以将图片分开
    \subfloat[Diff-Reg(steps=20)]{%
        \includegraphics[width=0.15\textwidth, height=2.2cm]{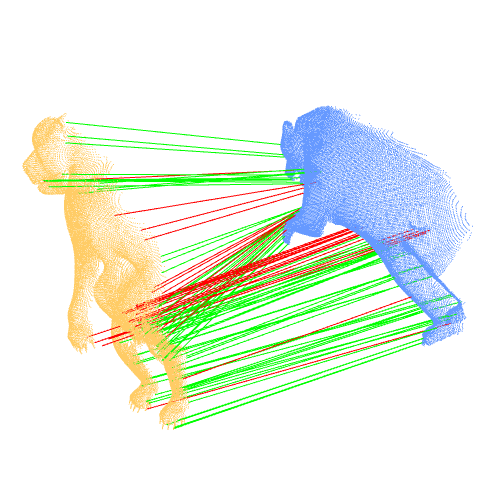} 
         \hfill
        \includegraphics[width=0.15\textwidth, height=2.2cm]{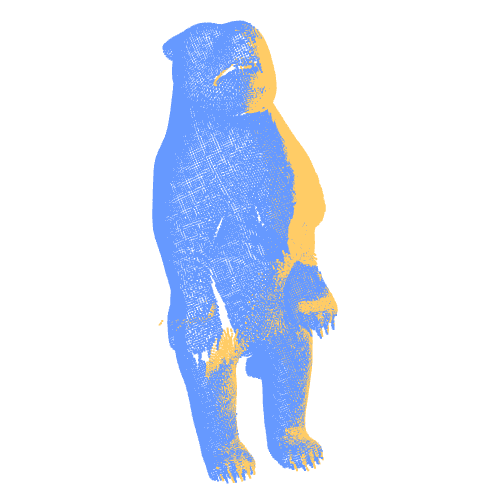}%
        \label{fig:sub5c}
    }

    \label{4dmatchi_res_vis}
    \end{minipage}
    
    \caption{The qualitative results of non-rigid registration in the 4DMatch/4DLoMatch benchmark. The top two lines are from 4DMatch, while the bottom three are from 4DLoMatch. The \textcolor{blue}{blue} and \textcolor{yellow}{yellow} colors denote the source and target point cloud, respectively. The \textcolor{green}{green} and \textcolor{red}{red} lines indicate whether the threshold accepts the predicted deformable flow from the source points. The deformable registration is built by GraphSCNet\cite{li20232D3D}. Zoom in for details.}
    \label{non_rigid_registration_vis}
\end{figure*}

\noindent\textbf{Qualitative results.}
We provide a visualization to demonstrate our method's effectiveness in Fig.~\ref{non_rigid_registration_vis}. For a fair comparison, we exploit the ``metric index'' (i.e., the test point set in the 4DMatch/4DLoMatch dataset, in \textcolor{blue}{blue} color) of the source point cloud for all methods. Taking the predicted correspondences from RoITr \cite{yu2023rotation}, GeoTr \cite{yu2023rotation}, and ``Our (steps=20)'' as anchor correspondences, we calculate the deformation flow for the source test points by applying neighborhood k-nearest neighbors (knn) interpolation based on the anchors. The deformable registration of the bear's two front paws in the first row and fourth column reveals that dealing with ambiguous matching patches of asymmetric objects can be highly challenging. However, our denoising process is capable of handling this scenario perfectly. The deformable registration results from the top three rows (from the 4DMatch benchmark) indicate that the baseline methods struggle to compute reliable and consistent correspondences between scans with large deformations. The bottom two rows (from the 4DLoMatch benchmark) also demonstrate that low overlapping, combined with deformation, results in a disaster. These visualizations demonstrate that our denoising module in the matching matrix space may provide a more effective approach for tackling these challenges. 

\begin{table}
\caption{Quantitative results on the 4DMatch and 4DLoMatch benchmarks. The best results are highlighted in bold, and the second-best results are underlined.}

\centering
\resizebox{0.5\textwidth}{!}{
\begin{tabular}{c|c|cc|cc}
\toprule
\midrule
\multirow{2}{*}{Category} & \multirow{2}{*}{Method} & \multicolumn{2}{c}{4DMatch} & \multicolumn{2}{c}{4DLoMatch} \\
                          &                          & NFMR$\uparrow$ & IR$\uparrow$ & NFMR$\uparrow$ & IR$\uparrow$ \\
\midrule
\multirow{2}{*}{Scene Flow} & PointPWC \cite{wu2019pointpwc} & 21.60 & 20.0 & 10.0 & 7.20 \\
                            & FLOT \cite{puy2020flot}        & 27.10 & 24.90 & 15.20 & 10.70 \\
\midrule
\multirow{5}{*}{Feature Matching} & D3Feat \cite{bai2020d3feat} & 55.50 & 54.70 & 27.40 & 21.50 \\
                                   & Predator \cite{huang2021predator} & 56.40 & 60.40 & 32.10 & 27.50 \\
                                   & Lepard \cite{li2022lepard}      & 83.60 & 82.64 & 66.63 & 55.55 \\
                                   & GeoTR \cite{qin2022geometric}  & 83.20 & 82.20 & 65.40 & 63.60 \\
                                   & RoITr \cite{yu2023rotation}    & 83.00 & \underline{84.40} & 69.40 & \textbf{67.60} \\
\midrule
\multirow{3}{*}{DDPM} & Diff-Reg (Backbone)       & \underline{85.47} & 81.15 & 72.37 & 59.50 \\
                       & Diff-Reg (steps=1)       & 85.23 & 83.85 & \underline{73.19} & 65.26 \\
                       & Diff-Reg (steps=20)      & \textbf{90.25} & \textbf{87.98} & \textbf{77.15} & \textbf{\underline{67.00}} \\
\bottomrule
\end{tabular}
}
\label{tab_4dmatch}
\end{table}

\subsubsection{\textbf{Rigid 3DMatch/3DLoMatch benchmark}}

\noindent\textbf{\quad \quad Datasets.} 3DMatch~\cite{zeng20173dmatch} is an indoor benchmark for 3D matching and registration. Following~\cite{huang2021predator,qin2022geometric,li2022lepard}, we split it into 46/8/8 scenes for training/validation/testing. The overlap ratio between scan pairs in 3DMatch/3DLoMatch is about $>30\%$/$10\%-30\%$.

\noindent\textbf{Metrics.}
Following~\cite{huang2021predator,qin2022geometric,lee2011hyper}, we utilize three evaluation metrics to evaluate our method and other baselines: (1) Inlier Ratio (IR): The proportion of accurate correspondences in which the distance falls below a threshold (i.e., $0.1m$) based on the ground truth transformation. (2) Feature Matching Recall (FMR): The percentage of matched pairs that have an inlier ratio exceeding a specified threshold (i.e., $5\%$). (3) Registration Recall (RR): The fraction of successfully registered point cloud pairs with a predicted transformation error below a certain threshold (e.g., RMSE $<$ 0.2).

\noindent\textbf{Results.} We compared our method with some state-of-the-art feature matching based methods: FCGF~\cite{choy2019fully}, D3Feat~\cite{bai2020d3feat}, Predator~\cite{huang2021predator}, Lepard~\cite{li2022lepard}, GeoTr~\cite{qin2022geometric}, and RoITr~\cite{yu2023rotation}. As demonstrated in Table~\ref{tab.3dmatch}, our method, Diff-Reg, achieves the highest registration recall on the 3DMatch benchmark. The IR and FMR metrics are lower than those of other methods, indicating noisy candidate matching sets. The high RR metric on the 3DMatch benchmark indicates that our method outputs a better-optimized feature backbone capable of recalling highly salient, sparse key matches that were missed by other methods. Subsequently, the RANSAC step can identify these key matches within the noisy candidate sets.

\begin{table}
\caption{Quantitative results on the 3DMatch and 3DLoMatch benchmarks. The best results are highlighted in bold, and the second-best results are underlined.}

\centering
	 \resizebox{0.5\textwidth}{!}{
\begin{tabular}{c|c|ccc|ccc}
\toprule
 \midrule
\multirow{2}{*}{Method}&& \multicolumn{3}{c}{3DMatch} & \multicolumn{3}{c}{3DLoMatch} \\
 &  Reference                                   & FMR$\uparrow$     &  IR$\uparrow$      &   RR$\uparrow$    &  FMR$\uparrow$     &  IR$\uparrow$      &   RR  $\uparrow$    \\

 \midrule
 FCGF&ICCV2019~\cite{choy2019fully}&    95.20   &  56.90     &   88.20   &  60.90     &  21.40     &  45.80    \\
 D3Feat& CVPR2020~\cite{bai2020d3feat}&  95.80    &   39.00    &  85.80     &   69.30    &  13.20     & 40.20     \\
 Predator &  CVPR2021~\cite{huang2021predator}& 96.70    &   58.00    &  91.80    &  78.60     &    26.70   &  62.40    \\
 Lepard&CVPR2022~\cite{li2022lepard}&    97.95   &   57.61    &  93.90    &  84.22    &    27.83  &    70.63  \\
  GeoTR&CVPR2022~\cite{qin2022geometric}&   98.1  &   72.7   &  92.3    &  88.7    &    44.7   &   75.4 \\ 
  RoITr&CVPR2023~\cite{yu2023rotation}&    98.0   &   82.6    &  91.9    &  89.6     &    54.3   &    74.8  \\

  PEAL-3D&CVPR2023~\cite{yu2023peal}& 98.5& 73.3&94.2&  87.6 &49.0&79.0\\

  Diff-Reg&ECCV2024&96.28& 30.92& \textbf{95.0}  &   69.6 &  9.6   & 73.8     \\     

 \bottomrule
\end{tabular}}\label{tab.3dmatch}
\end{table}

\begin{table}[H]

	\centering
	
\caption{Evaluation results on RGB-D Scenes V2\cite{li20232D3D}. The best results are highlighted in bold, and the second-best results are underlined.}
\resizebox{0.5\textwidth}{!}{

\begin{tabular}{c|ccccc}
\toprule
\midrule
Model & Scene-11 &Scene-12 &Scene-13 &Scene-14 &Mean \\
\midrule
Mean depth (m) &1.74& 1.66 &1.18 &1.39 &1.49\\
\midrule

\multicolumn{5}{c}{Registration Recall$\uparrow$}\\
\midrule
FCGF-2D-3D\cite{choy2019fully} &26.4& 41.2& 37.1 &16.8 &30.4\\
P2-Net\cite{Wang2021P2NetJD} &40.3& 40.2& 41.2& 31.9& 38.4\\
Predator-2D-3D\cite{huang2021predator} &44.4 &41.2& 21.6 &13.7& 30.2\\
2D-3D-MATR\cite{li20232D3D}& 63.9 &53.9 &58.8& 49.1& 56.4\\
FreeReg+Kabsch\cite{Wang2023FreeRegIC}& 38.7& 51.6& 30.7& 15.5& 34.1 \\
FreeReg+PnP\cite{Wang2023FreeRegIC}& 74.2& 72.5& 54.5& 27.9& 57.3\\
\midrule

Diff-Reg(dino) &87.5&86.3&63.9&60.6&74.6\\
Diff-Reg(dino/backbone) &79.2&86.3&75.3&\textbf{71.2}&78.0\\

Diff-Reg(dino/steps=1) &\underline{94.4}&\underline{98.0}&85.6&\underline{63.7}&85.4\\

Diff-Reg(dino/steps=10) &\textbf{98.6}&\textbf{99.0}&\underline{86.6}&\underline{63.7}&\underline{87.0}\\
\midrule
Diff-Reg(dino/backbone$_{epnp}$) &95.8&96.1&\textbf{88.7}&69.0&\textbf{87.4}\\

\bottomrule
\end{tabular}
}

\label{tab:non_rigid_registration}

% \vspace{-0.7cm}

\end{table}

\noindent\subsection{2D-3D Registration Task}\label{design_2D-3D_task}

We utilize the coarse-level circle loss\cite{li20232D3D} and fine-level matching loss\cite{li20232D3D} for the single-pass backbone\cite{li20232D3D} and a focal loss for our denoising module $g_{\theta}$. The framework is trained and evaluated with PyTorch on one NVIDIA RTX 3090 GPU. We train our model about 30 epochs with batch size 1.

\noindent\textbf{Datasets.}
RGB-D Scenes V2\cite{lai2014unsupervised} is generated from 14 indoor scenes with 11427 RGB-D frames. Following\cite{li20232D3D}, we split 14 sequences of image-to-point-clouds pairs data, where scenes $0-8/9-10/11-14$ for training/validation/testing. The resulting dataset contains $1748/236/497$ image-to-point-cloud pairs.  

7-Scenes\cite{shotton2013scene} is generated from 7 indoor scenes with 46 RGB-D sequences. Following~\cite{li20232D3D}, we split these sequences to 4048/1011/2304 image-to-point-cloud pairs for training/validation/testing. 

\noindent\textbf{Metrics.}
In this section, following~\cite{li20232D3D}, we give a detailed definition of three evaluation protocols: (1) Inlier Ratio (IR), the ratio of pixel-point matches whose 3D distance is under a certain threshold (i.e., 5cm). (2) Feature Matching Recall (FMR), the ratio of image-to-point-cloud pairs whose inlier ratio is above a certain threshold (i.e., 10$\%$). (3) Registration Recall (RR), the ratio of image-to-point-cloud pairs whose RMSE is under a certain threshold (i.e., 10cm).

% \vspace{1mm}
\noindent\textbf{Quantitative results.}
Following~\cite{li20232D3D}, we compare our diffusion matching model with FCGF~\cite{choy2019fully}, P2-Net~\cite{Wang2021P2NetJD}, Predator~\cite{huang2021predator}, 2D3D-MATR~\cite{li20232D3D}, and FreeReg~\cite{Wang2023FreeRegIC}. We introduce a robust single-step baseline, Diff-Reg(dino), where we integrate the visual feature backbone DINOv2~\cite{oquab2023dinov2} into the single-step model 2D-3D-MATR~\cite{li20232D3D}. Diff-Reg(dino/backbone) represents the single-step prediction head derived from Diff-Reg(dino) after joint training with our denoising module $g_{\theta}$. Diff-Reg(steps=1) and Diff-Reg(steps=10) refer to our diffusion matching model with one step and ten steps of reverse denoising sampling, respectively. For all variants of Diff-Reg(*), we drop the three scales 6 × 8, 12 × 16, 24 × 32 (refer to Section 4.1 in~\cite{li20232D3D}) after the coarsest level of ResNet and preserve only the 24 × 32 resolution.

As shown in Table~\ref{tab:non_rigid_registration}, Diff-Reg(dino/backbone)'s results indicate that the diffused training samples in the matrix space serve as data augmentation, which enhances the representation of the ResNet feature backbone and single-step prediction head. Additionally, the outcome for Diff-Reg(dino/steps=10) demonstrates that our denoising module $g_{\theta}$ effectively addresses the scale ambiguity issue in 2D-3D registration. Moreover, the result for Diff-Reg(dino/steps=1) reveals that the diffused training samples within the matrix space improve the single-step prediction head in Diff-Reg(dino).

We also conducted an additional experiment to show that the performance improvement of Diff-Reg(dino/backbone) over Diff-Reg(dino) is due to our diffusion matching model. In this case, the image depth map $D^{\hat{X}}$ is not a prerequisite. We input only superpixels and superpoints, with weights computed from the associated superpixel and superpoint features. Then, we employ the EPnP solver~\cite{li2012robust} instead of weighted SVD to compute the transformation $R, t$, which we denote as Diff-PnP. We use Diff-PnP(dino/backbone) to denote the single-pass prediction head that is jointly trained with our denoising module $g_\theta$. Diff-PnP(dino/steps=*) follows the same definition as Diff-Reg(dino/steps=*). We evaluate our Diff-PnP method on the 7Scenes~\cite{shotton2013scene} dataset. The results in Table~\ref{tab:2D-3D_7scenes} show that our method achieves competitive performance.

\begin{table}[H]
\centering
	
\caption{Evaluation results on 7Scenes~\cite{shotton2013scene}. The best results are highlighted in bold, and the second-best results are underlined.}
\resizebox{0.5\textwidth}{!}{

\begin{tabular}{c|cccccccc}
\toprule
\midrule
Model & Chess & Fire &Heads &Office &Pumpkin &Kitchen &Stairs &Mean \\
\midrule
Mean depth (m) &1.78 &1.55& 0.80& 2.03& 2.25& 2.13& 1.84& 1.77\\
\midrule
\multicolumn{9}{c}{Inlier Ratio$\uparrow$}\\
\midrule
FCGF-2D-3D~\cite{choy2019fully}&34.2& 32.8& 14.8& 26.0& 23.3& 22.5& 6.0 &22.8\\
P2-Net~\cite{choy2019fully}&55.2 &46.7& 13.0& 36.2& 32.0 &32.8& 5.8& 31.7 \\
Predator-2D-3D~\cite{huang2021predator}&34.7& 33.8 &16.6& 25.9 &23.1 &22.2 &7.5 &23.4 \\
2D-3D-MATR~\cite{li20232D3D} &72.1& 66.0& 31.3& 60.7& 50.2& 52.5& 18.1& 50.1\\
\midrule
Diff-PnP(dino/backbone)&79.2&71.0&54.1&70.4&55.8&60.2&22.9&59.1\\
Diff-PnP(dino/steps=10)&73.3&60.8&45.5&63.1&47.8&53.3&20.4&52.0\\

\midrule

\multicolumn{9}{c}{Feature Matching Recall$\uparrow$}\\
\midrule
FCGF-2D-3D~\cite{choy2019fully}&99.7& 98.2& 69.9& 97.1& 83.0& 87.7& 16.2& 78.8 \\
P2-Net~\cite{choy2019fully}&100.0& 99.3& 58.9& 99.1& 87.2& 92.2& 16.2& 79.0 \\
Predator-2D-3D~\cite{huang2021predator}&91.3 &95.1& 76.7& 88.6& 79.2& 80.6& 31.1& 77.5 \\
2D-3D-MATR~\cite{li20232D3D}&100.0&99.6& 98.6& 100.0& 92.4& 95.9& 58.1& 92.1\\

\midrule
Diff-PnP(dino/backbone)&100.0&100.0&100.0&100.0&91.3&98.1&58.1&92.5\\
Diff-PnP(dino/steps=10)&100.0&98.5&97.3&100.0&87.8&96.8&60.8&91.6\\

\midrule

\multicolumn{9}{c}{Registration Recall$\uparrow$}\\
\midrule
FCGF-2D-3D~\cite{choy2019fully} &89.5 &79.7 &19.2& 85.9& 69.4& 79.0& 6.8& 61.4\\
P2-Net~\cite{choy2019fully} &96.9& 86.5 &20.5& 91.7& 75.3& 85.2& 4.1& 65.7\\
Predator-2D-3D~\cite{huang2021predator} &69.6& 60.7& 17.8 &62.9& 56.2& 62.6& 9.5 &48.5\\
2D-3D-MATR~\cite{li20232D3D}&96.9& 90.7 &52.1& 95.5& 80.9& 86.1& 28.4& 75.8\\
\midrule
Diff-PnP(dino/backbone)&100.0&94.0&90.4&99.3&81.2&94.6&27.0&83.8\\
Diff-PnP(dino/steps=10) &99.3&94.3&91.8&99.1&79.9&91.8&25.7&83.1\\ 

\bottomrule
\end{tabular}
}

\label{tab:2D-3D_7scenes}
\end{table}

\begin{figure*}
    \centering  
    
   \begin{minipage}{\textwidth}
    \quad  % 插入一些空格以将图片分开
    \subfloat[]{%
        \includegraphics[width=0.3\textwidth, height=1.8cm]{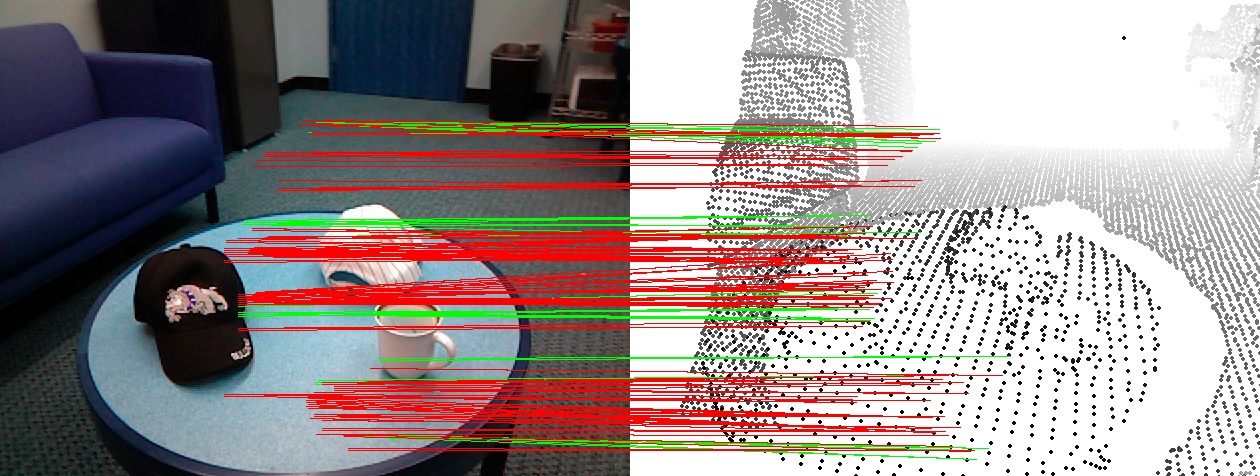} 
    }
    \hspace{0.01\textwidth}
    \subfloat[]{%
        \includegraphics[width=0.3\textwidth, height=1.8cm]{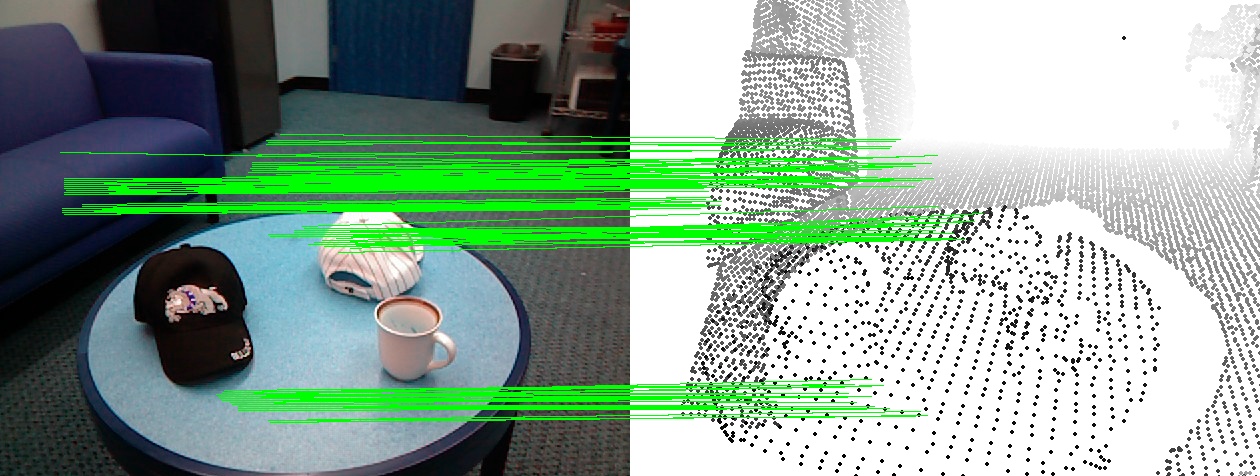} 
         \hspace{0.01\linewidth}
        \includegraphics[width=0.3\textwidth, height=1.8cm]{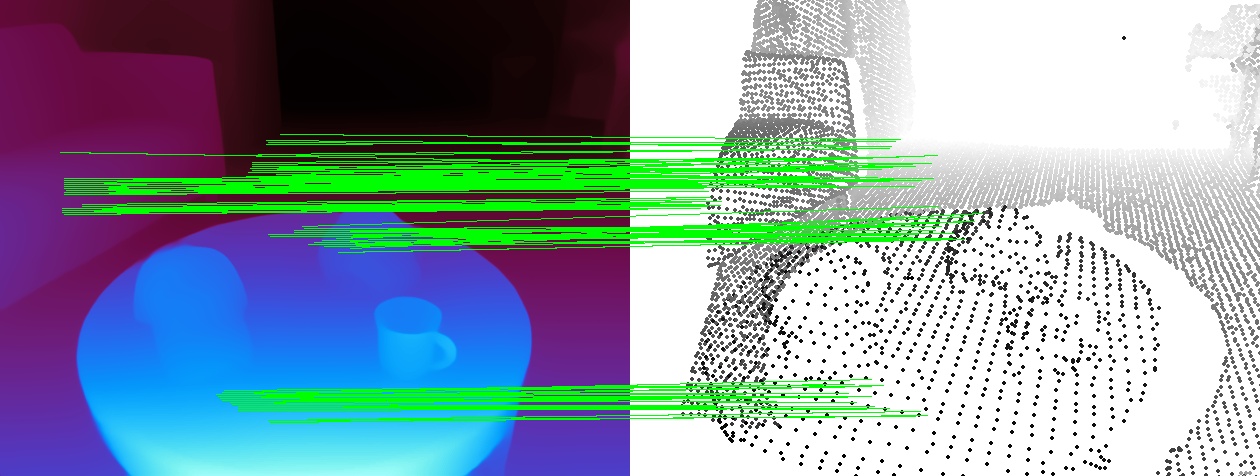}%
        \label{fig:sub5d}
    }
    \end{minipage}

   \begin{minipage}{\textwidth}
    \quad  % 插入一些空格以将图片分开
    \subfloat[]{%
        \includegraphics[width=0.3\textwidth, height=1.8cm]{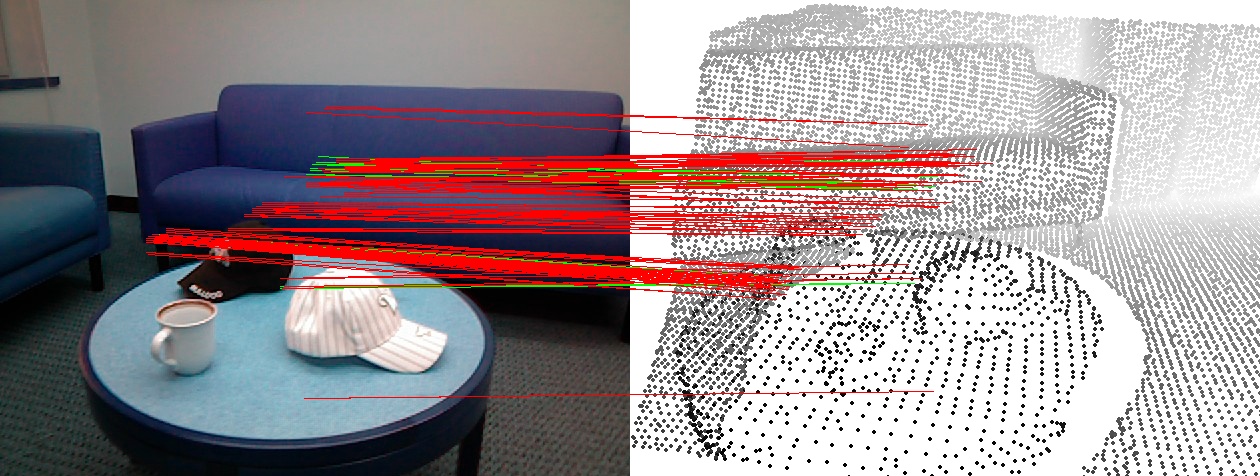} 
    }
    \hspace{0.01\textwidth}
    \subfloat[]{%
        \includegraphics[width=0.3\textwidth, height=1.8cm]{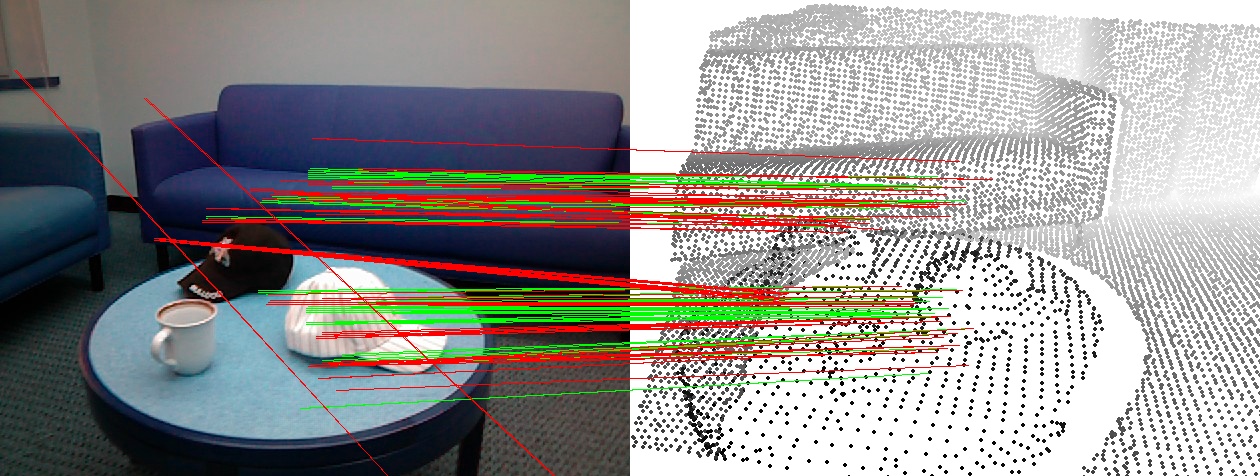} 
         \hspace{0.01\linewidth}
        \includegraphics[width=0.3\textwidth, height=1.8cm]{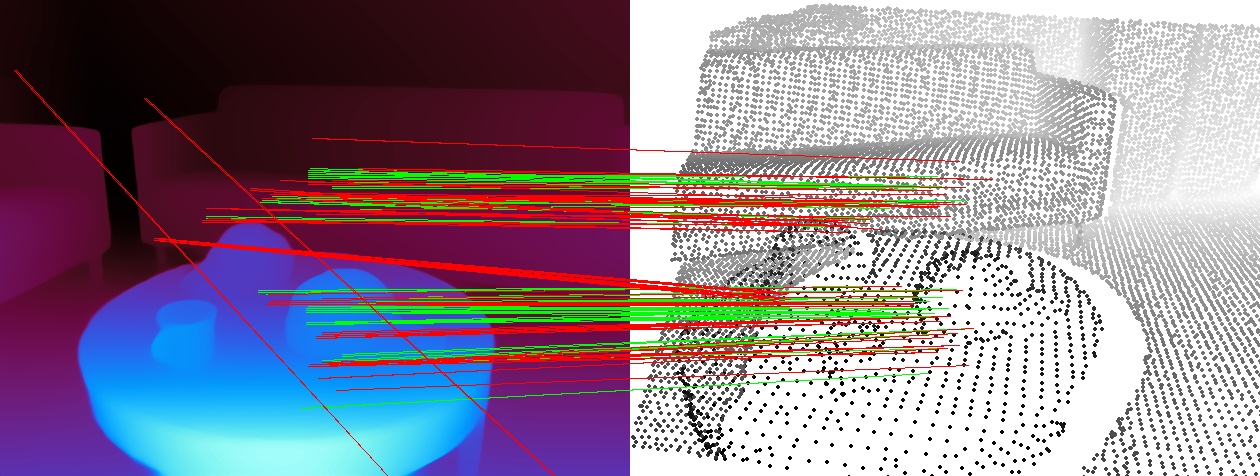}%
        \label{fig:sub5e}
    }
    \end{minipage}

  \begin{minipage}{\textwidth}
    \quad  % 插入一些空格以将图片分开
    \subfloat[]{%
        \includegraphics[width=0.3\textwidth, height=1.8cm]{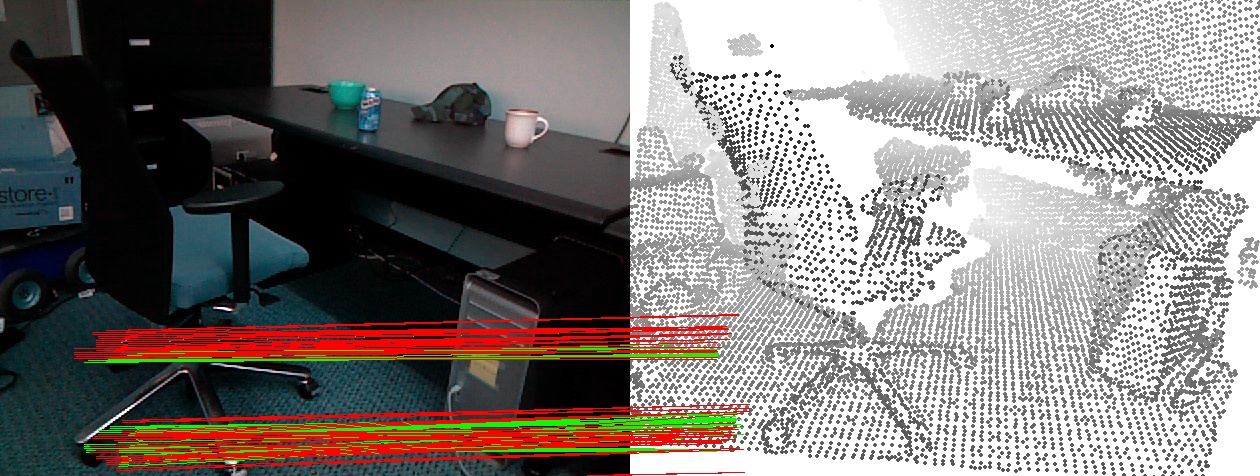} 
    }
    \hspace{0.01\textwidth}
    \subfloat[]{%
        \includegraphics[width=0.3\textwidth, height=1.8cm]{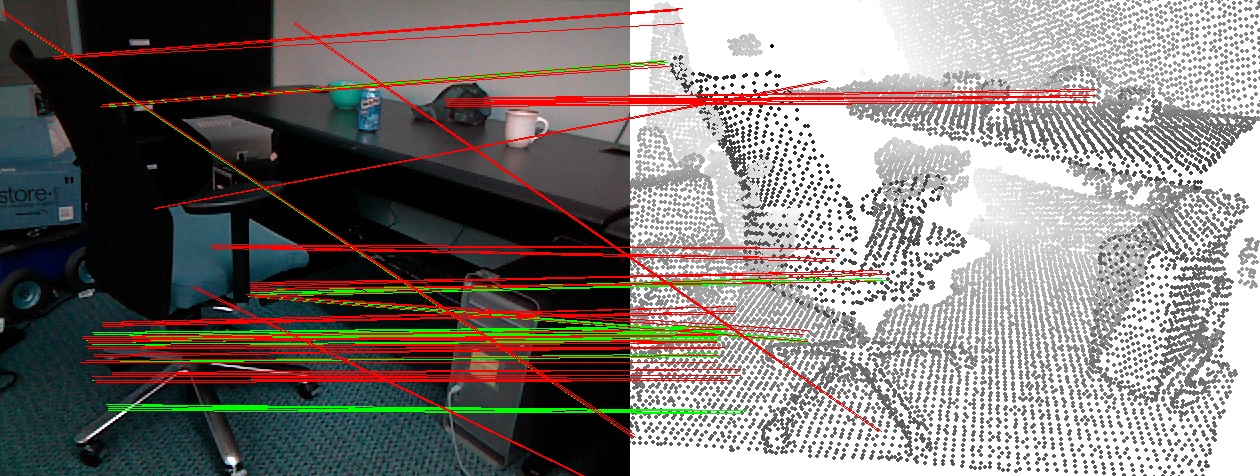} 
         \hspace{0.01\linewidth}
        \includegraphics[width=0.3\textwidth, height=1.8cm]{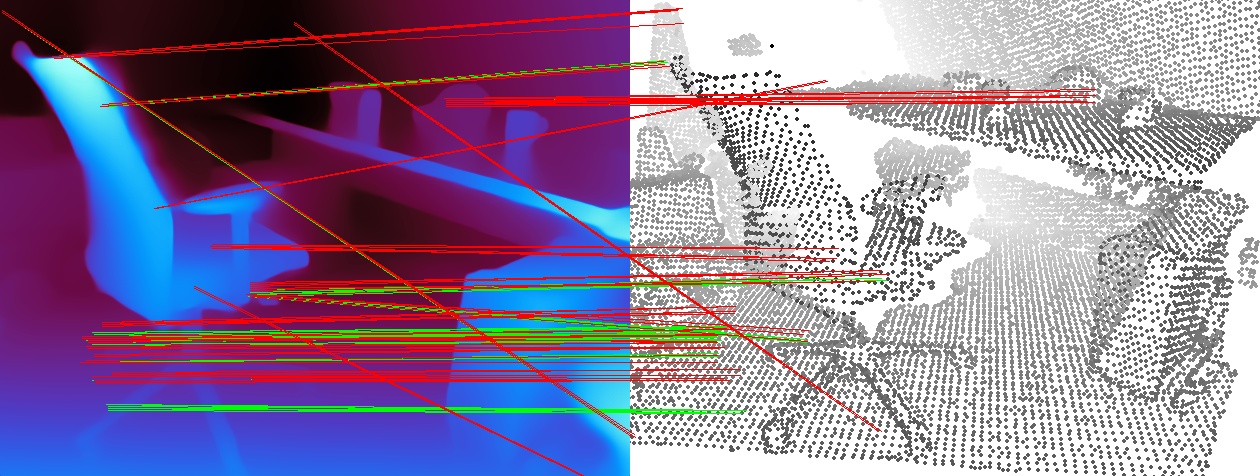}%
        \label{fig:sub5f}
    }
    \end{minipage}

  \begin{minipage}{\textwidth}
    \quad  % 插入一些空格以将图片分开
    \subfloat[Diff-Reg(dino)]{%
        \includegraphics[width=0.3\textwidth, height=1.8cm]{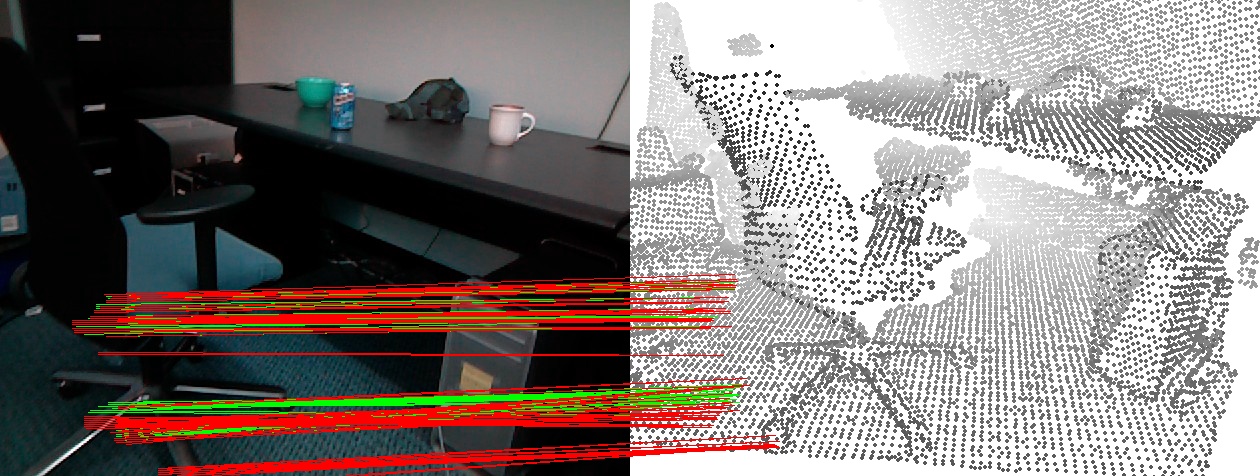} 
    }
    \hspace{0.01\textwidth}
    \subfloat[Diff-Reg(dino/steps=10)]{%
        \includegraphics[width=0.3\textwidth, height=1.8cm]{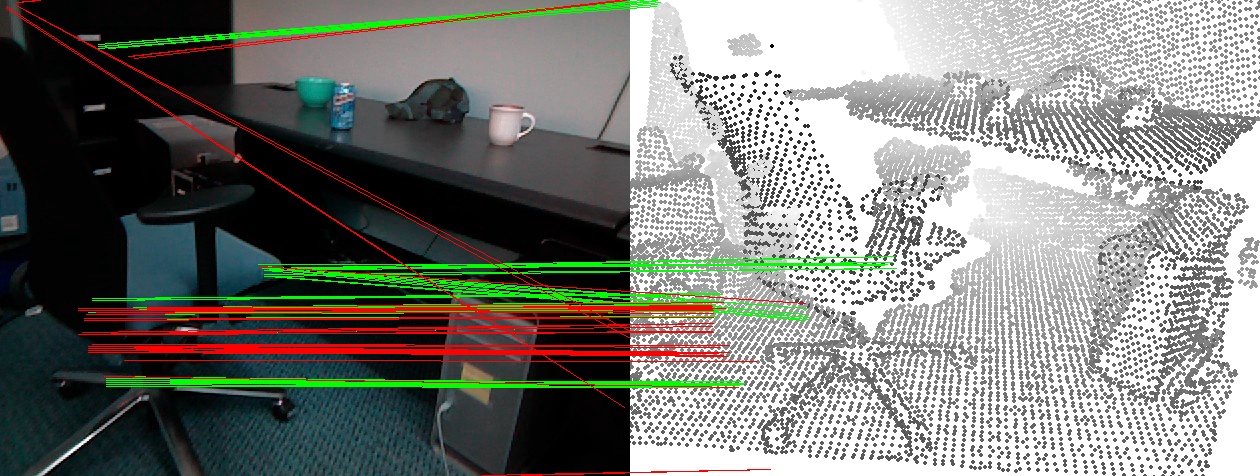} 
         \hspace{0.01\linewidth}
        \includegraphics[width=0.3\textwidth, height=1.8cm]{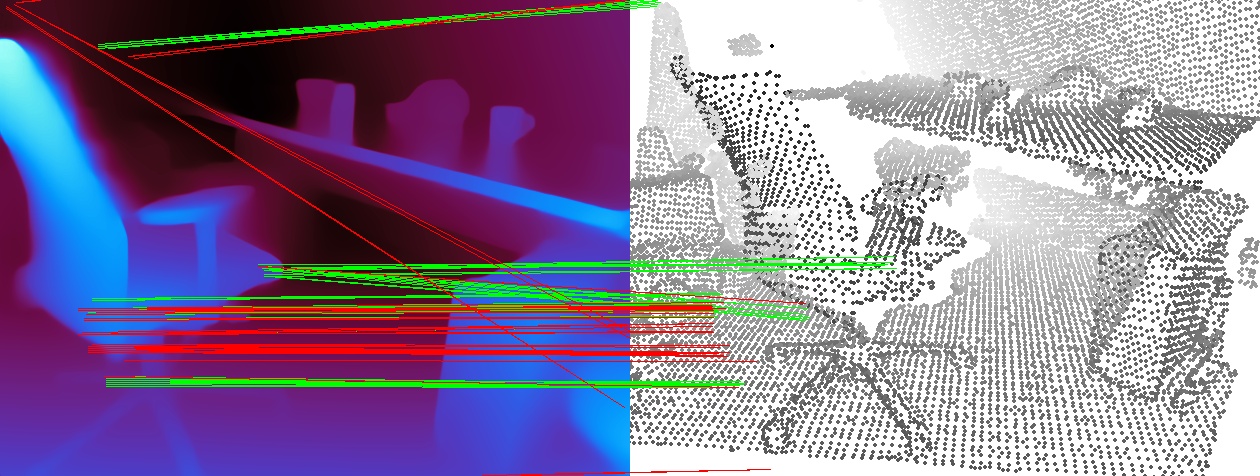}%
        \label{fig:sub5g}
    }
    \end{minipage}

    \caption{The qualitative results of top-200 predicted correspondences on the RGB-D Scenes V2 benchmark\cite{lai2014unsupervised}. The \textcolor{green}{green}/\textcolor{red}{red} color indicates whether the matching score is accepted based on a threshold value. Zoom in for details.}
    \label{2D-3D_non_rigid_registration_vis}
\end{figure*}

\noindent\textbf{Qualitative results.}
The examples of Diff-Reg(dino/steps=10) in Fig. \ref{2D-3D_non_rigid_registration_vis} reveal that our diffusion matching model excels at capturing salient correspondences crucial for combinatorial consistency, regardless of their distance from the camera. On the other hand, the matches generated by Diff-Reg(dino) tend to be more focused at specific distances (from camera). For instance, in the third row, the correspondences produced by Diff-Reg(dino) are located very close to the camera, while the correspondences from Diff-Reg(dino/steps=10) encompass objects such as hats on the black table that are situated at a greater distance. In the first row, Diff-Reg(dino) fails to capture the correspondences on the sofa, and in the second row, the correspondences of the white hat on the table are lost. An extreme case in the fourth row demonstrates that Diff-Reg(dino) misses the farthest correspondence on the column bookshelf or wall. 

% \vspace{-0.5cm}
\section{Conclusion}
In this paper, we present a novel approach by deploying the diffusion model in the matrix space for iteratively exploring the optimal matching matrix through the reverse denoising sampling process. We have successfully implemented the matching matrix encoding module and the denoising module in a unified manner, applicable to both 2D image matching and 3D registration tasks. Additionally, the lightweight design of our reverse denoising module enables faster convergence during the reverse sampling process. Through extensive experiments on real-world datasets, we demonstrate the effectiveness of our method in 3D registration, 2D-3D registration, and 2D image matching tasks.

\bibliographystyle{IEEEtran} 
\bibliography{jrnl}

\begin{thebibliography}{100}
\providecommand{\url}[1]{#1}
\csname url@samestyle\endcsname
\providecommand{\newblock}{\relax}
\providecommand{\bibinfo}[2]{#2}
\providecommand{\BIBentrySTDinterwordspacing}{\spaceskip=0pt\relax}
\providecommand{\BIBentryALTinterwordstretchfactor}{4}
\providecommand{\BIBentryALTinterwordspacing}{\spaceskip=\fontdimen2\font plus
\BIBentryALTinterwordstretchfactor\fontdimen3\font minus
  \fontdimen4\font\relax}
\providecommand{\BIBforeignlanguage}[2]{{%
\expandafter\ifx\csname l@#1\endcsname\relax
\typeout{** WARNING: IEEEtran.bst: No hyphenation pattern has been}%
\typeout{** loaded for the language `#1'. Using the pattern for}%
\typeout{** the default language instead.}%
\else
\language=\csname l@#1\endcsname
\fi
#2}}
\providecommand{\BIBdecl}{\relax}
\BIBdecl

\bibitem{sun2021loftr}
J.~Sun, Z.~Shen, Y.~Wang, H.~Bao, and X.~Zhou, ``Loftr: Detector-free local
  feature matching with transformers,'' in \emph{CVPR}, 2021.

\bibitem{edstedt2024roma}
J.~Edstedt, Q.~Sun, G.~B{\"o}kman, M.~Wadenb{\"a}ck, and M.~Felsberg, ``Roma:
  Robust dense feature matching,'' in \emph{Proceedings of the IEEE/CVF
  Conference on Computer Vision and Pattern Recognition}, 2024, pp.
  19\,790--19\,800.

\bibitem{zhong2009intrinsic}
Y.~Zhong, ``Intrinsic shape signatures: A shape descriptor for 3d object
  recognition,'' in \emph{2009 IEEE 12th international conference on computer
  vision workshops, ICCV workshops}, 2009.

\bibitem{yu2023rotation}
H.~Yu, Z.~Qin, J.~Hou, M.~Saleh, D.~Li, B.~Busam, and S.~Ilic,
  ``Rotation-invariant transformer for point cloud matching,'' in \emph{CVPR},
  2023.

\bibitem{bai2020d3feat}
X.~Bai, Z.~Luo, L.~Zhou, H.~Fu, L.~Quan, and C.-L. Tai, ``D3feat: Joint
  learning of dense detection and description of 3d local features,'' in
  \emph{CVPR}, 2020.

\bibitem{giang2023topicfm}
K.~T. Giang, S.~Song, and S.~Jo, ``Topicfm: Robust and interpretable
  topic-assisted feature matching,'' in \emph{Proceedings of the AAAI
  conference on artificial intelligence}, vol.~37, no.~2, 2023, pp. 2447--2455.

\bibitem{li20232D3D}
M.~Li, Z.~Qin, Z.~Gao, R.~Yi, C.~Zhu, Y.~Guo, and K.~Xu, ``2d3d-matr: 2d-3d
  matching transformer for detection-free registration between images and point
  clouds,'' in \emph{Proceedings of the IEEE/CVF International Conference on
  Computer Vision}, 2023.

\bibitem{deng2018ppfnet}
H.~\vspace{0mm}Deng, T.~Birdal, and S.~Ilic, ``Ppfnet: Global context aware
  local features for robust 3d point matching,'' in \emph{Proceedings of the
  IEEE conference on computer vision and pattern recognition}, 2018.

\bibitem{qin2022geometric}
Z.~Qin, H.~Yu, C.~Wang, Y.~Guo, Y.~Peng, and K.~Xu, ``Geometric transformer for
  fast and robust point cloud registration,'' in \emph{Proceedings of the
  IEEE/CVF Conference on Computer Vision and Pattern Recognition}, 2022.

\bibitem{wu2023sgfeat}
Q.~Wu, Y.~Ding, L.~Luo, C.~Zhou, J.~Xie, and J.~Yang, ``Sgfeat: Salient
  geometric feature for point cloud registration,'' \emph{arXiv preprint
  arXiv:2309.06207}, 2023.

\bibitem{li2022lepard}
Y.~Li and T.~Harada, ``Lepard: Learning partial point cloud matching in rigid
  and deformable scenes,'' in \emph{Proceedings of the IEEE/CVF Conference on
  Computer Vision and Pattern Recognition}, 2022.

\bibitem{yew2022regtr}
Z.~J. Yew and G.~H. Lee, ``Regtr: End-to-end point cloud correspondences with
  transformers,'' in \emph{Proceedings of the IEEE/CVF Conference on Computer
  Vision and Pattern Recognition}, 2022, pp. 6677--6686.

\bibitem{mei2023unsupervised}
G.~Mei, H.~Tang, X.~Huang, W.~Wang, J.~Liu, J.~Zhang, L.~Van~Gool, and Q.~Wu,
  ``Unsupervised deep probabilistic approach for partial point cloud
  registration,'' in \emph{Proceedings of the IEEE/CVF Conference on Computer
  Vision and Pattern Recognition}, 2023.

\bibitem{huang2021predator}
S.~Huang, Z.~Gojcic, M.~Usvyatsov, A.~Wieser, and K.~Schindler, ``Predator:
  Registration of 3d point clouds with low overlap,'' in \emph{Proceedings of
  the IEEE/CVF Conference on computer vision and pattern recognition}, 2021.

\bibitem{yu2021cofinet}
H.~Yu, F.~Li, M.~Saleh, B.~Busam, and S.~Ilic, ``Cofinet: Reliable
  coarse-to-fine correspondences for robust pointcloud registration,''
  \emph{Advances in Neural Information Processing Systems}, 2021.

\bibitem{yao2023hunter}
R.~Yao, S.~Du, W.~Cui, A.~Ye, F.~Wen, H.~Zhang, Z.~Tian, and Y.~Gao, ``Hunter:
  Exploring high-order consistency for point cloud registration with severe
  outliers,'' \emph{IEEE Transactions on Pattern Analysis and Machine
  Intelligence}, 2023.

\bibitem{chen2022aspanformer}
H.~Chen, Z.~Luo, L.~Zhou, Y.~Tian, M.~Zhen, T.~Fang, D.~Mckinnon, Y.~Tsin, and
  L.~Quan, ``Aspanformer: Detector-free image matching with adaptive span
  transformer,'' in \emph{European Conference on Computer Vision}.\hskip 1em
  plus 0.5em minus 0.4em\relax Springer, 2022, pp. 20--36.

\bibitem{thomas2019kpconv}
H.~Thomas, C.~R. Qi, J.-E. Deschaud, B.~Marcotegui, F.~Goulette, and L.~J.
  Guibas, ``Kpconv: Flexible and deformable convolution for point clouds,'' in
  \emph{Proceedings of the IEEE/CVF international conference on computer
  vision}, 2019.

\bibitem{he2016deep}
K.~He, X.~Zhang, S.~Ren, and J.~Sun, ``Deep residual learning for image
  recognition,'' in \emph{Proceedings of the IEEE conference on computer vision
  and pattern recognition}, 2016.

\bibitem{ding2021repvgg}
X.~Ding, X.~Zhang, N.~Ma, J.~Han, G.~Ding, and J.~Sun, ``Repvgg: Making
  vgg-style convnets great again,'' in \emph{Proceedings of the IEEE/CVF
  conference on computer vision and pattern recognition}, 2021, pp.
  13\,733--13\,742.

\bibitem{yang2020teaser}
H.~Yang, J.~Shi, and L.~Carlone, ``Teaser: Fast and certifiable point cloud
  registration,'' \emph{IEEE Transactions on Robotics}, vol.~37, no.~2, pp.
  314--333, 2020.

\bibitem{bai2021pointdsc}
X.~Bai, Z.~Luo, L.~Zhou, H.~Chen, L.~Li, Z.~Hu, H.~Fu, and C.-L. Tai,
  ``Pointdsc: Robust point cloud registration using deep spatial consistency,''
  in \emph{Proceedings of the IEEE/CVF Conference on Computer Vision and
  Pattern Recognition}, 2021.

\bibitem{chen2022sc2}
Z.~Chen, K.~Sun, F.~Yang, and W.~Tao, ``Sc2-pcr: A second order spatial
  compatibility for efficient and robust point cloud registration,'' in
  \emph{Proceedings of the IEEE/CVF Conference on Computer Vision and Pattern
  Recognition}, 2022.

\bibitem{jiang2023robust}
H.~Jiang, Z.~Dang, Z.~Wei, J.~Xie, J.~Yang, and M.~Salzmann, ``Robust outlier
  rejection for 3d registration with variational bayes,'' in \emph{Proceedings
  of the IEEE/CVF Conference on Computer Vision and Pattern Recognition}, 2023.

\bibitem{zhang20233dmac}
X.~Zhang, J.~Yang, S.~Zhang, and Y.~Zhang, ``3d registration with maximal
  cliques,'' in \emph{Proceedings of the IEEE/CVF Conference on Computer Vision
  and Pattern Recognition}, 2023.

\bibitem{xue2023imp}
F.~Xue, I.~Budvytis, and R.~Cipolla, ``Imp: Iterative matching and pose
  estimation with adaptive pooling,'' in \emph{Proceedings of the IEEE/CVF
  Conference on Computer Vision and Pattern Recognition}, 2023, pp.
  21\,317--21\,326.

\bibitem{sarlin2020superglue}
P.-E. Sarlin, D.~DeTone, T.~Malisiewicz, and A.~Rabinovich, ``Superglue:
  Learning feature matching with graph neural networks,'' in \emph{Proceedings
  of the IEEE/CVF conference on computer vision and pattern recognition}, 2020,
  pp. 4938--4947.

\bibitem{lindenberger2023lightglue}
P.~Lindenberger, P.-E. Sarlin, and M.~Pollefeys, ``Lightglue: Local feature
  matching at light speed,'' in \emph{Proceedings of the IEEE/CVF International
  Conference on Computer Vision}, 2023, pp. 17\,627--17\,638.

\bibitem{qi2019deep}
C.~R. Qi, O.~Litany, K.~He, and L.~J. Guibas, ``Deep hough voting for 3d object
  detection in point clouds,'' in \emph{proceedings of the IEEE/CVF
  International Conference on Computer Vision}, 2019.

\bibitem{fu2021robust}
K.~Fu, S.~Liu, X.~Luo, and M.~Wang, ``Robust point cloud registration framework
  based on deep graph matching,'' in \emph{CVPR}, 2021.

\bibitem{yan2024tri}
Z.~Yan, Y.~Lin, K.~Wang, Y.~Zheng, Y.~Wang, Z.~Zhang, J.~Li, and J.~Yang,
  ``Tri-perspective view decomposition for geometry-aware depth completion,''
  in \emph{Proceedings of the IEEE/CVF Conference on Computer Vision and
  Pattern Recognition}, 2024, pp. 4874--4884.

\bibitem{yan2023desnet}
Z.~Yan, K.~Wang, X.~Li, Z.~Zhang, J.~Li, and J.~Yang, ``Desnet: Decomposed
  scale-consistent network for unsupervised depth completion,'' in
  \emph{Proceedings of the AAAI Conference on Artificial Intelligence},
  vol.~37, 2023, pp. 3109--3117.

\bibitem{yan2022learning}
Z.~Yan, K.~Wang, X.~Li, Z.~Zhang, G.~Li, J.~Li, and J.~Yang, ``Learning
  complementary correlations for depth super-resolution with incomplete data in
  real world,'' \emph{IEEE transactions on neural networks and learning
  systems}, 2022.

\bibitem{yan2022rignet}
Z.~Yan, K.~Wang, X.~Li, Z.~Zhang, J.~Li, and J.~Yang, ``Rignet: Repetitive
  image guided network for depth completion,'' in \emph{European Conference on
  Computer Vision}.\hskip 1em plus 0.5em minus 0.4em\relax Springer, 2022, pp.
  214--230.

\bibitem{bokman2025affine}
G.~B{\"o}kman, J.~Edstedt, M.~Felsberg, and F.~Kahl, ``Affine steerers for
  structured keypoint description,'' in \emph{European Conference on Computer
  Vision}.\hskip 1em plus 0.5em minus 0.4em\relax Springer, 2025, pp. 449--468.

\bibitem{urain2023se}
J.~Urain, N.~Funk, J.~Peters, and G.~Chalvatzaki, ``Se (3)-diffusionfields:
  Learning smooth cost functions for joint grasp and motion optimization
  through diffusion,'' in \emph{2023 IEEE International Conference on Robotics
  and Automation (ICRA)}, 2023.

\bibitem{jiang2023se}
H.~Jiang, M.~Salzmann, Z.~Dang, J.~Xie, and J.~Yang, ``Se (3) diffusion
  model-based point cloud registration for robust 6d object pose estimation,''
  \emph{arXiv preprint arXiv:2310.17359}, 2023.

\bibitem{chen2023diffusionpcr}
Z.~Chen, Y.~Ren, T.~Zhang, Z.~Dang, W.~Tao, S.~S{\"u}sstrunk, and M.~Salzmann,
  ``Diffusionpcr: Diffusion models for robust multi-step point cloud
  registration,'' \emph{arXiv preprint arXiv:2312.03053}, 2023.

\bibitem{zhang2024diffsf}
Y.~Zhang, B.~Wandt, M.~Magnusson, and M.~Felsberg, ``Diffsf: Diffusion models
  for scene flow estimation,'' \emph{arXiv preprint arXiv:2403.05327}, 2024.

\bibitem{nam2023diffusion}
J.~Nam, G.~Lee, S.~Kim, H.~Kim, H.~Cho, S.~Kim, and S.~Kim, ``Diffusion model
  for dense matching,'' \emph{arXiv preprint arXiv:2305.19094}, 2023.

\bibitem{wang2024efficient}
Y.~Wang, X.~He, S.~Peng, D.~Tan, and X.~Zhou, ``Efficient loftr: Semi-dense
  local feature matching with sparse-like speed,'' in \emph{Proceedings of the
  IEEE/CVF Conference on Computer Vision and Pattern Recognition}, 2024, pp.
  21\,666--21\,675.

\bibitem{ho2020denoising}
J.~Ho, A.~Jain, and P.~Abbeel, ``Denoising diffusion probabilistic models,''
  \emph{Advances in neural information processing systems}, 2020.

\bibitem{song2020denoising}
J.~Song, C.~Meng, and S.~Ermon, ``Denoising diffusion implicit models,''
  \emph{arXiv preprint arXiv:2010.02502}, 2020.

\bibitem{vignac2022digress}
C.~Vignac, I.~Krawczuk, A.~Siraudin, B.~Wang, V.~Cevher, and P.~Frossard,
  ``Digress: Discrete denoising diffusion for graph generation,'' \emph{arXiv
  preprint arXiv:2209.14734}, 2022.

\bibitem{austin2021structured}
J.~Austin, D.~D. Johnson, J.~Ho, D.~Tarlow, and R.~Van Den~Berg, ``Structured
  denoising diffusion models in discrete state-spaces,'' \emph{Advances in
  Neural Information Processing Systems}, vol.~34, pp. 17\,981--17\,993, 2021.

\bibitem{xie2020fast}
Y.~Xie, X.~Wang, R.~Wang, and H.~Zha, ``A fast proximal point method for
  computing exact wasserstein distance,'' in \emph{Uncertainty in artificial
  intelligence}.\hskip 1em plus 0.5em minus 0.4em\relax PMLR, 2020, pp.
  433--453.

\bibitem{leordeanu2009integer}
M.~Leordeanu, M.~Hebert, and R.~Sukthankar, ``An integer projected fixed point
  method for graph matching and map inference,'' \emph{Advances in neural
  information processing systems}, vol.~22, 2009.

\bibitem{zaslavskiy2008path}
M.~Zaslavskiy, F.~Bach, and J.-P. Vert, ``A path following algorithm for the
  graph matching problem,'' \emph{IEEE Transactions on Pattern Analysis and
  Machine Intelligence}, vol.~31, no.~12, pp. 2227--2242, 2008.

\bibitem{86bd1ad6-50bb-38d0-9978-0966b4dfc6d3}
\BIBentryALTinterwordspacing
R.~M. Caron, X.~Li, P.~Mikusiński, H.~Sherwood, and M.~D. Taylor, ``Nonsquare
  "doubly stochastic" matrices,'' \emph{Lecture Notes-Monograph Series},
  vol.~28, pp. 65--75, 1996. [Online]. Available:
  \url{http://www.jstor.org/stable/4355884}
\BIBentrySTDinterwordspacing

\bibitem{wugraph}
Q.~Wu, Y.~Shen, H.~Jiang, G.~Mei, Y.~Ding, L.~Luo, J.~Xie, and J.~Yang, ``Graph
  matching optimization network for point cloud registration,'' in \emph{2023
  IEEE/RSJ International Conference on Intelligent Robots and Systems (IROS)},
  2023.

\bibitem{mei2021cotreg}
G.~Mei, X.~Huang, L.~Yu, J.~Zhang, and M.~Bennamoun, ``Cotreg: Coupled optimal
  transport based point cloud registration,'' \emph{arXiv preprint
  arXiv:2112.14381}, 2021.

\bibitem{edstedt2023dkm}
J.~Edstedt, I.~Athanasiadis, M.~Wadenb{\"a}ck, and M.~Felsberg, ``Dkm: Dense
  kernelized feature matching for geometry estimation,'' in \emph{Proceedings
  of the IEEE/CVF Conference on Computer Vision and Pattern Recognition}, 2023,
  pp. 17\,765--17\,775.

\bibitem{tang2022quadtree}
S.~Tang, J.~Zhang, S.~Zhu, and P.~Tan, ``Quadtree attention for vision
  transformers,'' \emph{arXiv preprint arXiv:2201.02767}, 2022.

\bibitem{Guan_IJCV}
B.~Guan, J.~Zhao, D.~Barath, and F.~Fraundorfer, ``Minimal solvers for relative
  pose estimation of multi-camera systems using affine correspondences,''
  \emph{International Journal of Computer Vision}, vol. 131, no.~1, pp.
  324--345, 2023.

\bibitem{Guan_TCYB}
B.~Guan, J.~Zhao, Z.~Li, F.~Sun, and F.~Fraundorfer, ``Relative pose estimation
  with a single affine correspondence,'' \emph{IEEE Transactions on
  Cybernetics}, vol.~52, no.~10, pp. 10\,111--10\,122, 2022.

\bibitem{li2020dual}
X.~Li, K.~Han, S.~Li, and V.~Prisacariu, ``Dual-resolution correspondence
  networks,'' \emph{Advances in Neural Information Processing Systems},
  vol.~33, pp. 17\,346--17\,357, 2020.

\bibitem{cuturi2013sinkhorn}
M.~Cuturi, ``Sinkhorn distances: Lightspeed computation of optimal transport,''
  \emph{Advances in neural information processing systems}, 2013.

\bibitem{li2022non}
Y.~Li and T.~Harada, ``Non-rigid point cloud registration with neural
  deformation pyramid,'' \emph{Advances in Neural Information Processing
  Systems}, 2022.

\bibitem{zeng20173dmatch}
A.~Zeng, S.~Song, M.~Nie{\ss}ner, M.~Fisher, J.~Xiao, and T.~Funkhouser,
  ``3dmatch: Learning local geometric descriptors from rgb-d reconstructions,''
  in \emph{Proceedings of the IEEE conference on computer vision and pattern
  recognition}, 2017, pp. 1802--1811.

\bibitem{shotton2013scene}
J.~Shotton, B.~Glocker, C.~Zach, S.~Izadi, A.~Criminisi, and A.~Fitzgibbon,
  ``Scene coordinate regression forests for camera relocalization in rgb-d
  images,'' in \emph{Proceedings of the IEEE conference on computer vision and
  pattern recognition}, 2013, pp. 2930--2937.

\bibitem{li2018megadepth}
Z.~Li and N.~Snavely, ``Megadepth: Learning single-view depth prediction from
  internet photos,'' in \emph{Proceedings of the IEEE conference on computer
  vision and pattern recognition}, 2018, pp. 2041--2050.

\bibitem{dai2017Scannet}
A.~Dai, A.~X. Chang, M.~Savva, M.~Halber, T.~Funkhouser, and M.~Nie{\ss}ner,
  ``Scannet: Richly-annotated 3d reconstructions of indoor scenes,'' in
  \emph{Proceedings of the IEEE conference on computer vision and pattern
  recognition}, 2017, pp. 5828--5839.

\bibitem{wu2023diff}
Q.~Wu, H.~Jiang, Y.~Ding, L.~Luo, J.~Xie, and J.~Yang, ``Diff-pcr:
  Diffusion-based correspondence searching in doubly stochastic matrix space
  for point cloud registration,'' \emph{arXiv preprint arXiv:2401.00436}, 2023.

\bibitem{wu2024diff}
Q.~Wu, H.~Jiang, L.~Luo, J.~Li, Y.~Ding, J.~Xie, and J.~Yang, ``Diff-reg:
  Diffusion model in doubly stochastic matrix space for registration problem,''
  in \emph{European Conference on Computer Vision}.\hskip 1em plus 0.5em minus
  0.4em\relax Springer, 2024, pp. 160--178.

\bibitem{yu2022riga}
H.~Yu, J.~Hou, Z.~Qin, M.~Saleh, I.~Shugurov, K.~Wang, B.~Busam, and S.~Ilic,
  ``Riga: Rotation-invariant and globally-aware descriptors for point cloud
  registration,'' \emph{arXiv preprint arXiv:2209.13252}, 2022.

\bibitem{yu2023peal}
J.~Yu, L.~Ren, Y.~Zhang, W.~Zhou, L.~Lin, and G.~Dai, ``Peal: Prior-embedded
  explicit attention learning for low-overlap point cloud registration,'' in
  \emph{CVPR}, 2023.

\bibitem{deng2018ppf}
H.~Deng, T.~Birdal, and S.~Ilic, ``Ppf-foldnet: Unsupervised learning of
  rotation invariant 3d local descriptors,'' in \emph{Proceedings of the
  European Conference on Computer Vision (ECCV)}, 2018.

\bibitem{Li2021DeepI2PIC}
J.~Li and G.~H. Lee, ``Deepi2p: Image-to-point cloud registration via deep
  classification,'' \emph{2021 IEEE/CVF Conference on Computer Vision and
  Pattern Recognition (CVPR)}, 2021.

\bibitem{Wang2021P2NetJD}
B.~Wang, C.~Chen, Z.~Cui, J.~Qin, C.~X. Lu, Z.~Yu, P.~Zhao, Z.~Dong, F.~Zhu,
  N.~Trigoni, and A.~Markham, ``P2-net: Joint description and detection of
  local features for pixel and point matching,'' \emph{2021 IEEE/CVF
  International Conference on Computer Vision (ICCV)}, 2021.

\bibitem{Wang2023FreeRegIC}
H.~Wang, Y.~Liu, B.~Wang, Y.~Sun, Z.~Dong, W.~Wang, and B.~Yang, ``Freereg:
  Image-to-point cloud registration leveraging pretrained diffusion models and
  monocular depth estimators,'' \emph{ArXiv}, 2023.

\bibitem{oquab2023dinov2}
M.~Oquab, T.~Darcet, T.~Moutakanni, H.~Vo, M.~Szafraniec, V.~Khalidov,
  P.~Fernandez, D.~Haziza, F.~Massa, A.~El-Nouby \emph{et~al.}, ``Dinov2:
  Learning robust visual features without supervision,'' \emph{arXiv preprint
  arXiv:2304.07193}, 2023.

\bibitem{song2019generative}
Y.~Song and S.~Ermon, ``Generative modeling by estimating gradients of the data
  distribution,'' \emph{Advances in neural information processing systems},
  2019.

\bibitem{Gong2022DiffPoseTM}
J.~Gong, L.~G. Foo, Z.~Fan, Q.~Ke, H.~Rahmani, and J.~Liu, ``Diffpose: Toward
  more reliable 3d pose estimation,'' \emph{2023 IEEE/CVF Conference on
  Computer Vision and Pattern Recognition (CVPR)}, 2022.

\bibitem{Shan2023DiffusionBased3H}
W.~Shan, Z.~Liu, X.~Zhang, Z.~Wang, K.~Han, S.~Wang, S.~Ma, and W.~Gao,
  ``Diffusion-based 3d human pose estimation with multi-hypothesis
  aggregation,'' \emph{2023 IEEE/CVF International Conference on Computer
  Vision (ICCV)}, 2023.

\bibitem{Wang2023PoseDiffusionSP}
J.~Wang, C.~Rupprecht, and D.~Novotn{\'y}, ``Posediffusion: Solving pose
  estimation via diffusion-aided bundle adjustment,'' \emph{2023 IEEE/CVF
  International Conference on Computer Vision (ICCV)}, 2023.

\bibitem{chen2023diffusiondet}
S.~Chen, P.~Sun, Y.~Song, and P.~Luo, ``Diffusiondet: Diffusion model for
  object detection,'' in \emph{Proceedings of the IEEE/CVF International
  Conference on Computer Vision}, 2023.

\bibitem{Baranchuk2021LabelEfficientSS}
D.~Baranchuk, I.~Rubachev, A.~Voynov, V.~Khrulkov, and A.~Babenko,
  ``Label-efficient semantic segmentation with diffusion models,''
  \emph{ArXiv}, 2021.

\bibitem{Gu2022DiffusionInstDM}
Z.~Gu, H.~Chen, Z.~Xu, J.~Lan, C.~Meng, and W.~Wang, ``Diffusioninst: Diffusion
  model for instance segmentation,'' \emph{ArXiv}, 2022.

\bibitem{parisi1981correlation}
G.~Parisi, ``Correlation functions and computer simulations,'' \emph{Nuclear
  Physics B}, 1981.

\bibitem{harris1988combined}
C.~Harris, M.~Stephens \emph{et~al.}, ``A combined corner and edge detector,''
  in \emph{Alvey vision conference}, vol.~15, no.~50.\hskip 1em plus 0.5em
  minus 0.4em\relax Citeseer, 1988, pp. 10--5244.

\bibitem{lowe2004distinctive}
D.~G. Lowe, ``Distinctive image features from scale-invariant keypoints,''
  \emph{International journal of computer vision}, vol.~60, pp. 91--110, 2004.

\bibitem{detone2018superpoint}
D.~DeTone, T.~Malisiewicz, and A.~Rabinovich, ``Superpoint: Self-supervised
  interest point detection and description,'' in \emph{Proceedings of the IEEE
  conference on computer vision and pattern recognition workshops}, 2018, pp.
  224--236.

\bibitem{chen2021learning}
H.~Chen, Z.~Luo, J.~Zhang, L.~Zhou, X.~Bai, Z.~Hu, C.-L. Tai, and L.~Quan,
  ``Learning to match features with seeded graph matching network,'' in
  \emph{Proceedings of the IEEE/CVF international conference on computer
  vision}, 2021, pp. 6301--6310.

\bibitem{shi2022clustergnn}
Y.~Shi, J.-X. Cai, Y.~Shavit, T.-J. Mu, W.~Feng, and K.~Zhang, ``Clustergnn:
  Cluster-based coarse-to-fine graph neural network for efficient feature
  matching,'' in \emph{Proceedings of the IEEE/CVF conference on computer
  vision and pattern recognition}, 2022, pp. 12\,517--12\,526.

\bibitem{zhang2024diffglue}
S.~Zhang and J.~Ma, ``Diffglue: Diffusion-aided image feature matching,'' in
  \emph{Proceedings of the 32nd ACM International Conference on Multimedia},
  2024, pp. 8451--8460.

\bibitem{wang2022matchformer}
Q.~Wang, J.~Zhang, K.~Yang, K.~Peng, and R.~Stiefelhagen, ``Matchformer:
  Interleaving attention in transformers for feature matching,'' in
  \emph{Proceedings of the Asian Conference on Computer Vision}, 2022, pp.
  2746--2762.

\bibitem{song2020score}
Y.~Song, J.~Sohl-Dickstein, D.~P. Kingma, A.~Kumar, S.~Ermon, and B.~Poole,
  ``Score-based generative modeling through stochastic differential
  equations,'' \emph{arXiv preprint arXiv:2011.13456}, 2020.

\bibitem{SohlDickstein2015DeepUL}
J.~N. Sohl-Dickstein, E.~A. Weiss, N.~Maheswaranathan, and S.~Ganguli, ``Deep
  unsupervised learning using nonequilibrium thermodynamics,'' \emph{ArXiv},
  2015.

\bibitem{arun1987least}
K.~S. Arun, T.~S. Huang, and S.~D. Blostein, ``Least-squares fitting of two 3-d
  point sets,'' \emph{IEEE Transactions on pattern analysis and machine
  intelligence}, 1987.

\bibitem{vaswani2017attention}
A.~Vaswani, N.~Shazeer, N.~Parmar, J.~Uszkoreit, L.~Jones, A.~N. Gomez,
  {\L}.~Kaiser, and I.~Polosukhin, ``Attention is all you need,''
  \emph{Advances in neural information processing systems}, 2017.

\bibitem{mildenhall2021nerf}
B.~Mildenhall, P.~P. Srinivasan, M.~Tancik, J.~T. Barron, R.~Ramamoorthi, and
  R.~Ng, ``Nerf: Representing scenes as neural radiance fields for view
  synthesis,'' \emph{Communications of the ACM}, 2021.

\bibitem{su2024roformer}
J.~Su, M.~Ahmed, Y.~Lu, S.~Pan, W.~Bo, and Y.~Liu, ``Roformer: Enhanced
  transformer with rotary position embedding,'' \emph{Neurocomputing}, vol.
  568, p. 127063, 2024.

\bibitem{lai2014unsupervised}
K.~Lai, L.~Bo, and D.~Fox, ``Unsupervised feature learning for 3d scene
  labeling,'' in \emph{IEEE International Conference on Robotics and Automation
  (ICRA)}, 2014.

\bibitem{song2023consistency}
\BIBentryALTinterwordspacing
Y.~Song, P.~Dhariwal, M.~Chen, and I.~Sutskever, ``Consistency models,'' 2023.
  [Online]. Available: \url{https://arxiv.org/abs/2303.01469}
\BIBentrySTDinterwordspacing

\bibitem{liu2022flow}
X.~Liu, C.~Gong, and Q.~Liu, ``Flow straight and fast: Learning to generate and
  transfer data with rectified flow,'' \emph{arXiv preprint arXiv:2209.03003},
  2022.

\bibitem{li20214dcomplete}
Y.~Li, H.~Takehara, T.~Taketomi, B.~Zheng, and M.~Nie{\ss}ner, ``4dcomplete:
  Non-rigid motion estimation beyond the observable surface,'' in \emph{ICCV},
  2021.

\bibitem{wu2019pointpwc}
W.~Wu, Z.~Wang, Z.~Li, W.~Liu, and L.~Fuxin, ``Pointpwc-net: A coarse-to-fine
  network for supervised and self-supervised scene flow estimation on 3d point
  clouds,'' \emph{arXiv preprint arXiv:1911.12408}, 2019.

\bibitem{puy2020flot}
G.~Puy, A.~Boulch, and R.~Marlet, ``Flot: Scene flow on point clouds guided by
  optimal transport,'' in \emph{European conference on computer vision}, 2020.

\bibitem{qin2023deep}
Z.~Qin, H.~Yu, C.~Wang, Y.~Peng, and K.~Xu, ``Deep graph-based spatial
  consistency for robust non-rigid point cloud registration,'' in
  \emph{Proceedings of the IEEE/CVF Conference on Computer Vision and Pattern
  Recognition}, 2023.

\bibitem{lee2011hyper}
J.~Lee, M.~Cho, and K.~M. Lee, ``Hyper-graph matching via reweighted random
  walks,'' in \emph{CVPR}, 2011.

\bibitem{choy2019fully}
C.~Choy, J.~Park, and V.~Koltun, ``Fully convolutional geometric features,'' in
  \emph{Proceedings of the IEEE/CVF International Conference on Computer
  Vision}, 2019.

\bibitem{li2012robust}
S.~Li, C.~Xu, and M.~Xie, ``A robust o (n) solution to the perspective-n-point
  problem,'' \emph{IEEE transactions on pattern analysis and machine
  intelligence}, 2012.

\end{thebibliography}

\end{document}